\documentclass{article}

\usepackage[final]{neurips2021}
\usepackage{times}
\usepackage{latexsym}
\usepackage[T1]{fontenc}
\usepackage{multicol}
\usepackage{microtype}

\usepackage{times}
\usepackage{latexsym}
\usepackage{graphicx}
\usepackage{tikz}
\usepackage{enumitem}
\usepackage{tikz-qtree}
\usepackage{amsmath}
\usepackage{amssymb}
\usepackage{amsthm}
\usepackage{pgfplots}
\usepackage{booktabs}
\usepackage{tipa}
\usepackage{hyperref}
\usepackage{inconsolata}

\usepackage{caption}
\usepackage{subcaption}
\usepackage[T1]{fontenc}

\usepackage[utf8]{inputenc}

\theoremstyle{definition}
\newtheorem{definition}{Definition}[section]

\newcommand{\secref}[1]{Section~\ref{#1}}
\newcommand{\Appref}[1]{Appendix~\ref{#1}}
\newcommand{\appref}[1]{Appendix~\ref{#1}}

\newcommand{\Figref}[1]{Figure~\ref{#1}}
\newcommand{\figref}[1]{Figure~\ref{#1}}
\newcommand{\dashfigref}[2]{Figures~\ref{#1}--\ref{#2}}

\newcommand{\tabref}[1]{Table~\ref{#1}}

\newcommand{\MQNLI}{\text{MQNLI}}

\newcommand{\nodevar}{\mathit{N}}

\renewcommand{\textit}[1]{\emph{#1}}

\newcommand{\addnet}{N_+}
\newcommand{\nlinet}{N_{\emph{NLI}}}
\newcommand{\addmod}{C_+}
\newcommand{\natmod}{C_{\emph{NatLog}}}

\newcommand{\Subj}{\text{Subj}}
\newcommand{\Obj}{\text{Obj}}

\newcommand{\QSubj}{\text{Q}_\text{Subj}}
\newcommand{\AdjSubj}{\text{Adj}_\text{Subj}}
\newcommand{\NSubj}{\text{N}_\text{Subj}}
\newcommand{\NPSubj}{\text{NP}_\text{Subj}}
\newcommand{\Neg}{\text{Neg}}
\newcommand{\NegP}{\text{NegP}}
\newcommand{\Adv}{\text{Adv}}
\newcommand{\V}{\text{V}}
\newcommand{\VP}{\text{VP}}
\newcommand{\QObj}{\text{Q}_\text{Obj}}
\newcommand{\QPObj}{\text{QP}_\text{Obj}}
\newcommand{\AdjObj}{\text{Adj}_\text{Obj}}
\newcommand{\NObj}{\text{N}_\text{Obj}}
\newcommand{\NPObj}{\text{NP}_\text{Obj}}
\newcommand{\AObj}{\text{A}_\text{Obj}}

\title{Causal Abstractions of Neural Networks}

\author{Atticus Geiger\thanks{equal contribution}, Hanson Lu\footnotemark[1], Thomas Icard, and Christopher Potts \\
  Stanford \\
  Stanford, CA 94305-2150\\
  \texttt{\{atticusg, hansonlu, icard, cgpotts\}@stanford.edu} \\
}

\begin{document}
\setlength{\abovedisplayskip}{5.5pt}
\setlength{\belowdisplayskip}{5.5pt}
\maketitle

\begin{abstract}
Structural analysis methods (e.g., probing and feature attribution) are increasingly important tools for neural network analysis. We propose a new structural analysis method grounded in a formal theory of \textit{causal abstraction} that provides rich characterizations of model-internal representations and their roles in input/output behavior. In this method, neural representations are aligned with variables in interpretable causal models, and then \textit{interchange interventions} are used to experimentally verify that the neural representations have the causal properties of their aligned variables. We apply this method in a case study to analyze neural models trained on Multiply Quantified Natural Language Inference (MQNLI) corpus, a highly complex NLI dataset that was constructed with a tree-structured natural logic causal model. We discover that a BERT-based model with state-of-the-art performance successfully realizes parts of the natural logic model's causal structure, whereas a simpler baseline model fails to show any such structure, demonstrating that BERT representations encode the compositional structure of MQNLI.
\end{abstract}

\section{Introduction}\label{sec:intro}

Explainability and interpretability have long been central issues for
neural networks, and they have taken on renewed importance as such models are now ubiquitous
in research and technology. Recent structural evaluation methods seek to reveal the internal structure of these ``black box'' models. Structural methods include probes, attributions (feature importance methods), and interventions (manipulations
of model-internal states). These methods can complement standard behavioral techniques (e.g., performance on gold evaluation sets), and they can yield insights into how and why models make the predictions they do.  However, these tools have their limitations, and it has often been assumed that more ambitious and systematic causal analysis of such models is beyond reach.

Although there is a sense in which neural networks are ``black boxes'', they have the virtue of being completely closed and controlled systems. This means that standard empirical challenges of causal inference due to lack of observability simply do not arise. The challenge is rather to identify high-level causal regularities that \emph{abstract away} from irrelevant (but arbitrarily observable and manipulable) low-level details. Our contribution in this paper is to show that this challenge can be met. Drawing on recent innovations in the formal theory of causal abstraction \citep{Beckers_Halpern_2019,beckers20a,chalupka16,Rubensteinetal17}, we offer a methodology for meaningful causal explanations of neural network behavior.



Our methodology \textit{causal abstraction analysis}\footnote{We provide tools for causal abstraction analysis at \url{http://github.com/hansonhl/antra} and the code base for this paper at \url{http://github.com/atticusg/Interchange} } consists of three stages. (1) Formulate a hypothesis by defining a causal model that might explain network behavior. Candidate causal models can be naturally adapted from theoretical and empirical modeling work in linguistics and cognitive sciences. (2) Search for an alignment between neural representations in the network and variables in the high-level causal model. (3)~Verify experimentally that the neural representations have the same causal properties as their aligned high-level variables using the \emph{interchange intervention} method of \citet{geiger-etal-2020-neural}.

As a case study, we apply this methodology to LSTM-based and BERT-based natural language inference (NLI) models trained on the logically complex Multiply Quantified NLI (MQNLI) dataset of \citet{Geiger-etal:2019}. This challenging dataset was constructed with a tree-structured natural logic causal model \citep{MacCartney:07,vanBenthem:08,Icard:Moss:2013:LILT}. Our BERT-based model has the structure of a standard NLI classifier, and yet it is able to perform well on MQNLI (88\%), a result \citeauthor{Geiger-etal:2019}\ achieved only with highly customized task-specific models. By contrast, our LSTM-based model is much less successful (46\%).

The obvious scientific question in this case study is what drives the success of the BERT-based model on this challenging task. To answer this we employ our methodology. (1) We formulate hypotheses by defining simplified variants of the natural logic causal model. (2) We search over potential alignments between neural representations in BERT and variables in our high-level causal models. (3) We perform interchange interventions on the BERT model for each alignment. We find that our BERT model partially realizes the causal structure of the natural logic causal model; crucially, the LSTM model does not. High-level causal explanation for system behavior is often considered a gold standard for interpretability, one that may be thought quixotic for complex neural models \citep{lillicrap2019does}. The point of our case study is to show that this high standard can be achieved.

We conclude by comparing our methodology to probing and the attribution method of integrated gradients \citep{sundararajan17a}. We argue probing is unable to provide a causal characterization of models. We show formally that attribution methods do measure causal properties, and in that way they are similar to the tool of interchange interventions. However, our methodology of causal abstraction analysis provides a framework for systematically measuring and aggregating such causal properties in order to evaluate a precise hypothesis about abstract causal structure.

\section{Related Work}\label{sec:relatedwork}

\begin{figure*}[t]
  \centering
  \begin{subfigure}[t]{0.96\textwidth}
  \centering
\begin{tikzpicture}[thick,scale=0.55, every node/.style={scale=0.55}]
\node[minimum size=25pt] (x) at (0,0.3) {$x$};
\node[minimum size=25pt] (y) at (2,0.3) {$y$};
\node[minimum size=25pt] (z) at (4,0.3) {$z$};

\node[fill=gray!20, draw, rectangle, minimum size=25pt] (N00) at (0,1.5) {$D_{x}$};
\node[fill=gray!20, draw, rectangle, minimum size=25pt] (N01) at (2,1.5) {$D_{y}$};
\node[fill=gray!20, draw, rectangle, minimum size=25pt] (N02) at (4,1.5) {$D_{z}$};

\node[fill=red!20, draw, rectangle, minimum size=25pt] (N10) at (0,3.5) {$L_{2}$};
\node[fill=gray!20, draw, rectangle, minimum size=25pt] (N11) at (2,3.5) {};
\node[fill=red!20, draw, rectangle, minimum size=25pt] (N12) at (4,3.5) {$L_{1}$};

\node[fill=gray!20, draw, rectangle, minimum size=25pt] (output) at (2,5.2) {$O$};

\draw [-] (x) -- (N00) ;
\draw [-] (y) -- (N01) ;
\draw [-] (z) -- (N02) ;

\draw [->] (N00) -- (N10) ;
\draw [->] (N00) -- (N11) ;
\draw [->] (N00) -- (N12) ;

\draw [->] (N01) -- (N10) ;
\draw [->] (N01) -- (N11) ;
\draw [->] (N01) -- (N12) ;

\draw [->] (N02) -- (N10) ;
\draw [->] (N02) -- (N11) ;
\draw [->] (N02) -- (N12) ;

\draw [->] (N10) -- (output) ;
\draw [->] (N11) -- (output) ;
\draw [->] (N12) -- (output) ;

\def\x{10}
\def\y{1.5}
\node[fill=green!20, draw, circle, minimum size=35pt] (D1) at (0+\x,0+\y) {\large $X$};
\node[fill=green!20, draw, circle, minimum size=35pt] (D2) at (2+\x,0+\y) {\large$Y$};
\node[fill=green!20, draw, circle, minimum size=35pt] (D3) at (4+\x,0+\y) {\large$Z$};
\node[fill=green!20, draw, circle, minimum size=35pt] (D4) at (3+\x,1.5+\y) {\large$W$};
\node[fill=green!20, draw, circle, minimum size=35pt] (S1) at (1+\x,1.5+\y) {\large$S_1$};
\node[fill=green!20, draw, circle, minimum size=35pt] (S2) at (2+\x,3+\y) {\large$S_2$};
\draw [->] (D1) -- (S1);
\draw [->] (D2) -- (S1);
\draw [->] (S1) -- (S2);
\draw [->] (D4) -- (S2);
\draw [->] (D3) -- (D4);

\draw[dashed](S1) to[bend right=15pt] (N12);
\draw[dashed](D4) to[bend right=22pt] (N10);
\end{tikzpicture}
  \caption{The causal model $\addmod$ (right) that first computes $S_1 = X + Y$ and $W = Z$, before computing the final output $S_2 = W + S_1$ aligned with the neural network $\addnet$ (left) with $L_{1}$ highlighted as the hypothesized location encoding $S_1 = X + Y$ and $L_{2}$ as the location encoding $W = Z$.}
  \label{fig:addalign}
  \end{subfigure} 
  
  \begin{subfigure}[t]{0.48\textwidth}
  \centering
      \resizebox{\textwidth}{!}{
\begin{tikzpicture}[thick,scale=0.6, every node/.style={scale=0.6}]
\node[minimum size=25pt] (ax) at (0,0.3) {$1$};
\node[minimum size=25pt] (ay) at (2,0.3) {$2$};
\node[minimum size=25pt] (az) at (4,0.3) {$3$};

\node[fill=gray!20, draw, rectangle, minimum size=25pt] (aN00) at (0,1.5) {$D_{1}$};
\node[fill=gray!20, draw, rectangle, minimum size=25pt] (aN01) at (2,1.5) {$D_{2}$};
\node[fill=gray!20, draw, rectangle, minimum size=25pt] (aN02) at (4,1.5) {$D_{3}$};

\node[fill=gray!20, draw, rectangle, minimum size=25pt] (aN10) at (0,3.5) {};
\node[fill=gray!20, draw, rectangle, minimum size=25pt] (aN11) at (2,3.5) {};
\node[fill=orange!20, draw, rectangle, minimum size=25pt] (aN12) at (4,3.5) {$L_{1}$};

\node[fill=blue!30, draw, rectangle, minimum size=25pt] (aoutput) at (2,5.2) {$12$};

\draw [-] (ax) -- (aN00) ;
\draw [-] (ay) -- (aN01) ;
\draw [-] (az) -- (aN02) ;

\draw [->] (aN00) -- (aN10) ;
\draw [->] (aN00) -- (aN11) ;
\draw [->] (aN00) -- (aN12) ;

\draw [->] (aN01) -- (aN10) ;
\draw [->] (aN01) -- (aN11) ;
\draw [->] (aN01) -- (aN12) ;

\draw [->] (aN02) -- (aN10) ;
\draw [->] (aN02) -- (aN11) ;
\draw [->] (aN02) -- (aN12) ;

\draw [->] (aN10) -- (aoutput) ;
\draw [->] (aN11) -- (aoutput) ;
\draw [->] (aN12) -- (aoutput) ;

\def\x{0}

\node[minimum size=25pt] (bx) at (6+\x,0.3) {$4$};
\node[minimum size=25pt] (by) at (8+\x,0.3) {$5$};
\node[minimum size=25pt] (bz) at (10+\x,0.3) {$6$};

\node[fill=orange!20, draw, rectangle, minimum size=25pt] (bN00) at (6+\x,1.5) {$D_{4}$};
\node[fill=orange!20, draw, rectangle, minimum size=25pt] (bN01) at (8+\x,1.5) {$D_{5}$};
\node[fill=orange!20, draw, rectangle, minimum size=25pt] (bN02) at (10+\x,1.5) {$D_{6}$};

\node[fill=orange!20, draw, rectangle, minimum size=25pt] (bN10) at (6+\x,3.5) {};
\node[fill=orange!20, draw, rectangle, minimum size=25pt] (bN11) at (8+\x,3.5) {};
\node[fill=orange!20, draw, rectangle, minimum size=25pt] (bN12) at (10+\x,3.5) {$L_{1}$};

\node[fill=orange!20, draw, rectangle, minimum size=25pt] (boutput) at (8+\x,5.2) {$15$};

\draw [-] (bx) -- (bN00) ;
\draw [-] (by) -- (bN01) ;
\draw [-] (bz) -- (bN02) ;

\draw [->] (bN00) -- (bN10) ;
\draw [->] (bN00) -- (bN11) ;
\draw [->] (bN00) -- (bN12) ;

\draw [->] (bN01) -- (bN10) ;
\draw [->] (bN01) -- (bN11) ;
\draw [->] (bN01) -- (bN12) ;

\draw [->] (bN02) -- (bN10) ;
\draw [->] (bN02) -- (bN11) ;
\draw [->] (bN02) -- (bN12) ;

\draw [->] (bN10) -- (boutput) ;
\draw [->] (bN11) -- (boutput) ;
\draw [->] (bN12) -- (boutput) ;

\path(bN12.north) edge[->, dotted, bend right=20pt, thick] (aN12.north);

\end{tikzpicture}}

\caption{Low-level neural network interchange intervention. The network processes two different input sequences. The neural representation created at $L_{1}$ for input sequence $(1, 2, 3)$ is replaced by the corresponding representation created for input sequence $(4, 5, 6)$.}

\label{fig:addnet-ii}
  \end{subfigure}
  \hfill
  \begin{subfigure}[t]{0.48\textwidth}
  \centering
    \resizebox{\textwidth}{!}{
\begin{tikzpicture}[thick,scale=0.6, every node/.style={scale=0.6}]
\node (trash) at (0,0) {};
\def\x{0}
\def\y{1.9}
\node[fill=purple!20, draw, circle, minimum size=35pt] (D1) at (0+\x,0+\y) {\large $1$};
\node[fill=purple!20, draw, circle, minimum size=35pt] (D2) at (2+\x,0+\y) {\large$2$};
\node[fill=purple!20, draw, circle, minimum size=35pt] (D3) at (4+\x,0+\y) {\large$3$};
\node[fill=purple!20, draw, circle, minimum size=35pt] (D4) at (3+\x,1.5+\y) {\large$3$};
\node[fill=green!20, draw, circle, minimum size=35pt] (S1) at (1+\x,1.5+\y) {\large$9$};
\node[fill=brown!20, draw, circle, minimum size=35pt] (S2) at (2+\x,3+\y) {\large$12$};
\draw [->] (D1) -- (S1);
\draw [->] (D2) -- (S1);
\draw [->] (S1) -- (S2);
\draw [->] (D4) -- (S2);
\draw [->] (D3) -- (D4);
\def\x{6}
\def\y{1.9}
\node[fill=green!20, draw, circle, minimum size=35pt] (D12) at (0+\x,0+\y) {\large $4$};
\node[fill=green!20, draw, circle, minimum size=35pt] (D22) at (2+\x,0+\y) {\large$5$};
\node[fill=green!20, draw, circle, minimum size=35pt] (D32) at (4+\x,0+\y) {\large$6$};
\node[fill=green!20, draw, circle, minimum size=35pt] (D42) at (3+\x,1.5+\y) {\large$6$};
\node[fill=green!20, draw, circle, minimum size=35pt] (S12) at (1+\x,1.5+\y) {\large$9$};
\node[fill=green!20, draw, circle, minimum size=35pt] (S22) at (2+\x,3+\y) {\large$15$};
\draw [->] (D12) -- (S12);
\draw [->] (D22) -- (S12);
\draw [->] (S12) -- (S22);
\draw [->] (D42) -- (S22);
\draw [->] (D32) -- (D42);

\path(S12.west) edge[->, dotted, bend right=40pt, thick] (S1.east);

\end{tikzpicture}}

\caption{High-level symbolic computation interchange intervention. The computation processes two different input sequences. The sum at $S_1$ for input sequence $(1, 2, 3)$ is replaced by the corresponding representation created for input sequence $(4, 5, 6)$.}

\label{fig:addcomp-ii}
  \end{subfigure}

 \caption{Our motivating example where we hypothesis that a symbolic computation $C_+$ is a causal abstraction of a neural network $N_+$ under a particular alignment (top). We can experimentally confirm this hypothesis by conducting an interchange intervention on both the network and the computation with every pair of inputs and evaluating whether the intervened network and intervened computation have the same counterfactual output behavior. We schematically depict an interchange intervention on the network $N_+$ (bottom left) and the computation $C_+$ (bottom right) with the base input $(1,2,3)$ and the source input $(4,5,6)$. Observe that the output of the intervened neural network matches the output of the intervened symbolic computation, so we have success for this pair of inputs.}
 
 \label{fig:addmod-example}
\end{figure*}
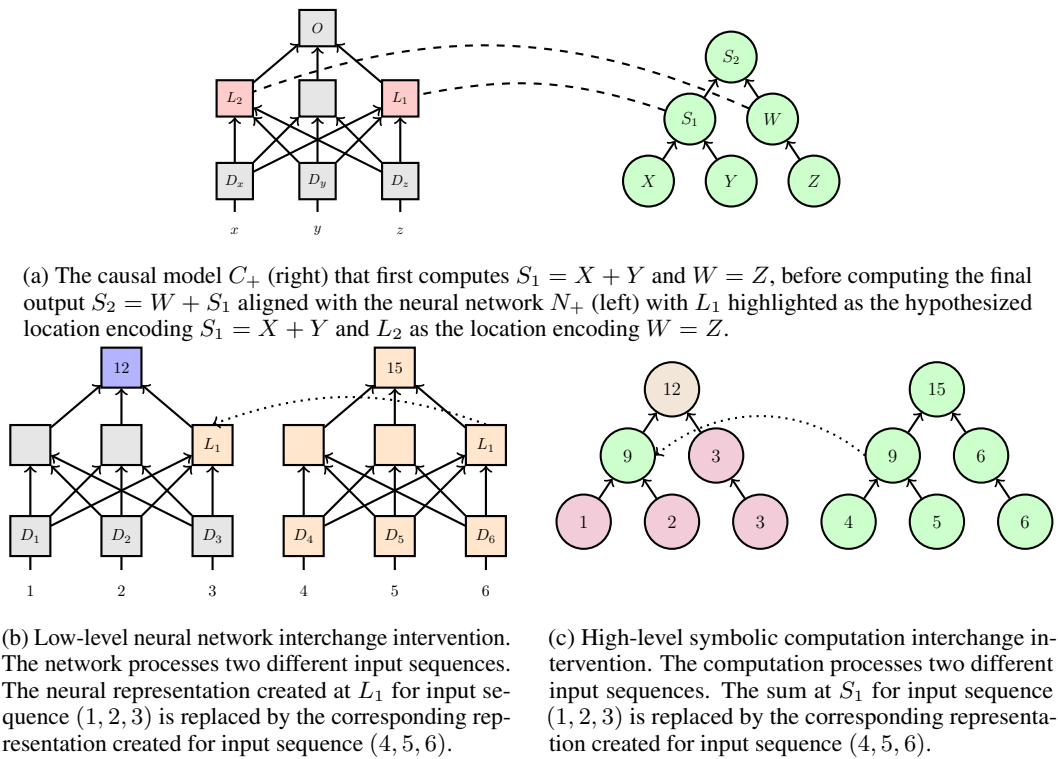

\paragraph{Probes}

Probes are generally supervised models trained on the internal representations of networks with the goal of determining what those internal representations encode \citep{clark-etal-2019-bert,hupkes-etal-2018-analysing, peters-etal-2018-dissecting,tenney-etal-2019-bert}. Probes are fundamentally unable to directly measure causal properties of neural representations, and \citet{ravichander2020probing}, \citet{elazar-etal-2020}, and \citet{geiger-etal-2020-neural} have argued that probes are limited in their ability to provide even indirect evidence of causal properties.


We now present an analytic example in which probing identifies seemingly crucial information in representations that have no causal impact on behavior. We assume the structure of the simple addition network $\addnet$ in \figref{fig:addmod-example}. For our embedding, we simply map every integer $i$ in $\mathbb{N}_{9}$ to the 1-dimensional vector $[i]$. The weight matrices are
\[\small
W_{1} = 
\left(
\begin{array}{c}
1 \\ 1 \\ 0 
\end{array}
\right)
\quad
W_{2} = 
\left(
\begin{array}{c}
1 \\ 1 \\ 1
\end{array}
\right)
\quad
W_{3} = 
\left(
\begin{array}{c}
0 \\ 0 \\ 1
\end{array}
\right)
\quad
\mathbf{w} = 
\left(
\begin{array}{c}
0 \\ 1 \\ 0 
\end{array}
\right)
\]
The output for an input sequence $\mathbf{x} = (i, j, k)$ is given by $\left(\mathbf{x}W_{1} ; \mathbf{x}W_{2} ;\mathbf{x}W_{3}\right)\mathbf{w}$.

In this network, $\mathbf{x}W_{1}$ perfectly encodes $i + j$, and $\mathbf{x}W_{3}$ perfectly encodes $k$. Thus, the identity model probe will be perfect in probing those representations for this information. However, neither representation plays a causal role in the network behavior; only $\mathbf{x}W_{2}$ contributes to the output.

\paragraph{Attribution Methods}

Attribution methods aim to quantify the degree to which a network representation contributes to the output prediction of the model, for a specific example or set of examples \citep{Binder16,Shrikumar16,springerberg2014,sundararajan17a,Zeiler2014}.  In contrast to probing, the well known integrated gradients method (IG) can be given an unambiguous causal interpretation. Following \cite{sundararajan17a} we define the vector $\textit{IG}(\mathbf{x})$, for an input $\mathbf{x}$ relative to a baseline $\mathbf{b}$, to have $i$th component $\textit{IG}_i(\mathbf{x})$ given by the expression on the left: \begin{eqnarray*} (x_i - b_i) \cdot \int_{\alpha = 0}^1 \frac{\partial F(\alpha\mathbf{x}+(1-\alpha)\mathbf{b})}{\partial x_i} d\alpha &=&  (x_i - b_i) \cdot \int_{\alpha = 0}^1 \underset{\epsilon \rightarrow 0}{\textnormal{lim}} \frac{F(\mathbf{x}^{\alpha,\epsilon}) - F(\mathbf{x}^{\alpha})}{\epsilon} d\alpha \label{ig}
\end{eqnarray*} 
Abbreviating the weighted average  $\alpha\mathbf{x}+(1-\alpha)\mathbf{b}$ by $\mathbf{x}^{\alpha}$,  letting $\mathbf{x}^{\alpha,\epsilon}$ be the vector that differs from $\mathbf{x}^{\alpha}$ in that the $i$th coordinate is increased by $\epsilon$, and then expanding the definition of partial derivative, this can be written in the form given on the right. The difference $F(\mathbf{x}^{\alpha,\epsilon}) - F(\mathbf{x}^{\alpha})$ is known in the causal literature as the (individual) \emph{causal effect} on the output (e.g., \cite{Imbens2015}) of increasing neuron $i$ by $\epsilon$ relative to the fixed input 
$\mathbf{x}^{\alpha}$.
So, essentially, $\textit{IG}_i(\mathbf{x})$ is measuring the average ``limiting'' causal effect of increasing neuron $i$ along the straight line from the baseline vector to the input vector~$\mathbf{x}$, weighted by the difference at $i$ between input and baseline. More recently, \citet{chattopadhyay19a} develop an attribution method that explicitly treats neural models as structured causal models and directly computes the individual causal effect of a feature to determine its attribution. 


Attribution methods can measure causal properties, and, in that way, they are similar to the tool of interchange interventions. However, our methodology of causal abstraction analysis provides a framework for systematically measuring and aggregating such causal properties in order to evaluate a precise hypothesis about abstract causal structure.

\paragraph{Causal Abstraction}

Our goal is to evaluate whether the internal structure of a neural network realizes an abstract causal process. To concretize this, we turn to formal, broadly interventionist theories of causality \citep{spirtes,pearl}, in which causal processes are characterized by effects of interventions, and theories of abstraction \citep{Beckers_Halpern_2019,beckers20a,chalupka16,Rubensteinetal17} where relationships between two causal processes are determined by the presence of systematic correspondences between the effects of interventions.

%
%
The notion of abstraction that we employ here is a relatively simple one called  \emph{constructive abstraction} \citep{Beckers_Halpern_2019}. Informally, a high-level model is a constructive abstraction of a low-level model if there is a way to partition the variables in the low-level model where each high-level variable can be assigned to a low-level partition cell, such that there is a systematic correspondence between interventions on the low-level partition cells and interventions on the high-level variables. 

There are two properties of constructive abstraction that make it ideal for neural network analysis. First, the information content of partition cells of low-level variables can be determined by the high-level variables that they correspond to. For neural networks, the  partition cells of low-level variables are sets of neurons, and our method supports reasoning at the level of vector representations (sets of neurons). Second, the causal dependencies between partitions of low-level variables are not necessarily preserved as causal dependencies between the high-level variables corresponding to these partitions. For example, the low-level model might be a fully connected neural network, whereas the high-level model might have much sparser connections.
For neural network analysis, this means we can find causal abstractions that have far simpler causal structures than the underlying neural networks. We provide an example in the next section.


\section{Causal Abstraction Analysis of Neural Networks}\label{sec:causality}

We now describe our methodology in more detail, illustrating the relevant concepts with an example of a neural network performing basic arithmetic. Specifically, suppose that we have a neural network $\addnet$ that takes in three vector representations $D_x,D_y,D_z$ representing the integers $x$, $y$, and $z$, and outputs the sum of the three inputs: $\addnet(D_x,D_y,D_z) = x + y + z$. We seek an informative causal explanation of this network's behavior.

\paragraph{Formulating a Hypothesis}\label{sec:motivating-example}
A human performing this task might follow an algorithm in which they add together the first two numbers and then add that sum to the third number. We can hypothesize that the behavior of $\addnet$ is explained by this symbolic computation. Specifically, the network combines $D_{x}$ and $D_{y}$ to create an internal representation at some location $L_{1}$ encoding $x + y$; it encodes $z$ at some location $L_{2}$; and $L_{1}$ and $L_{2}$ are composed to encode $a + z$ at the location of the output representation. This hypothesis is given schematically in \figref{fig:addalign}. 

Following our methodology, we first define the causal model $\addmod$ in \figref{fig:addalign}. Our informal hypothesis that a neural network's behavior is explained by a simple algorithm can then be restated more formally: $\addmod$ is a constructive abstraction of the neural network $\addnet$. 


 

\paragraph{Alignment Search} Now that we have hypothesized that the causal model $\addmod$ is a causal abstraction of the network $\addnet$, the next step is to align the neural representations in $\addnet$ with the variables in $\addmod$. The input embeddings $D_x$, $D_y$, and $D_z$ must be aligned with the input variables $X$, $Y$, and $Z$ and the output neuron $O$ must be aligned with the output variable $S_2$. That leaves the intermediate variables $S_1$ and $W$ to be aligned with neural representations at some undetermined locations $L_1$ and $L_2$. If this were an actual experiment (see below), we would perform an \textit{alignment search} to consider many possible values for $L_1$ and $L_2$. Each alignment is a hypothesis about where the network $\addnet$ stores and uses the values of $S_1$ and $W$. For the example, we assume the alignment in  \figref{fig:addalign}.

\paragraph{Interchange Interventions}

Finally, for a given alignment, we experimentally determine whether the neural representations at $L_1$ and $L_2$ have the same causal properties as $S_1$ and $W$. The basic experimental technique is an \textit{interchange intervention}, in which a neural representation created during prediction on a ``base'' input is interchanged with the representation created for a ``source'' input \citep{geiger-etal-2020-neural}. We now show informally that this method can be used to prove that the causal model $\addmod$ is a constructive abstraction of the neural network $\addnet$ (\appref{app:addition} has formal details).

We first intervene on the causal model. Consider two inputs $\mathbf{a}, \mathbf{a}' \in (\mathbb{N}_9)^3$ where $\mathbb{N}_9$ is the set of integers 0--9. Let $\mathbf{a} = (x, y, z)$ be the base input and $\mathbf{a}' = (x', y', z')$ be the source input. Define 
\begin{equation}
\addmod^{S_1 \leftarrow \mathbf{a}'}(\mathbf{a}) = x' + y' + z
\label{eq:addmod-ii}
\end{equation}
to be the output provided by $\addmod$ when $S_1$, the variable representing the intermediate sum, is intervened on and set to the value $x' + y'$. Thus, for example, if the base input is $\addmod(1, 2, 3) = 6$, and the source input is $\mathbf{a}' = (4, 5, 6)$, then $\addmod^{S_1 \leftarrow \mathbf{a}'}(1, 2, 3) = 4 + 5 + 3 = 12$. This process is depicted in \figref{fig:addcomp-ii}.

Next, we intervene on the neural network $\addnet$. Let $\mathbf{D}$ be an embedding space that provides unique representations for $\mathbb{N}_9$, and consider two inputs $D = (D_{x}, D_{y}, D_{z})$ and $D' = (D_{x'}, D_{y'}, D_{z'})$, where all $D_{i}$ and $D_{i'}$ are drawn from $\mathbf{D}$. In parallel with \eqref{eq:addmod-ii}, define
\begin{equation}
\addnet^{L_{1} \leftarrow D'}(D)
\label{eq:addnet-ii}
\end{equation}
to be the output provided by $\addnet$ processing the input $D$ when the representation at location $L_{1}$ is replaced with the representation at location $L_{1}$ created when $\addnet$ is processing the input $D'$. This process is depicted in \figref{fig:addnet-ii}.

With these two definitions, we can define what it means to test the hypothesis that $\addnet$ computes $x + y$ at position $L_{1}$. Where $D_{\mathbf{a}}$ is an embedding for $\mathbf{a}$ and $D_{\mathbf{a}'}$ is an embedding for $\mathbf{a}'$, we test:
\begin{equation}
\addmod^{S_1 \leftarrow \mathbf{a}'}(\mathbf{a}) = \addnet^{L_1 \leftarrow D_{\mathbf{a}'}}(D_{\mathbf{a}})
\label{eq:add-linking}
\end{equation}
If this equality holds for all source and base inputs $\mathbf{a}$ and $\mathbf{a}'$, then we can conclude that, for every intervention on $S_1$, there is an equivalent intervention on $L_{1}$. If we can establish a corresponding claim for $W$ and $L_{2}$, then we have shown that $\addmod$ is a constructive abstraction of $\addnet$, since the inputs' relationships are established by our embedding and there are no other interventions on $\addmod$ to test.



\paragraph{Analysis}
Suppose that all of our intervention experiments verify our hypothesis that $\addmod$ is a constructive abstraction of $\addnet$ with variables $S_1$ and $W$ aligned to neural representations at $L_1$ and $L_2$. This explains network behavior by resolving two crucial questions.



First, we learn what information is encoded in the representations $L_1$ and $L_2$. Neural representations encode the values of the high-level variables they are aligned with. The location $L_1$ encodes the variable $S_1$ and the location $L_2$ encodes the variable $W$. This is similar to what probing achieves. However, our method is crucially different from probing. In probing, information content is established through purely correlational properties, meaning a neural representation with no causal role in network behavior can be successfully probed, as we showed in \secref{sec:relatedwork}. In causal abstraction analysis, information content is established through purely causal properties, ensuring that the neural representation is actually implicated in model behavior.

Second, we learn what causal role $L_1$ and $L_2$ play in network behavior.
Neural representations play a parallel causal role to their aligned high-level variables. At the location $L_1$, $D_x$ and $D_y$ are composed to form a neural representation with content $x+y$ that is then composed with $L_2$ to create an output. The fact that $S_1$ doesn't depend on $z$ tells us that while $L_1$ depends on $D_z$ and representations at $L_1$ may even correlate with $z$, the information about $z$ is not causally represented at $L_1$. At the location $L_2$, the value of $z$ is simply repeated and then composed with $L_1$ to create a final output.


Our method assigns causally impactful information content, but also identifies the abstract causal structure along which representations are composed. It thus encompasses and improves on both correlational (probing) and attribution methods.




\section{The Natural Language Inference Task and Models}\label{sec:datasetandmodels}


\paragraph{Multiply Quantified NLI Dataset}

The Multiply Quantified NLI (MQNLI) dataset of \citet{Geiger-etal:2019} contains templatically generated English-language NLI examples that involve very complex interactions between quantifiers, negation, and modifiers. We provide a few examples in \figref{tab:mqnli-examples}; the empty-string symbol $\varepsilon$ ensures perfect alignments at the token level both between premises and hypotheses and across all examples.

The MQNLI examples are labeled using an algorithmic implementation of the natural logic of \citet{maccartney-manning-2009-extended} over tree structures, and MQNLI has train/dev/test splits that vary in their difficulty. In the hardest setting, the train set is provably the minimal set of examples required to ensure that the dev and test sets can be perfectly solved by a simple symbolic model; in the easier settings, the train set redundantly encodes necessary information, which might allow a model to perform perfectly in assessment by memorization despite not having found a truly general solution. For a fuller review of the dataset, see \appref{app:mqnli-data}.

MQNLI is a fitting benchmark given our goals for a few reasons. First, we can focus on the hardest splits that can be generated, which will stress-test our NLI architectures in a standard behavioral way. Second, the MQNLI labeling algorithm itself suggests an appropriate causal model of the data-generating process. \figref{fig:bigtree} summarizes this model in tree form, and it is presented in full detail in \citet{Geiger-etal:2019}. This allows us to rigorously assess whether a neural network has learned to implement variants of this causal model. The complexity of the MQNLI examples creates many opportunities to do this in linguistically interesting ways.

\begin{figure}

\begin{subfigure}{0.98\textwidth}
  \centering
  \input{Figures/CompTree.tex}
  \caption{The causal structure of the high-level natural logic causal model $\natmod$ that performs inference on \MQNLI. The superscripts $P$ and $H$ stand for `premise' and `hypothesis' and the subscripts `Subj' and `Obj' stand for `Subject' and `Object'. The node labels are used to explain the experimental results in \secref{sec:experiments}}
  \label{fig:bigtree}
\end{subfigure}

\vspace{12pt}

\begin{subfigure}{0.53\textwidth}
\centering
\fontsize{8.2pt}{8.2pt}\selectfont
\begin{tabular}{c}
$\varepsilon$ every $\varepsilon$ baker $\varepsilon$ $\varepsilon$ $\varepsilon$ eats $\varepsilon$ no $\varepsilon$ bread \\
\textbf{contradiction} \\
$\varepsilon$ no angry baker $\varepsilon$ $\varepsilon$ $\varepsilon$ eats $\varepsilon$ no $\varepsilon$ bread \\
\\
$\varepsilon$ every silly professor $\varepsilon$ $\varepsilon$ $\varepsilon$ sells not every $\varepsilon$ book \\
\textbf{neutral}\\
$\varepsilon$ every silly professor $\varepsilon$ $\varepsilon$ $\varepsilon$ sells not every $\varepsilon$ chair \\
\\
not every sad baker $\varepsilon$ $\varepsilon$ fairly admits not every odd idea \\
\textbf{entailment} \\
$\varepsilon$ some $\varepsilon$ baker does not $\varepsilon$ admits $\varepsilon$ no $\varepsilon$ idea 
\end{tabular}    
 \caption{MQNLI examples. The $\varepsilon$ token serves as padding (but still attended to by the model) and ensures a perfect alignment between both premises and hypotheses and across all examples. It is semantically an identity element.}
  \label{tab:mqnli-examples}
  \end{subfigure}
  \hfill
  \begin{subfigure}{0.45\textwidth}
  \centering
  \begin{tabular}{l c c c}
    \toprule
    Model &  Train & Dev & Test\\
    \midrule
    CBoW       & 88.04 & 54.18 & 53.99 \\
    TreeNN     & 67.01 & 54.01 & 53.73 \\
    CompTreeNN & 99.65 & 80.17 & 80.21 \\[0.25ex]
    BiLSTM     & 99.42 & 46.41 & 46.32 \\
    BERT       & 99.99 & \textbf{88.25} & \textbf{88.50} \\
    \bottomrule
  \end{tabular}
  \caption{MQNLI results. The first three models are from \citealt{Geiger-etal:2019}, where the CompTreeNN is a task-specific model not suitable for general NLI and functions as an idealized upperbound. Our results show that BERT-based models can surpass this without such alignments.}
  \label{tab:accuracy}
  \end{subfigure}
    \caption{The natural logic causal model (top), MQNLI examples (left) and MQNLI results (right).}
\end{figure}

\paragraph{Models}

We evaluated two models on \MQNLI: a randomly initialized multilayered Bidirectional LSTM (BiLSTM; \cite{schuster1997bidirectional}) and a BERT-based classifier model in which the English \texttt{bert-base} parameters \citep{devlin-etal-2019-bert} are fine-tuned on the MQNLI train set.  Output predictions are computed using the final representation above the [CLS] token. Models are trained to predict the relation of every pair of aligned phrases in \figref{fig:bigtree}. Additional model and training details are given in  \appref{app:trainingdetails}.



\paragraph{Results}

\Figref{tab:accuracy} summarizes the results of our BERT and BiLSTM models on the hardest fair generalization task \citet{Geiger-etal:2019} creates with \MQNLI. We find that our BiLSTM model is not able to learn this task, and that our BERT model is able to achieve high accuracy. The only models in \citet{Geiger-etal:2019} able to achieve above 50\% accuracy were task-specific tree-structured models with the structure of the tree in \figref{fig:bigtree}. Thus, our BERT-based model is the first general-purpose model able to achieve good performance on this hard generalization task. Without pretraining, the BERT-based model achieves $\approx$49.1\%, confirming that pretraining is essential, as expected.

A natural hypothesis is that the BERT-based model achieves this high performance \emph{because} it has in effect induced some approximation to the tree-like structure of the data-generating process in its own internal layers. With causal abstraction analysis, we are actually in a position to test this hypothesis.

\section{A Case Study in Structural Neural Network Analysis}\label{sec:experiments}

\subsection{Causal Abstractions of Neural NLI models}

\paragraph{Formulating Our Hypotheses}\label{sec:abstractionexperiments}
We proceed just as we did for the simple motivating example in \secref{sec:motivating-example}, except that we are now seeking to assess the extent to which the natural logic algebra in \figref{fig:bigtree} is a causal abstraction of the trained neural models in the above section.

The hallmark of \figref{fig:bigtree} is that it defines an alignment between premise and hypothesis at both lexical and phrasal levels. This permits us to run interchange interventions in a naturally compositional way. For a given non-leaf node ${\nodevar}$ in \figref{fig:bigtree}, let $\natmod^{\nodevar}$ be a submodel of $\natmod$ that computes the relation between the aligned phrases under ${\nodevar}$ and uses them to compute the final output relation between premise and hypothesis. For example, let $\natmod^{\text{NP}_{\text{Obj}}}$ be the submodel of $\natmod$ that computes the relation between the two aligned object noun phrases and then uses that relation in computing the final output relation between premise and hypothesis (see \figref{fig:mainalignment} right). We would like to ask whether our trained neural models also compute this relation between object noun phrases and use it to make a final prediction. We can pose this same question for other nodes which correspond to a pair of aligned subphrases.


\paragraph{Alignment Search}\label{sec:locations} For each ${\nodevar}$, we search for an alignment between a neural representation in $\nlinet$ and the variable ${\nodevar}$ in $\natmod^{\nodevar}$. In principle, any location in the network could be the right one for any causal model. Testing every hypothesis in this space would be intractable. Thus, for each $\natmod^{\nodevar}$, we consider a restricted set of hidden representations based on the identity of ${\nodevar}$. The BERT model we use has 12 Transformer layers \citep{Vaswani-etal:2017}, meaning that there are 12 hidden representations for each input token. Each alignment search considers aligning the intermediate high-level variable with dozens of possible locations in the grid of BERT representations. Specifically, the following locations were considered for each ${\nodevar}$:
{
\vspace{-10pt}
\setlength\columnsep{30pt}
\begin{multicols}{2}
\begin{itemize}[ labelsep=1em, leftmargin=*]
\item $\QSubj$, $\AdjSubj$, $\NSubj$, $\Neg$, $\Adv$, $\V$, $\QObj$, $\AdjObj$, $\NObj$: hidden representations above the two descendant leaf tokens.
\item $\NPSubj, \VP$, and $\NPObj$: same but above the four descendant leaf tokens.
\item $\QPObj$: hidden representations above $\QObj^P$ and $\QObj^H$.
\item $\NegP$: same but above $\Neg^P$ and $\Neg^H$.
\item All nodes (for BERT): same but above [CLS] and [SEP].
\end{itemize}
\end{multicols}
\vspace{-10pt}
}
 For each alignment considered, we performed a full causal abstraction analysis. We report the results from the best alignments in \tabref{tab:all-results}, and we summarize the results from all alignments in \appref{app:heatmaps}. 
 
{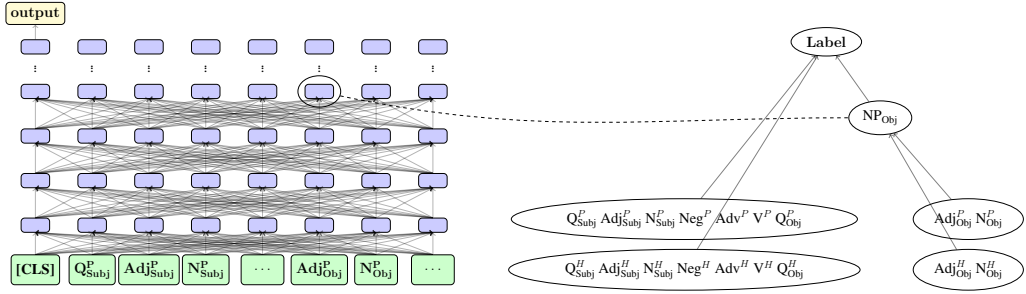
\begin{figure}[tp]
    \centering
    \hspace{-68pt}
\resizebox{!}{0.275\textwidth}{
\usetikzlibrary{shapes}
\begin{tikzpicture}[thick,scale=0.6, every node/.style={scale=0.6}]
\def\scalex{2.8}
\def\scaley{2.2}
\def\opac{0.2}
\huge

\node[fill=yellow!20, draw, rectangle, rounded corners=3pt, minimum width=60pt, minimum height=20pt](out) at (0 * \scalex,5.75 * \scaley) {\huge $\mathbf{output}$};
\node[fill=green!20, draw, rectangle, rounded corners=3pt, minimum width=60pt, minimum height=42pt](BERT00) at (0 * \scalex,0 * \scaley) {\huge\textbf{[CLS]}};
\node[fill=green!20, draw, rectangle, rounded corners=3pt, minimum width=60pt, minimum height=42pt](BERT01) at (1 * \scalex,0 * \scaley) {\huge$\mathbf{Q^P_{Subj}}$};
\node[fill=green!20, draw, rectangle, rounded corners=3pt, minimum width=60pt, minimum height=42pt](BERT02) at (2 * \scalex,0 * \scaley) {\huge$\mathbf{Adj^P_{Subj}}$};
\node[fill=green!20, draw, rectangle, rounded corners=3pt, minimum width=60pt, minimum height=42pt](BERT03) at (3 * \scalex,0 * \scaley) {\huge$\mathbf{N^P_{Subj}}$};
\node[fill=green!20, draw, rectangle, rounded corners=3pt, minimum width=60pt, minimum height=42pt](BERT04) at (4 * \scalex,0 * \scaley) {\huge$\mathbf{\dots}$};
\node[fill=green!20, draw, rectangle, rounded corners=3pt, minimum width=60pt, minimum height=42pt](BERT05) at (5 * \scalex,0 * \scaley) {\huge$\mathbf{Adj^P_{Obj}}$};
\node[fill=green!20, draw, rectangle, rounded corners=3pt, minimum width=60pt, minimum height=42pt](BERT06) at (6 * \scalex,0 * \scaley) {\huge$\mathbf{N^P_{Obj}}$};
\node[fill=green!20, draw, rectangle, rounded corners=3pt, minimum width=60pt, minimum height=42pt](BERT07) at (7 * \scalex,0 * \scaley) {\huge$\mathbf{\dots}$};

\foreach \row in { 1,2,3}
\foreach \col in {0,1,2,3,4,5,6,7}{
\node[fill=blue!20, draw, rectangle, rounded corners=3pt, minimum width=40pt, minimum height=20pt] (BERT\row\col) at (\col * \scalex,\row * \scaley) {};

}

\foreach \row in {4}
\foreach \col in {0,1,2,3,4,5,6,7}{
\node[fill=blue!20, draw, rectangle, rounded corners=3pt, minimum width=40pt, minimum height=20pt] (BERT\row\col) at (\col * \scalex,\row * \scaley) {};

}

\foreach \row in {5}
\foreach \col in {0,1,2,3,4,5,6,7}{
\node[fill=blue!20, draw, rectangle, rounded corners=3pt, minimum width=40pt, minimum height=20pt] (BERT\row\col) at (\col * \scalex,\row * \scaley) {};

}


\foreach \row in {4.58}
\foreach \col in {0,1,2,3,4,5,6,7}{
\node (BERT\row\col) at (\col * \scalex,\row * \scaley-0.05) {\huge\vdots};
}

\foreach \col in {0,1,2,3,4,5,6,7}{
\foreach \coll in {0,1,2,3,4,5,6,7}{
\draw[opacity=0.3, ->] (BERT0\col.north) to (BERT1\coll.south);
\draw[opacity=0.3, ->] (BERT1\col.north) to (BERT2\coll.south);
\draw[opacity=0.3, ->] (BERT2\col.north) to (BERT3\coll.south);
\draw[opacity=0.3, ->] (BERT3\col.north) to (BERT4\coll.south);
}
}

\draw[opacity=0.3, ->] (BERT50) to (out);

\def\startx{32}
\def\starty{0}
\def\scaley{5}
\def\scalex{7}
\def\opac{0.5}

\huge

\node[ draw, ellipse, minimum size=30pt] (P1) at (0+ \startx,0.5*\scaley+ \starty) {$\QSubj^{P}$ $\AdjSubj^{P}$ $\NSubj^{P}$ $\Neg^{P}$ $\Adv^P$ $\V^P$ $\QObj^P$ };
\node[ draw, ellipse, minimum size=30pt] (H1) at (0+ \startx,0*\scaley+ \starty) {$\QSubj^H$ $\AdjSubj^H$ $\NSubj^H$ $\Neg^H$ $\Adv^H$ $\V^H$ $\QObj^H$ };

\node[draw, ellipse, minimum size=30pt] (P2) at (2*\scalex+ \startx,0.5*\scaley+ \starty) {$\AdjObj^P$ $\NObj^P$};
\node[draw, ellipse, minimum size=30pt] (H2) at (2*\scalex+ \startx,0+ \starty) {$\AdjObj^H$ $\NObj^H$};
        
        \node[ draw, ellipse, minimum size=30pt] (NP) at (1.38*\scalex + \startx,1.5*\scaley+ \starty) {$\NPObj$};        

\node[ draw, ellipse, minimum size=30pt] (O) at (1*\scalex + \startx,2.25*\scaley+ \starty) {$\mathbf{Label}$};

\draw [->,opacity=\opac] (P1) -- (O);
\draw [->,opacity=\opac] (H1) -- (O);
\draw [->,opacity=\opac] (P2) -- (NP);
\draw [->,opacity=\opac] (H2) -- (NP);
\draw [->,opacity=\opac] (NP) -- (O);

\def\scalex{2.8}
\def\scaley{2.2}
\def\opac{0.2}

\node[draw, ellipse,  minimum width=60pt, minimum height=44pt] (bigLHS) at (5 * \scalex,4 * \scaley) {};

\draw[dashed] (bigLHS) to[out=-12, in=-180] (NP);
\end{tikzpicture}
}
    \caption{A BERT-based NLI model (left) aligned with the natural logic causal model $\natmod^{\NPObj}$ (right), where the fourth vector representation above the $\AdjObj^{P}$ token in the network is aligned with $\NPObj$, the variable representing the relation between the object noun phrases. When analyzing a sample of 1000 examples, we found a subset of 383 where $\natmod^{\NPObj}$ is an abstraction of $\nlinet$ under this alignment.}
    \label{fig:mainalignment}
\end{figure}

\paragraph{Interchange Interventions}\label{sec:nliinterchange}
We first focus on our high-level causal models. Consider a non-leaf node ${\nodevar}$ from \figref{fig:bigtree} and two input token sequences $e$ and $e'$ from MQNLI. Define
\begin{equation}
\natmod^{{\nodevar} \leftarrow e'}(e)
\end{equation}
to be the output provided by the causal model $\natmod^{\nodevar}$ when processing input $e$ where the relation between the aligned subphrases under the node ${\nodevar}$ is changed to the relation between those subphrases in $e'$. For example, simplifying for the sake of exposition, suppose $e$ is (\emph{some happy baker}, \emph{no} $\epsilon$ \emph{baker}), which has output label \textbf{contradiction}, and suppose $e'$ is (\emph{every happy person}, \emph{some happy baker}), which has output label \textbf{entailment}. We wish to intervene on the noun phrase, so ${\nodevar} = \text{NP}$. In $e$, the noun phrase relation is entailment; in $e'$, it is reverse entailment. Thus, $\natmod^{{\text{NP}} \leftarrow e'}(e)$ changes the object noun phrase relation in $e$ to entailment while holding everything else about $e$ constant. This results in the output label for the example (\emph{some happy person}, \emph{no} $\epsilon$ \emph{baker}), which is \textbf{neutral}.


Next, we consider interventions in a neural model $\nlinet$. Define
\begin{equation}
\nlinet^{L \leftarrow e'}(e)
\end{equation}
to be the output provided by $\nlinet$ processing the input $e$ when the representation at location $L$ is replaced with the representation at location $L$ created when $\nlinet$ is processing $e'$. This is exactly the process depicted in \figref{fig:addmod-example}, except now the networks are the complex trained networks of \secref{sec:datasetandmodels}.

Our hypothesis linking \figref{fig:bigtree} with a model  $\nlinet$ takes the same form as \eqref{eq:add-linking}. The causal model $\natmod^{\nodevar}$ is a constructive abstraction of $\nlinet$ when, for some representation location $L$, it is the case that, for all MQNLI examples $e$ and $e'$, we have
\begin{equation}
    \natmod^{{\nodevar} \leftarrow e'}(e) = \nlinet^{L \leftarrow e'}(e) \label{eqn:abstraction}
\end{equation}
This asserts a correspondence between interventions on the representations at $L$ in network $\nlinet$ and interventions on the variable ${\nodevar}$ in the causal model $\natmod^{\nodevar}$. If it holds, then $\nlinet$ computes the relation between the aligned phrases under the node ${\nodevar}$ and uses this information to compute the relation between the premise and hypothesis.

We call a pair of examples $(e, e')$ \emph{successful} if it satisfies equation \eqref{eqn:abstraction}, i.e., interventions in both the target causal model and neural model produce equal results. In addition, to isolate the causal impact of our interventions, we specifically focus on pairs $(e, e')$  for which performing the intervention produces a different output value than without the intervention. We call a pair $(e, e')$ \emph{impactful} if:
\begin{equation}
\natmod^{{\nodevar} \leftarrow e'}(e) \ne \natmod^{\nodevar}(e) \label{eq:impactful}
\end{equation}

\begin{table}[t]
\caption{Largest subsets of examples on which specific models $\natmod^N$ are abstractions of an LSTM and BERT model trained on \MQNLI. We record the size of such subsets as a percentage of the total 1000 examples. On this subset, we know that the neural models compute a representation of the relation between the aligned subphrases under $N$ and use this information to make a final prediction.\\[-1.25ex] }
\label{tab:all-results}
  \centering
  \footnotesize
  \begin{subtable}[t]{0.35\textwidth}
  \centering  
      \begin{tabular}[b]{l c c}
        \toprule  
        Causal Model &  LSTM & BERT\\
        \midrule
        $\QSubj$    & 0.7 & 13.1 \\
        $\QObj$    & 0.9 & 7.3 \\
        $\Neg$    & 0.7 & 21.4 \\
        $\AdjSubj$    & 2.5 & 6.7 \\
        $\NSubj$    & 1.2 & 5.5 \\
        $\AdjObj$    & 0.9 & 14.1 \\
        $\NObj$    & 0.7  & 8.8 \\
        $\V$    & 0.4 & 11.4 \\
        $\Adv$    & 1.4 & 7.9 \\
        $\NPSubj$  & 1.0 & 6.7 \\
        $\NPObj$  & 0.7 & \textbf{38.3} \\
        $\VP$  & 0.4 & 11.4 \\
        $\NegP$  & 0.9 & 11.8 \\
        \bottomrule
      \end{tabular}%
    \caption{Main results (clique sizes) for non-leaf nodes of the tree in \figref{fig:bigtree}. The hypothesis we have most evidence for is that the BERT model computes a representation of the $\NPObj$ node with the alignment shown in \figref{fig:mainalignment}. Remarkably, with 1000 examples sampled, we found a subset of 383 examples where $\natmod^{\NPObj}$ is an abstraction of BERT. }
    \label{tab:main-results}
  \end{subtable}
  \hfill
  \begin{subtable}[t]{0.57\textwidth}    
      \begin{tabular}[b]{l c}
        \toprule        
        Nodes removed &   BERT\\
        \midrule
        $\NObj^{H}$ & 31.9\\
        $\AObj^{H}$ & 15.7\\
        $\NObj^{P}$ & 33.8\\
        $\AObj^{P}$& 15.8\\
        $\NObj^{H}, \AObj^{H}$ & 31.9 \\
        $\NObj^{H}, \NObj^{P}$ & 14.1\\
        $\NObj^{H}, \AObj^{P}$  & 32.2 \\
        $\NObj^{P}, \AObj^{H}$ & 31.6\\
        $\AObj^{H}, \AObj^{P}$ & 8.8\\
        $\NObj^{P}, \AObj^{P}$ & 32.1\\
        \bottomrule
      \end{tabular}
    \hfill    
      \begin{tabular}[b]{l c}
        \toprule
         Nodes added  &   BERT\\
        \midrule
        $\AdjSubj^{P}$ & 30.5 \\
        $\NSubj^{P}$ & 37.2\\
        $\Neg^P$ & 14.9\\
        $\Adv^P$ & 26.9\\
        $\V^P$ & 35.6\\
        $\QObj^{H}$ & 16.2 \\
        $\AdjSubj^{H}$ & 13.4 \\
        $\NSubj^{H}$ & 12.0 \\
        $\Neg^H$ &  34.4\\
        $\Adv^H$ & 16.2\\
        $\V^H$ & 13.4\\
        $\QObj^H$ & 12.0 \\
        \bottomrule
      \end{tabular}
    \caption{  Detailed results (clique sizes) for Alternative causal models in a ``neighborhood'' around the model $\natmod^{\NPObj}$, which has a single intermediate variable composed of four lexical items (See \figref{fig:mainalignment}). At left, we have alternative causal models where one or two of those lexical items are removed from the composition. At right, we have alternatives obtained by adding one lexical item to the composition. We observe that no alternative hypothesis about causal structure considered has more evidence. }
    \label{tab:results:npobj}
  \end{subtable}  
\end{table}    

}

\paragraph{Quantifying Partial Success}

Equation~\eqref{eqn:abstraction} universally quantifies over all examples. We do not expect this kind of perfect correspondence to emerge in practice for real problems: neural network training is often approximate and variable in nature, and even our best model does not achieve \textit{perfect} performance. However, we can still ask how widely \eqref{eqn:abstraction} holds for a given model. To do this, we seek to find the \textit{largest subset} of MQNLI on which $\natmod^{\nodevar}$ is an abstraction of our neural models, for each non-leaf node ${\nodevar}$ in $\natmod$.


More specifically, considering each example in MQNLI as a vertex in a graph, we add an undirected edge between two examples $e_i$ and $e_j$ if and only if both the ordered pairs $(e_i, e_j)$ and $(e_j, e_i)$ satisfy \eqref{eqn:abstraction}. In other words, $\natmod^{\nodevar}$ is an abstraction of a neural model on a subset of examples $S$ of MQNLI if and only if all examples in $S$ form a \textit{clique}.

The number of interventions we need to run scales quadratically with the number of inputs we consider, so we sample $1000$ \MQNLI\ examples, producing a total of $1000^{2} = $ 1M ordered pairs. We only consider examples for which the neural network outputs a correct label. For each node ${\nodevar}$ and each of its corresponding neural network locations $L$, we perform interventions on all of these pairs.

We choose to measure the largest clique with at least one impactful edge, because (1) the causal abstraction relation holds with full force on that clique, but other measures such as the total number of connections lack this theoretical grounding, and (2) if a clique has at least one impactful edge, that guarantees the high-level variable is being used.

\paragraph{Results and Analysis}

For each target causal model node ${\nodevar}$ and neural network representation location $L$, we construct a graph as described above with 1000 examples as vertices and add an edge between two examples $e_i$ and $e_j$ if and only if \textit{both} $(e_i, e_j)$ and $(e_j, e_i)$ are successful. We then find the largest clique in this graph with at least one impactful edge and record its size.

\tabref{tab:main-results} shows, for each causal model node ${\nodevar}$, the maximum size of cliques found among all neural locations. With this stricter \emph{impactful} criterion (as opposed to simply using intervention success), our results show that, for almost all nodes ${\nodevar}$, our target causal model $\natmod^{\nodevar}$ is indeed a causal abstraction of BERT on a significant number of examples in our dataset. These subsets are much smaller for the BiLSTM model.

We also investigated alternative high-level causal structures that are not variants of $\natmod$ from \figref{fig:bigtree}. Specifically, we consider alternative models in a ``neighborhood'' around the model $\natmod^{\NPObj}$ that can be obtained by adding one leaf, or by removing one or two leaves to the composition. These results are in \tabref{tab:results:npobj}. Remarkably, all of these alternative models result in smaller clique sizes, significantly so for many of them. This further supports the significance of our results.


This analysis is similar to the analysis of our hypothetical addition example in \secref{sec:causality}, except for two crucial differences. First, for each variable ${\nodevar}$, we are hypothesizing that the causal model $\natmod^{\nodevar}$ is an abstraction of $\nlinet$, whereas in the addition example there was only one model. To investigate this difference, we take ${\nodevar} = \NPObj$ as a paradigm case, as it is the model with the strongest results. (The results for other nodes are in \appref{app:heatmaps}.) Second, we only achieved partial experimental success, whereas in the addition example we assumed complete success. Crucially, this means that the following analysis will be valid only on subsets of the input space on which the abstraction relation holds between $\nlinet$ and $\natmod^{\NPObj}$. 

\begin{figure*}[t]

    \makebox[\textwidth][c]{
    \begin{subfigure}[t]{0.24\textwidth}
        \centering
        \includegraphics[width=1\textwidth]{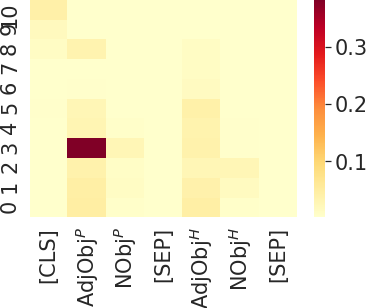}
        \caption{Clique size.}
        \label{fig:N_Subj_clq_size}
    \end{subfigure}%
    \hfill
    \begin{subfigure}[t]{0.24\textwidth}
        \centering
        \includegraphics[width=1\textwidth]{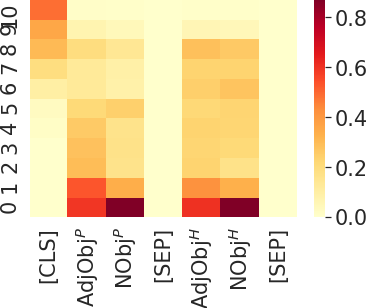}
        \caption{Interchange success.}
        \label{fig:N_Subj_interx}
    \end{subfigure}
    \hfill
    \begin{subfigure}[t]{0.24\textwidth}
        \centering
        \includegraphics[width=1\textwidth]{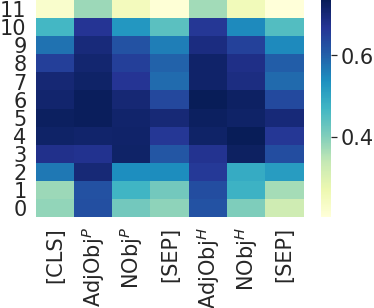}
        \caption{Probe selectivity.}
        \label{fig:N_Subj_probing_sel}
    \end{subfigure}%
    \hfill
    \begin{subfigure}[t]{0.24\textwidth}
        \centering
        \includegraphics[width=1\textwidth]{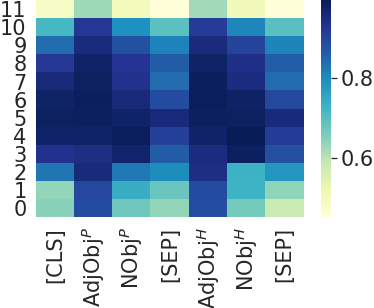}
        \caption{Probe Accuracy. }
        \label{fig:N_Subj_probing_acc}
    \end{subfigure}%
    }
    

    \caption{Interchange intervention and probing results for the $\text{NP}_\text{Obj}$ position. Vertical axes denote layers of BERT and horizontal axes denote the token position of hidden representations. The intervention success rates reported here are calculated based on intervention experiments with a change in the output label. Clique sizes are reported as \% of 1000 examples. 
    }
    \label{fig:probing-results}
\end{figure*}

We visualize the results of our intervention experiments for the node $\NPObj$ in \figref{fig:probing-results}. The alignment with the largest subset of inputs aligns the $\NPObj$ variable in $\natmod^{\NPObj}$ with the neural representation on the fourth layer of BERT above the $\AdjObj^P$ token (see \figref{fig:mainalignment}). Because neural representations encode the value of their aligned variables and play a parallel causal role to their high-level variables, we know that, on this subset of input examples, at the fourth neural representation above the $\AdjObj^P$ token, the four input embeddings for the object nouns and adjectives in the premise and hypothesis are composed to form a neural representation with information content of the relation between the object noun phrases in the premise and hypothesis.  Then this representation is composed with the other input-embeddings to create an output representing the relation between the premise and hypothesis. 
%




\subsection{Comparison with Other Structural Analysis Methods}\label{sec:other}

\paragraph{Probes}

We probed neural representation locations for the relation between aligned subexpressions on a subset of 12,800 randomly selected \MQNLI\ examples. For a pair of aligned subexpressions below a node ${\nodevar}$ in \figref{fig:bigtree}, we probe the columns above the same set of restricted class of tokens as described in \secref{sec:locations}. 

To evaluate these probes, we report accuracy as well as \emph{selectivity} as defined by \citet{hewitt-liang-2019-designing}: probe accuracy minus control accuracy, where \emph{control accuracy} is the train set accuracy of a probe with the same architecture but trained on a control task to factor out probe success that can be attributed to the probe model itself. Our control task is to learn a random mapping from node types to semantic relations; see \appref{app:probing} for full details on how this task was constructed.

\Figref{fig:probing-results} summarizes our probing results for ${\nodevar}=\NPObj$, along with corresponding interchange intervention results for comparison. Probes tell us that information about the relation between the aligned noun phrases is encoded in nearly all of the locations we considered, and using the selectivity metric does not result in any qualitative change. In contrast, our intervention heatmaps indicate only a small number of locations store this information in a causally relevant way. Clearly, our intervention experiments are far more discriminating than probes. \Appref{app:heatmaps} provides examples involving other variables along with the intervention experiments, where the general trend of interchange interventions being more discriminating holds.




\paragraph{Integrated Gradients}

Attribution methods that estimate feature importance can measure causal properties of neural representations, but a single feature importance method is an impoverished characterization of a representation's role in network behavior. Whereas our interchange interventions gave us high-level information about how a neural representation is composed and what it is composed into, attribution methods simply tell us ``how much'' a representation contributes to the network output on a give input. Moreover, intervention interchanges provide a rich, high-level characterization of causal structure on a space of inputs.

We use integrated gradients on our models to verify the intuitive hypothesis that if a premise and hypothesis differ by a single token, then the neural representations above that token should be more causally responsible for the network output than other representations. For example, given premise `Every sleepy cat meows' and hypothesis `Some hungry cat meows', the attributive modifier position is different and the rest are matched. The neural representations above the adjective tokens \textit{sleepy} and \textit{hungry} should be more important for the network output than others, because if those adjectives were the same, the example label would change from \textbf{neutral} to \textbf{entailment}.
We summarize the results of our integrated gradient experiments in \appref{app:IG}, where we confirm our intuitive hypothesis.

\section{Conclusion}\label{sec:conclusion}

We have introduced a methodology for deriving interpretable causal explanations of neural network behaviors, grounded in a formal theory of causal abstraction. The methodology involves first \emph{formulating a hypothesis} in the form of a high-level, interpretable causal model, then \emph{searching for an alignment} between the neural network and the causal model, and finally \emph{verifying experimentally} that the neural representations encode the same causal properties and information content as the corresponding components of the high-level causal model. As a case study demonstrating the feasibility of the approach, we analyzed neural models trained on the semantically formidable MQNLI dataset. Guided by the intuition that success on this challenging task may call for a way of recapitulating the causal structure of the natural logic model that generates the MQNLI data, we were able to verify the hypothesis that a state-of-the-art BERT-based model partially realizes this structure, whereas baseline models that do not perform as well fail to do so. This suggestive case study demonstrates that our theoretically grounded methodology can work in practice. 





\ack{Our thanks to Amir Feder, Noah Goodman, Elisa Kreiss, Josh Rozner, Zhengxuan Wu, and our anonymous reviewers. This research is supported in part by grants from Facebook and Google.}

\bibliography{anthology,custom}

\begin{thebibliography}{32}
\providecommand{\natexlab}[1]{#1}
\providecommand{\url}[1]{\texttt{#1}}
\expandafter\ifx\csname urlstyle\endcsname\relax
  \providecommand{\doi}[1]{doi: #1}\else
  \providecommand{\doi}{doi: \begingroup \urlstyle{rm}\Url}\fi

\bibitem[Beckers and Halpern(2019)]{Beckers_Halpern_2019}
S.~Beckers and J.~Y. Halpern.
\newblock Abstracting causal models.
\newblock \emph{Proceedings of the AAAI Conference on Artificial Intelligence},
  33\penalty0 (01):\penalty0 2678--2685, Jul. 2019.
\newblock \doi{10.1609/aaai.v33i01.33012678}.
\newblock URL \url{https://ojs.aaai.org/index.php/AAAI/article/view/4117}.

\bibitem[Beckers et~al.(2020)Beckers, Eberhardt, and Halpern]{beckers20a}
S.~Beckers, F.~Eberhardt, and J.~Y. Halpern.
\newblock Approximate causal abstractions.
\newblock In R.~P. Adams and V.~Gogate, editors, \emph{Proceedings of The 35th
  Uncertainty in Artificial Intelligence Conference}, volume 115 of
  \emph{Proceedings of Machine Learning Research}, pages 606--615, Tel Aviv,
  Israel, 22--25 Jul 2020. PMLR.
\newblock URL \url{http://proceedings.mlr.press/v115/beckers20a.html}.

\bibitem[Binder et~al.(2016)Binder, Montavon, Bach, M{\"{u}}ller, and
  Samek]{Binder16}
A.~Binder, G.~Montavon, S.~Bach, K.~M{\"{u}}ller, and W.~Samek.
\newblock Layer-wise relevance propagation for neural networks with local
  renormalization layers.
\newblock \emph{CoRR}, abs/1604.00825, 2016.
\newblock URL \url{http://arxiv.org/abs/1604.00825}.

\bibitem[Bongers et~al.(2020)Bongers, Forr{\'e}, Peters, Sch{\"o}lkopf, and
  Mooij]{Bongers2016}
S.~Bongers, P.~Forr{\'e}, J.~Peters, B.~Sch{\"o}lkopf, and J.~M. Mooij.
\newblock Foundations of structural causal models with cycles and latent
  variables.
\newblock \emph{arXiv.org preprint}, arXiv:1611.06221v4 [stat.ME], Oct. 2020.
\newblock URL \url{https://arxiv.org/abs/1611.06221v4}.

\bibitem[Chalupka et~al.(2016)Chalupka, Eberhardt, and Perona]{chalupka16}
K.~Chalupka, F.~Eberhardt, and P.~Perona.
\newblock Multi-level cause-effect systems.
\newblock In A.~Gretton and C.~C. Robert, editors, \emph{Proceedings of the
  19th International Conference on Artificial Intelligence and Statistics},
  volume~51 of \emph{Proceedings of Machine Learning Research}, pages 361--369,
  Cadiz, Spain, 09--11 May 2016. PMLR.
\newblock URL \url{http://proceedings.mlr.press/v51/chalupka16.html}.

\bibitem[Chattopadhyay et~al.(2019)Chattopadhyay, Manupriya, Sarkar, and
  Balasubramanian]{chattopadhyay19a}
A.~Chattopadhyay, P.~Manupriya, A.~Sarkar, and V.~N. Balasubramanian.
\newblock Neural network attributions: A causal perspective.
\newblock In K.~Chaudhuri and R.~Salakhutdinov, editors, \emph{Proceedings of
  the 36th International Conference on Machine Learning}, volume~97 of
  \emph{Proceedings of Machine Learning Research}, pages 981--990, Long Beach,
  California, USA, 09--15 Jun 2019. PMLR.
\newblock URL \url{http://proceedings.mlr.press/v97/chattopadhyay19a.html}.

\bibitem[Clark et~al.(2019)Clark, Khandelwal, Levy, and
  Manning]{clark-etal-2019-bert}
K.~Clark, U.~Khandelwal, O.~Levy, and C.~D. Manning.
\newblock What does {BERT} look at? an analysis of {BERT}{'}s attention.
\newblock In \emph{Proceedings of the 2019 ACL Workshop BlackboxNLP: Analyzing
  and Interpreting Neural Networks for NLP}, pages 276--286, Florence, Italy,
  Aug. 2019. Association for Computational Linguistics.
\newblock \doi{10.18653/v1/W19-4828}.
\newblock URL \url{https://www.aclweb.org/anthology/W19-4828}.

\bibitem[Devlin et~al.(2019)Devlin, Chang, Lee, and
  Toutanova]{devlin-etal-2019-bert}
J.~Devlin, M.-W. Chang, K.~Lee, and K.~Toutanova.
\newblock {BERT}: Pre-training of deep bidirectional transformers for language
  understanding.
\newblock In \emph{Proceedings of the 2019 Conference of the North {A}merican
  Chapter of the Association for Computational Linguistics: Human Language
  Technologies, Volume 1 (Long and Short Papers)}, pages 4171--4186,
  Minneapolis, Minnesota, June 2019. Association for Computational Linguistics.
\newblock \doi{10.18653/v1/N19-1423}.
\newblock URL \url{https://www.aclweb.org/anthology/N19-1423}.

\bibitem[Elazar et~al.(2020)Elazar, Ravfogel, Jacovi, and
  Goldberg]{elazar-etal-2020}
Y.~Elazar, S.~Ravfogel, A.~Jacovi, and Y.~Goldberg.
\newblock Amnesic probing: Behavioral explanation with amnesic counterfactuals.
\newblock In \emph{Proceedings of the 2020 {EMNLP} Workshop {B}lackbox{NLP}:
  Analyzing and Interpreting Neural Networks for {NLP}}. Association for
  Computational Linguistics, Nov. 2020.
\newblock \doi{10.18653/v1/W18-5426}.

\bibitem[Geiger et~al.(2019)Geiger, Cases, Karttunen, and
  Potts]{Geiger-etal:2019}
A.~Geiger, I.~Cases, L.~Karttunen, and C.~Potts.
\newblock Posing fair generalization tasks for natural language inference.
\newblock In \emph{Proceedings of the 2019 Conference on Empirical Methods in
  Natural Language Processing and the 9th International Joint Conference on
  Natural Language Processing (EMNLP-IJCNLP)}, pages 4475--4485, Stroudsburg,
  PA, November 2019. Association for Computational Linguistics.
\newblock \doi{10.18653/v1/D19-1456}.
\newblock URL \url{https://www.aclweb.org/anthology/D19-1456}.

\bibitem[Geiger et~al.(2020)Geiger, Richardson, and
  Potts]{geiger-etal-2020-neural}
A.~Geiger, K.~Richardson, and C.~Potts.
\newblock Neural natural language inference models partially embed theories of
  lexical entailment and negation.
\newblock In \emph{Proceedings of the Third BlackboxNLP Workshop on Analyzing
  and Interpreting Neural Networks for NLP}, pages 163--173, Online, Nov. 2020.
  Association for Computational Linguistics.
\newblock \doi{10.18653/v1/2020.blackboxnlp-1.16}.
\newblock URL \url{https://www.aclweb.org/anthology/2020.blackboxnlp-1.16}.

\bibitem[Hewitt and Liang(2019)]{hewitt-liang-2019-designing}
J.~Hewitt and P.~Liang.
\newblock Designing and interpreting probes with control tasks.
\newblock In \emph{Proceedings of the 2019 Conference on Empirical Methods in
  Natural Language Processing and the 9th International Joint Conference on
  Natural Language Processing (EMNLP-IJCNLP)}, pages 2733--2743, Hong Kong,
  China, Nov. 2019. Association for Computational Linguistics.
\newblock \doi{10.18653/v1/D19-1275}.
\newblock URL \url{https://www.aclweb.org/anthology/D19-1275}.

\bibitem[Hupkes et~al.(2018)Hupkes, Bouwmeester, and
  Fern{\'a}ndez]{hupkes-etal-2018-analysing}
D.~Hupkes, S.~Bouwmeester, and R.~Fern{\'a}ndez.
\newblock Analysing the potential of seq-to-seq models for incremental
  interpretation in task-oriented dialogue.
\newblock In \emph{Proceedings of the 2018 {EMNLP} Workshop {B}lackbox{NLP}:
  Analyzing and Interpreting Neural Networks for {NLP}}, pages 165--174,
  Brussels, Belgium, Nov. 2018. Association for Computational Linguistics.
\newblock \doi{10.18653/v1/W18-5419}.
\newblock URL \url{https://www.aclweb.org/anthology/W18-5419}.

\bibitem[Icard and Moss(2013)]{Icard:Moss:2013:LILT}
T.~F. Icard and L.~S. Moss.
\newblock Recent progress on monotonicity.
\newblock \emph{Linguistic Issues in Language Technology}, 9\penalty0
  (7):\penalty0 1--31, January 2013.

\bibitem[Imbens and Rubin(2015)]{Imbens2015}
G.~W. Imbens and D.~B. Rubin.
\newblock \emph{Causal inference in statistics, social, and biomedical
  sciences}.
\newblock Cambridge University Press, 2015.

\bibitem[Lillicrap and Kording(2019)]{lillicrap2019does}
T.~P. Lillicrap and K.~P. Kording.
\newblock What does it mean to understand a neural network?, 2019.

\bibitem[MacCartney and Manning(2007)]{MacCartney:07}
B.~MacCartney and C.~D. Manning.
\newblock Natural logic for textual inference.
\newblock In \emph{Proceedings of the ACL-PASCAL Workshop on Textual Entailment
  and Paraphrasing}, RTE '07, pages 193--200, Stroudsburg, PA, USA, 2007.
  Association for Computational Linguistics.
\newblock URL \url{http://dl.acm.org/citation.cfm?id=1654536.1654575}.

\bibitem[MacCartney and Manning(2009)]{maccartney-manning-2009-extended}
B.~MacCartney and C.~D. Manning.
\newblock An extended model of natural logic.
\newblock In \emph{Proceedings of the Eight International Conference on
  Computational Semantics}, pages 140--156, Tilburg, The Netherlands, Jan.
  2009. Association for Computational Linguistics.
\newblock URL \url{https://www.aclweb.org/anthology/W09-3714}.

\bibitem[Pearl(2001)]{pearl}
J.~Pearl.
\newblock Direct and indirect effects.
\newblock In \emph{Proceedings of the Seventeenth Conference on Uncertainty in
  Artificial Intelligence}, UAI’01, page 411–420, San Francisco, CA, USA,
  2001. Morgan Kaufmann Publishers Inc.
\newblock ISBN 1558608001.

\bibitem[Peters et~al.(2018)Peters, Neumann, Zettlemoyer, and
  Yih]{peters-etal-2018-dissecting}
M.~Peters, M.~Neumann, L.~Zettlemoyer, and W.-t. Yih.
\newblock Dissecting contextual word embeddings: Architecture and
  representation.
\newblock In \emph{Proceedings of the 2018 Conference on Empirical Methods in
  Natural Language Processing}, pages 1499--1509, Brussels, Belgium, Oct.-Nov.
  2018. Association for Computational Linguistics.
\newblock \doi{10.18653/v1/D18-1179}.
\newblock URL \url{https://www.aclweb.org/anthology/D18-1179}.

\bibitem[Ravichander et~al.(2020)Ravichander, Belinkov, and
  Hovy]{ravichander2020probing}
A.~Ravichander, Y.~Belinkov, and E.~Hovy.
\newblock Probing the probing paradigm: Does probing accuracy entail task
  relevance?, 2020.

\bibitem[Rubenstein et~al.(2017)Rubenstein, Weichwald, Bongers, Mooij, Janzing,
  Grosse-Wentrup, and Sch{\"o}lkopf]{Rubensteinetal17}
P.~K. Rubenstein, S.~Weichwald, S.~Bongers, J.~M. Mooij, D.~Janzing,
  M.~Grosse-Wentrup, and B.~Sch{\"o}lkopf.
\newblock Causal consistency of structural equation models.
\newblock In \emph{Proceedings of the 33rd Conference on Uncertainty in
  Artificial Intelligence (UAI)}. Association for Uncertainty in Artificial
  Intelligence (AUAI), Aug. 2017.
\newblock URL \url{http://auai.org/uai2017/proceedings/papers/11.pdf}.
\newblock *equal contribution.

\bibitem[Schuster and Paliwal(1997)]{schuster1997bidirectional}
M.~Schuster and K.~K. Paliwal.
\newblock Bidirectional recurrent neural networks.
\newblock \emph{IEEE transactions on Signal Processing}, 45\penalty0
  (11):\penalty0 2673--2681, 1997.

\bibitem[Shrikumar et~al.(2016)Shrikumar, Greenside, Shcherbina, and
  Kundaje]{Shrikumar16}
A.~Shrikumar, P.~Greenside, A.~Shcherbina, and A.~Kundaje.
\newblock Not just a black box: Learning important features through propagating
  activation differences.
\newblock \emph{CoRR}, abs/1605.01713, 2016.
\newblock URL \url{http://arxiv.org/abs/1605.01713}.

\bibitem[Spirtes et~al.(2001)Spirtes, Glymour, and Scheines]{spirtes}
P.~Spirtes, C.~N. Glymour, and R.~Scheines.
\newblock \emph{{Causation, Prediction, and Search}}.
\newblock MIT Press, 2nd edition, 2001.
\newblock ISBN 9780262194402.

\bibitem[Springenberg et~al.(2014)Springenberg, Dosovitskiy, Brox, and
  Riedmiller]{springerberg2014}
J.~Springenberg, A.~Dosovitskiy, T.~Brox, and M.~Riedmiller.
\newblock Striving for simplicity: The all convolutional net.
\newblock \emph{CoRR}, 12 2014.

\bibitem[Sundararajan et~al.(2017)Sundararajan, Taly, and Yan]{sundararajan17a}
M.~Sundararajan, A.~Taly, and Q.~Yan.
\newblock Axiomatic attribution for deep networks.
\newblock In D.~Precup and Y.~W. Teh, editors, \emph{Proceedings of the 34th
  International Conference on Machine Learning}, volume~70 of \emph{Proceedings
  of Machine Learning Research}, pages 3319--3328, International Convention
  Centre, Sydney, Australia, 06--11 Aug 2017. PMLR.
\newblock URL \url{http://proceedings.mlr.press/v70/sundararajan17a.html}.

\bibitem[Tenney et~al.(2019)Tenney, Das, and Pavlick]{tenney-etal-2019-bert}
I.~Tenney, D.~Das, and E.~Pavlick.
\newblock {BERT} rediscovers the classical {NLP} pipeline.
\newblock In \emph{Proceedings of the 57th Annual Meeting of the Association
  for Computational Linguistics}, pages 4593--4601, Florence, Italy, July 2019.
  Association for Computational Linguistics.
\newblock \doi{10.18653/v1/P19-1452}.
\newblock URL \url{https://www.aclweb.org/anthology/P19-1452}.

\bibitem[van Benthem(2008)]{vanBenthem:08}
J.~van Benthem.
\newblock A brief history of natural logic.
\newblock In M.~Chakraborty, B.~L{\"o}we, M.~Nath~Mitra, and S.~Sarukki,
  editors, \emph{Logic, Navya-Nyaya and Applications: Homage to {B}imal
  {M}atilal}, 2008.

\bibitem[Vaswani et~al.(2017)Vaswani, Shazeer, Parmar, Uszkoreit, Jones, Gomez,
  Kaiser, and Polosukhin]{Vaswani-etal:2017}
A.~Vaswani, N.~Shazeer, N.~Parmar, J.~Uszkoreit, L.~Jones, A.~N. Gomez, L.~u.
  Kaiser, and I.~Polosukhin.
\newblock Attention is all you need.
\newblock In I.~Guyon, U.~V. Luxburg, S.~Bengio, H.~Wallach, R.~Fergus,
  S.~Vishwanathan, and R.~Garnett, editors, \emph{Advances in Neural
  Information Processing Systems 30}, pages 5998--6008. Curran Associates,
  Inc., 2017.
\newblock URL
  \url{http://papers.nips.cc/paper/7181-attention-is-all-you-need.pdf}.

\bibitem[Wolf et~al.(2019)Wolf, Debut, Sanh, Chaumond, Delangue, Moi, Cistac,
  Rault, Louf, Funtowicz, and Brew]{Wolf2019HuggingFacesTS}
T.~Wolf, L.~Debut, V.~Sanh, J.~Chaumond, C.~Delangue, A.~Moi, P.~Cistac,
  T.~Rault, R.~Louf, M.~Funtowicz, and J.~Brew.
\newblock Huggingface's transformers: State-of-the-art natural language
  processing.
\newblock \emph{ArXiv}, abs/1910.03771, 2019.

\bibitem[Zeiler and Fergus(2014)]{Zeiler2014}
M.~D. Zeiler and R.~Fergus.
\newblock Visualizing and understanding convolutional networks.
\newblock In D.~Fleet, T.~Pajdla, B.~Schiele, and T.~Tuytelaars, editors,
  \emph{Computer Vision -- ECCV 2014}, pages 818--833, Cham, 2014. Springer
  International Publishing.
\newblock ISBN 978-3-319-10590-1.

\end{thebibliography}

\newpage
\clearpage

\appendix

\section{Additional Details on MQNLI}\label{app:mqnli-data}

\subsection{Dataset Description} \label{sec:datadesc}

\newcommand{\nlvar}[2]{\ensuremath{\text{#1}_{\text{#2}}}}
\newcommand{\word}[1]{\emph{#1}}
The MQNLI dataset contains sentences of the form
\begin{center}
  \nlvar{Q}{S} \nlvar{Adj}{S} \nlvar{N}{S} \nlvar{Neg}{} \nlvar{Adv}{}
  \nlvar{V}{} \nlvar{Q}{O} \nlvar{Adj}{O} \nlvar{N}{O}
\end{center}
where \nlvar{N}{S} and \nlvar{N}{O} are nouns, \nlvar{V}{} is a verb,
\nlvar{Adj}{S} and \nlvar{Adj}{O} are adjectives, and \nlvar{Adv}{} is
an adverb. These categories all have 100 words. \nlvar{Neg}{} is \word{does not}, and \nlvar{Q}{S} and
\nlvar{Q}{O} can be \word{every}, \word{not every}, \word{some}, or
\word{no}.
Additionally, \nlvar{Adj}{S}, \nlvar{Adj}{O}, \nlvar{Adv}{}, and
\nlvar{Neg}{} can be the empty string $\varepsilon$.

NLI examples are constructed so that non-identical non-empty
nouns, adjectives, verbs, and adverbs with identical positions in
$s_{p}$ and $s_{h}$ are semantically unrelated. This means that the learning task is trivial for these lexical items, as the correct relation is equivalence when they are identical and independence when they are not identical.

For our experiments, we used a train set with 500K examples,  a dev set with 60k examples, and a test set with 10K examples -- the most difficult generalization scheme of \citet{Geiger-etal:2019}.

\subsection{A Natural Logic Causal Model}\label{app:natlogmodel}

\citet{Geiger-etal:2019} construct a natural logic model that solves \MQNLI\ using a formalization they call \emph{composition trees}, which is easily translated into the causal model we call $\natmod$.  Natural logic is a flexible approach to doing logical inference directly on natural language expressions \citep{Icard:Moss:2013:LILT,MacCartney:07,vanBenthem:08} where the \textit{semantic relations} between phrases are compositionally computed from the semantic relations between aligned subphrases and \emph{projectivity signatures}, which encode how semantic operators interact compositionally with their arguments (which are semantic relations). The causal model $\natmod$ performs inference on aligned semantic parse trees that represent both the premise and hypothesis as a single structure and calculates semantic relations between all subphrases compositionally.

\section{Model Training and Interchange Experiment Details}\label{app:trainingdetails}

We evaluated two models on \MQNLI: a multi-layered bidirectional LSTM baseline and a Transformer-based model trained to do masked language modeling and next-sentence prediction \citep{devlin-etal-2019-bert}. We rely on the uncased BERT-base initial parameters from Hugging Face \texttt{transformers} \citep{Wolf2019HuggingFacesTS}.
%
%
For both models, we concatenate the premise $s_p$ and hypothesis $s_h$ into one string with special separator tokens: [CLS] $s_p$ [SEP] $s_h$ [SEP]. 

For the BiLSTM, we concatenate the hidden state above the last [SEP] and the [CLS] in the last layer for the forward and backward directions respectively  to obtain a representation for the whole input, and then apply three linear transformations on top of that. The final transformation outputs a logit score for each class in the label space.

For the BERT model, we apply one linear transformation to the final layer's hidden representation above the [CLS] token to obtain a logit score for each label class.

\subsection{Tokenization} 

In the original setting of \MQNLI, some positions in the premise and hypothesis consist of two words such as \textit{not every} in \nlvar{Q}{S} and \nlvar{Q}{O} and \textit{does not} in the leaf nodes Neg$^\text{P}$ and Neg$^\text{H}$ (as shown in the beginning of \secref{sec:datadesc}). We treat them as two separate tokens in order to utilize BERT's knowledge of these function words. To ensure all sentences have identical length, we introduce one extra empty string tokens $\varepsilon$ to single-word quantifiers and two such tokens in the place of Neg$^P$ and Neg$^H$ for sentences without negation.

For consistency, we use the same tokenization method for both models.

\subsection{Dataset Augmentation with Labeled Subphrases} 

The \textit{hard but fair} MQNLI generalization task requires the dataset to explicitly expose the model to labels for each intermediate node that is a relation in $\natmod$. For each training example $(s_p, s_h, y) \in \mathcal{S}$, we create an additional example $(s_p^N, s_h^N, y^N)$ for each node $N$. $(s_p^N, s_h^N)$ is a \textit{subphrase} pair made up of all the leaf tokens under node $N$ in the original input $(s_p, s_h)$, and $y^N$ is the relation computed by $\natmod$ for that subphrase pair. The set of labels we use for these subphrase examples is disjoint from that of the full-sentence examples. During training, the augmented examples are coupled with original examples in each batch. For BERT, the subphrase pairs occupy their original positions in the sentence, while we pad and apply an attention mask over all other positions. For the BiLSTM, we align them to the left, with [SEP] in between the two parts of the pair. 

We performed an ablation experiment to test whether removing the augmented examples would affect BERT's performance. Using the same grid-search setting, we see that BERT's dev set accuracy decreased from 88.25\% to 55.42\%, and test set accuracy decreased from 88.50\% to 54.51\%. This indeed shows that the above data augmentation method is important for BERT to learn the type of generalization required for the hard MQNLI task.

\begin{table}[t]
  \caption{Ablation results.}
  \label{tab:ablation}
  \centering
  \begin{tabular}{l c c}
    \toprule
    Model &  Dev & Test\\
    \midrule
    Fine-tuned BERT       & \textbf{88.25} & \textbf{88.50} \\
    Without augmented examples & 55.42 & 54.51 \\
    \bottomrule
  \end{tabular}
\end{table}

\subsection{Training Procedure} 

For the BiLSTM, we use 256 dimensions for token embeddings and 128 dimensions for the hidden states in each LSTM direction. We grid search for $\{2,4,6\}$ layers. We randomly initialize each element in the token embeddings from the distribution $\mathcal{N}(0,1)$ scaled down by a factor of 0.1. We use a batch size of $768 = 64 \times 12$, with 64 original examples per batch and 11 augmented examples for each one. We apply a dropout of 0.1, and grid search for learning rates in $\{ 0.001, 0.0001\}$. We train for a maximum of 400 epochs and perform early stopping when the dev set accuracy does not increase for 20 epochs. We train each grid search setting 3 times with different random seeds.

For BERT, we use the same model architecture for the uncased base variant. We use a batch size of $192 = 16 \times 12$, and grid search for learning rates in $\{2.0\times10^{-5}, 5.0\times10^{-5} \}$. We train for a maximum of \{3, 4\} epochs. We warm up the learning rate linearly from 0 to the specified value in the first 25\% of steps of the first epoch, and linearly decrease the learning rate to $0$ following that until the end of training.

All models were trained with 1 GPU core on a cluster with models including GeForce RTX 2080 Ti, GeForce GTX Titan X, Titan XP and Titan V, each with 11-12GB memory. Each instance of the grid search took on average 5.5 hours to train. We repeated each grid search setting with 4 different random seeds and took the instance with the highest dev set accuracy.

\subsection{Interchange experiment details}

There are 14 intermediate nodes in the high-level causal model ($\NegP$, $\QPObj$, $\QSubj$,  $\NPSubj$, $\AdjSubj$, $\NSubj$, $\Neg$, $\VP$, $\Adv,$ $\text{V}$, $\QObj$, $\NPObj$, $\AdjObj$, $\NObj$). For each high-level node, we conducted a set of interchange experiments on each one of 11 BERT layers (excluding the final layer, since only the [CLS] token causally impacts the output). Each high-level node has its own fixed set of hand-specified intervention locations in the time-step/sentence length dimension, and we use the same intervention locations on each layer. For each of the $14 \times 11 = 154$ interchange experiments, it took on average 1.15 hours to run using the same computation resources mentioned above.


%

\section{Probing Details}\label{app:probing}

\subsection{Probe Models}

Our probe models are single-layer softmax classifiers: $y_i \propto \mathrm{softmax}(Ah_i + b)$ where $h_i$ is a hidden representation and $y_i \in \mathbb{R}$. Following \citet{hewitt-liang-2019-designing}, to control the dimensionality of $A$, we factorize it in the form $A = LR$ where $L \in \mathbb{R}^{|\mathcal{R}| \times \ell}$ and $R \in \mathbb{R}^{\ell \times d}$ where $d$ is the dimensionality of $h_i$. 

We train the probes on hidden representations of a set of 12,800 examples that are randomly selected from the model's original training set. We additionally take 2,000 examples to form a development set for early stopping. We filter out examples for which the model outputs a wrong prediction.

For training, we perform a grid search, maximizing for selectivity. We set a dropout of 0.1, and apply early stopping when the development set loss does not increase for 4 epochs. We train for a maximum of 40 epochs. We also anneal the learning rate by a factor of 0.5 if the dev set loss did not increase in the last epoch. We use a batch size of 512,  learning rates in $\{0.001, 0.01\}$, weight decay regularization constants in $\{0.01, 0.1\}$. We set $\ell \in \{8, 32\}$ for restricting the maximum rank of the linear matrix $A$.

Using the same computation resources described above, each grid search setting took approximately 5 hours to run. For each grid search setting we trained a separate probe for every possible $\langle\text{causal model node, BERT representation}\rangle$ combination, where for the latter we use the intervention locations outlined in the ``Alignment Search" part of \secref{sec:abstractionexperiments} on each
BERT layer.

\subsection{Control Task} \label{sec:probe_control_task}

For each high-level node $N$, we construct a random mapping $\mathrm{Control}_N: \mathcal{S}_N \mapsto \mathcal{L}_N$ where $\mathcal{S}_N$ is the set of all aligned subexpressions under the node $N$ and $\mathcal{L}_N$ is the output label space. For phrasal nodes ($\VP, \NegP$, etc.) and aligned verbs and nouns, $\mathcal{L}_N$ is the set of 7 possible relations $\{\#, \equiv, \sqsubset, \sqsupset, |,\ \hat{} \ , \smile \}$ from \citet{MacCartney:07}. For aligned quantifiers, the label space is the set of all projectivity signatures that can be produced by their composition.

Similar to \citet{hewitt-liang-2019-designing}, $\mathrm{Control}_N$ will assign the same control label regardless of the context as long as its input consists of the same tokens. Consequently, the possible input space $\mathcal{S}_N$ grows exponentially larger if $N$ corresponds to longer subphrases (such as $\NegP$ and $\QPObj$), and the control task becomes much more difficult to solve, resulting in near random accuracies.

\subsection{Extended Probe Analysis}

\begin{figure*}[!ht]
    \makebox[\textwidth][c]{
    \begin{subfigure}[t]{0.24\textwidth}
        \centering
        \includegraphics[width=0.9\textwidth]{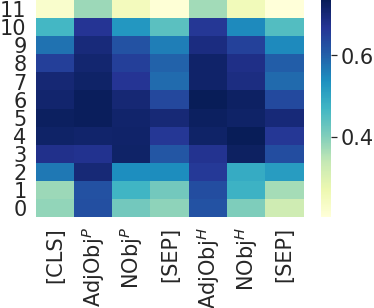}
        \caption{$\text{NP}_\text{Obj}$ Probe selectivity. 
        }
        \label{fig:NP_Obj_probing_sel}
    \end{subfigure}\hfill
    \begin{subfigure}[t]{0.24\textwidth}
        \centering
        \includegraphics[width=0.9\textwidth]{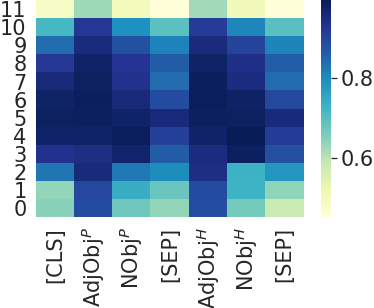}
        \caption{$\text{NP}_\text{Obj}$ Probe accuracy.
        Random baseline: 25\%}
        \label{fig:NP_Obj_probing_acc}
    \end{subfigure}\hfill
    \begin{subfigure}[t]{0.24\textwidth}
        \centering
        \includegraphics[width=0.9\textwidth]{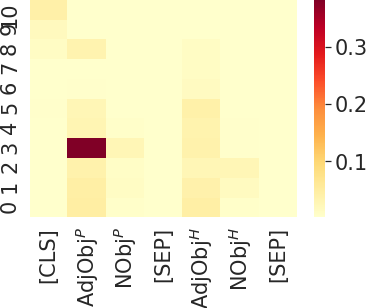}
        \caption{$\text{NP}_\text{Obj}$ Clique size.}
        \label{fig:NP_Obj_clq_size}
    \end{subfigure}\hfill
    \begin{subfigure}[t]{0.24\textwidth}
        \centering
        \includegraphics[width=0.9\textwidth]{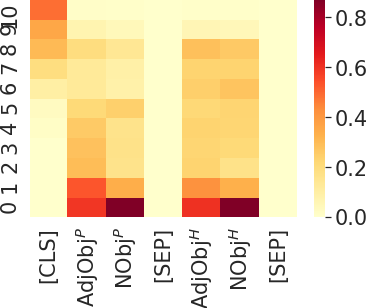}
        \caption{$\text{NP}_\text{Obj}$ Interchange success.}
        \label{fig:NP_Obj_interx}
    \end{subfigure}
    }
    
    \makebox[\textwidth][c]{
    \begin{subfigure}[t]{0.24\textwidth}
        \centering
        \includegraphics[width=0.9\textwidth]{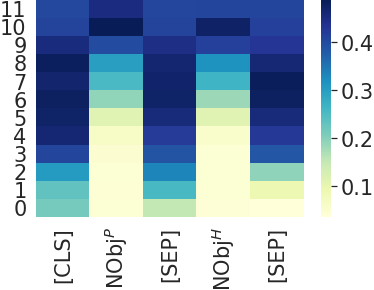}
        \caption{$\text{N}_\text{Obj}$ Probe selectivity.
        }
        \label{fig:N_Obj_probing_sel}
    \end{subfigure}\hfill
    \begin{subfigure}[t]{0.24\textwidth}
        \centering
        \includegraphics[width=0.9\textwidth]{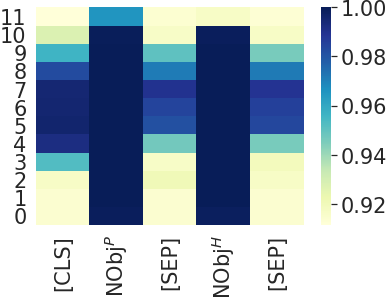}
        \caption{$\text{N}_\text{Obj}$ Probe accuracy.
        Random baseline: 14.3\%}
        \label{fig:N_Obj_probing_acc}
    \end{subfigure}\hfill
    \begin{subfigure}[t]{0.24\textwidth}
        \centering
        \includegraphics[width=0.9\textwidth]{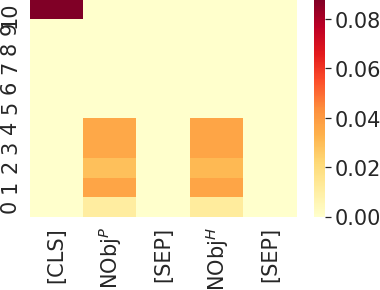}
        \caption{$\text{N}_\text{Obj}$ Clique size.}
        \label{fig:N_Obj_clq_size}
    \end{subfigure}\hfill
    \begin{subfigure}[t]{0.24\textwidth}
        \centering
        \includegraphics[width=0.9\textwidth]{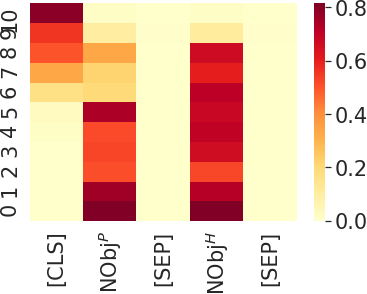}
        \caption{$\text{N}_\text{Obj}$ Interchange success.}
        \label{fig:N_Obj_interx}
    \end{subfigure}
    }
    \makebox[\textwidth][c]{
    \begin{subfigure}[t]{0.24\textwidth}
        \centering
        \includegraphics[width=0.9\textwidth]{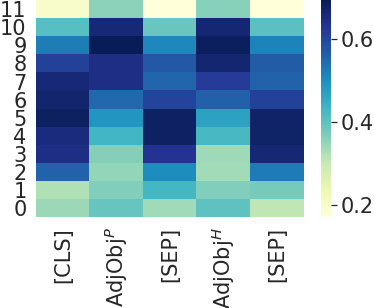}
        \caption{$\text{Adj}_\text{Obj}$ Probe selectivity.
        }
        \label{fig:Adj_Obj_probing_sel}
    \end{subfigure}\hfill
    \begin{subfigure}[t]{0.24\textwidth}
        \centering
        \includegraphics[width=0.9\textwidth]{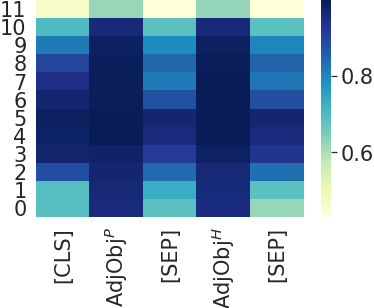}
        \caption{$\text{Adj}_\text{Obj}$ Probe accuracy.
        Random baseline: 14.3\%}
        \label{fig:Adj_Obj_probing_acc}
    \end{subfigure}\hfill
    \begin{subfigure}[t]{0.24\textwidth}
        \centering
        \includegraphics[width=0.9\textwidth]{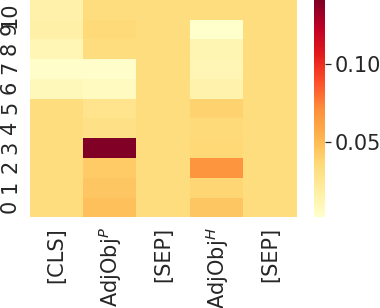}
        \caption{$\text{Adj}_\text{Obj}$ Clique size.}
        \label{fig:Adj_Obj_clq_size}
    \end{subfigure}\hfill
    \begin{subfigure}[t]{0.24\textwidth}
        \centering
        \includegraphics[width=0.9\textwidth]{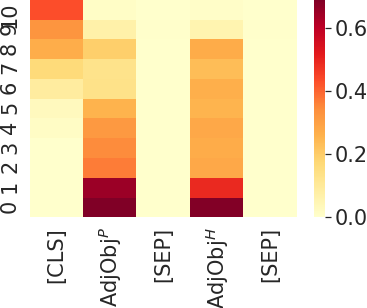}
        \caption{$\text{Adj}_\text{Obj}$ Interchange success.}
        \label{fig:Adj_Obj_interx}
    \end{subfigure}
    }
    
      \makebox[\textwidth][c]{
    \begin{subfigure}[t]{0.24\textwidth}
        \centering
        \includegraphics[width=0.9\textwidth]{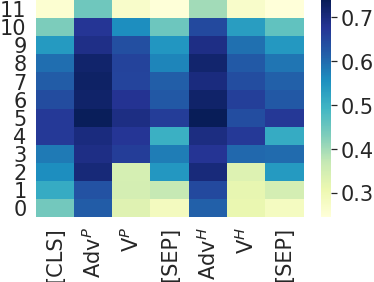}
        \caption{$\text{VP}$ Probe selectivity. 
        }
        \label{fig:VP_probing_sel}
    \end{subfigure}\hfill
    \begin{subfigure}[t]{0.24\textwidth}
        \centering
        \includegraphics[width=0.9\textwidth]{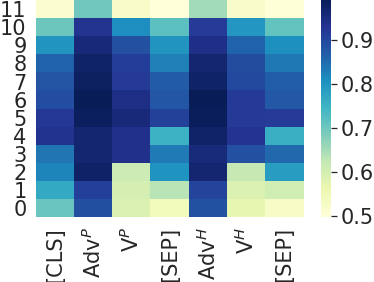}
        \caption{$\text{VP}$ Probe accuracy.
        Random baseline: 25\%}
        \label{fig:VP_probing_acc}
    \end{subfigure}\hfill
    \begin{subfigure}[t]{0.24\textwidth}
        \centering
        \includegraphics[width=0.9\textwidth]{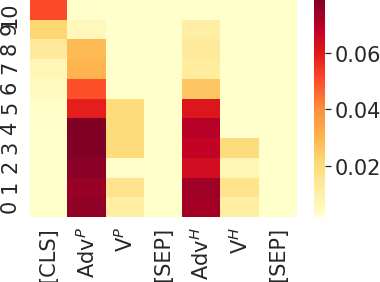}
        \caption{$\text{VP}$ Clique size.}
        \label{fig:VP_clq_size}
    \end{subfigure}\hfill
    \begin{subfigure}[t]{0.24\textwidth}
        \centering
        \includegraphics[width=0.9\textwidth]{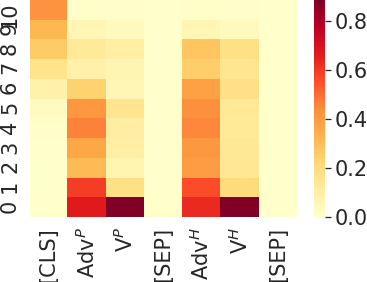}
        \caption{$\text{VP}$ Interchange success.}
        \label{fig:VP_interx}
    \end{subfigure}
    }
    
    \makebox[\textwidth][c]{
    \begin{subfigure}[t]{0.24\textwidth}
        \centering
        \includegraphics[width=0.9\textwidth]{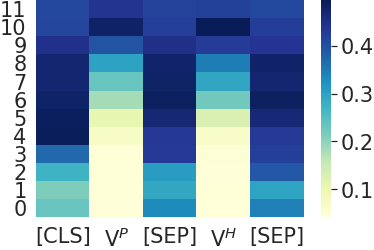}
        \caption{$\text{V}$ Probe selectivity.
        }
        \label{fig:V_probing_sel}
    \end{subfigure}\hfill
    \begin{subfigure}[t]{0.24\textwidth}
        \centering
        \includegraphics[width=0.9\textwidth]{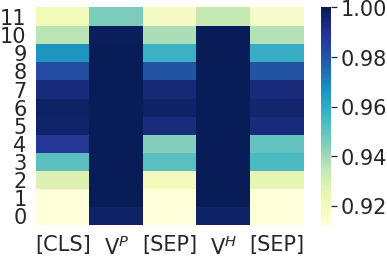}
        \caption{$\text{V}$ Probe accuracy.
        Random baseline: 14.3\%}
        \label{fig:V_probing_acc}
    \end{subfigure}\hfill
    \begin{subfigure}[t]{0.24\textwidth}
        \centering
        \includegraphics[width=0.9\textwidth]{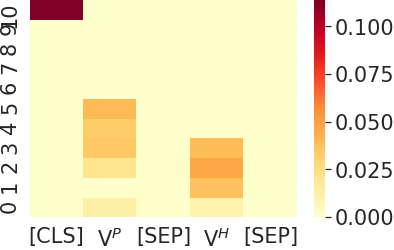}
        \caption{$\text{V}$ Clique size.}
        \label{fig:V_clq_size}
    \end{subfigure}\hfill
    \begin{subfigure}[t]{0.24\textwidth}
        \centering
        \includegraphics[width=0.9\textwidth]{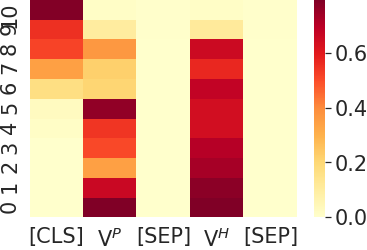}
        \caption{$\text{V}$ Interchange success.}
        \label{fig:V_interx}
    \end{subfigure}
    }
    \makebox[\textwidth][c]{
    \begin{subfigure}[t]{0.24\textwidth}
        \centering
        \includegraphics[width=0.9\textwidth]{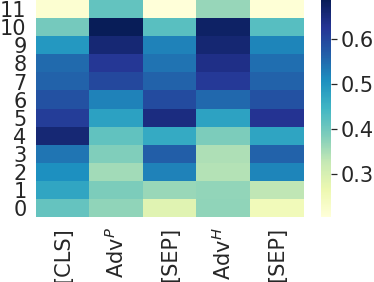}
        \caption{$\text{Adv}$ Probe selectivity.
        }
        \label{fig:Adv_probing_sel}
    \end{subfigure}\hfill
    \begin{subfigure}[t]{0.24\textwidth}
        \centering
        \includegraphics[width=0.9\textwidth]{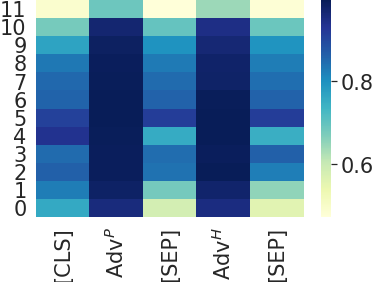}
        \caption{$\text{Adv}$ Probe accuracy.
        Random baseline: 14.3\%}
        \label{fig:Adv_probing_acc}
    \end{subfigure}\hfill
    \begin{subfigure}[t]{0.24\textwidth}
        \centering
        \includegraphics[width=0.9\textwidth]{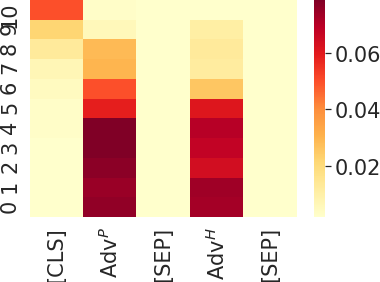}
        \caption{$\text{Adv}$ Clique size.}
        \label{fig:Adv_clq_size}
    \end{subfigure}\hfill
    \begin{subfigure}[t]{0.24\textwidth}
        \centering
        \includegraphics[width=0.9\textwidth]{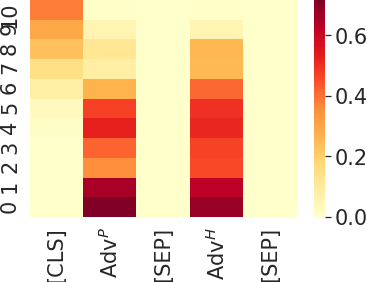}
        \caption{$\text{Adv}$ Interchange success.}
        \label{fig:Adv_interx}
    \end{subfigure}
    }

    \caption{Full probing and interchange intervention results for high-level nodes $\text{NP}_\text{Obj}$, $\text{N}_\text{Obj}$, $\text{Adj}_\text{Obj}$, VP, V, and Adv. Vertical axes denote BERT layers and horizontal axes denote the token position of hidden representations. Intervention success rates are based on experiments with a change in the output label. Clique sizes are reported as a percentage of all examples.}
    \label{fig:probing-results-extended-1}
\end{figure*}

In \dashfigref{fig:probing-results-extended-1}{fig:probing-results-extended-3} we report some more representative selectivity and accuracy results for our probing experiments on BERT trained on the hard variant, juxtaposed against intervention experiments on the same model. For open-class words and full phrases, probing and intervention show similar trends. For aligned closed-class words, we find near-zero selectivity because the domain of the control function is so small. 

In general, probing and intervention experiments for relations between aligned single open-class words (i.e., $\NSubj$, $\AdjSubj$, $ \NObj$, $\AdjObj$, $\Adv$, $\V$) show similar trends, which can be seen in Figures~\ref{fig:N_Subj_probing_sel}--\ref{fig:N_Subj_interx}. Every location except those above the [CLS] and [SEP] tokens has a near-100\% accuracy, while selectivity is only high in the last few layers. Lower layers of BERT contains more information about word identity and hence may allow the probe to memorize each input pair, resulting in higher control task accuracy and lower selectivity for lower layers. 

\begin{figure*}[!ht]
    \makebox[\textwidth][c]{
    \begin{subfigure}[t]{0.24\textwidth}
        \centering
        \includegraphics[width=0.9\textwidth]{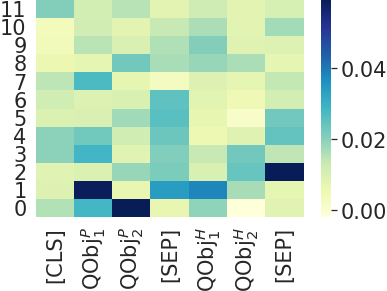}
        \caption{$\text{Q}_\text{Obj}$ Probe selectivity.}
        \label{fig:Q_Obj_probing_sel}
    \end{subfigure}\hfill
    \begin{subfigure}[t]{0.24\textwidth}
        \centering
        \includegraphics[width=0.9\textwidth]{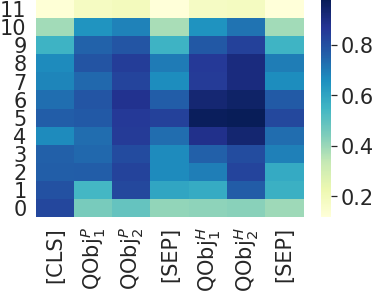}
        \caption{$\text{Q}_\text{Obj}$ Probe accuracy.
        Random Baseline: 6.25\%}
        \label{fig:Q_Obj_probing_acc}
    \end{subfigure}\hfill
    \begin{subfigure}[t]{0.24\textwidth}
        \centering
        \includegraphics[width=0.9\textwidth]{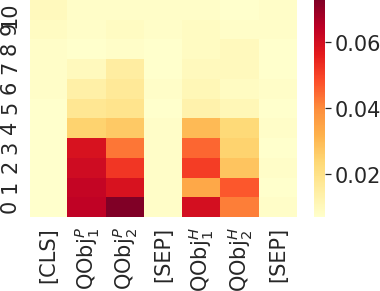}
        \caption{$\text{Q}_\text{Obj}$ Clique size.}
        \label{fig:Q_Obj_clq_size}
    \end{subfigure}\hfill
    \begin{subfigure}[t]{0.24\textwidth}
        \centering
        \includegraphics[width=0.9\textwidth]{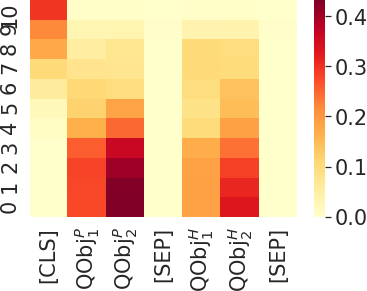}
        \caption{$\text{Q}_\text{Obj}$ Interchange success.}
        \label{fig:Q_Obj_interx}
    \end{subfigure}
    }

    \makebox[\textwidth][c]{
    \begin{subfigure}[t]{0.24\textwidth}
        \centering
        \includegraphics[width=0.9\textwidth]{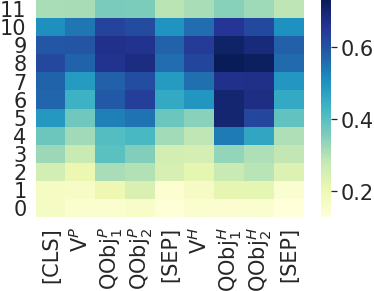}
        \caption{$\text{QP}_\text{Obj}$ Probe selectivity.}
        \label{fig:QP_Obj_probing_sel}
    \end{subfigure}\hfill
    \begin{subfigure}[t]{0.24\textwidth}
        \centering
        \includegraphics[width=0.9\textwidth]{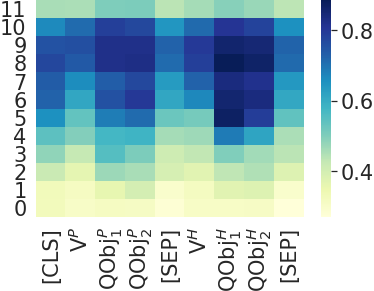}
        \caption{$\text{QP}_\text{Obj}$ Probe accuracy.
        Random Baseline 14.3\%}
        \label{fig:QP_Obj_probing_acc}
    \end{subfigure}\hfill
    \begin{subfigure}[t]{0.24\textwidth}
        \centering
        \includegraphics[width=0.9\textwidth]{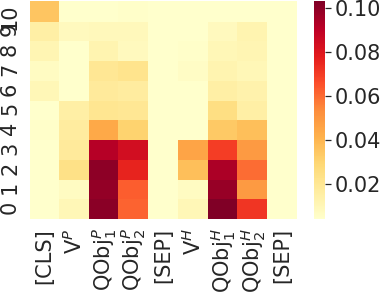}
        \caption{$\text{QP}_\text{Obj}$ Clique size.}
        \label{fig:QP_Obj_clq_size}
    \end{subfigure}\hfill
    \begin{subfigure}[t]{0.24\textwidth}
        \centering
        \includegraphics[width=0.9\textwidth]{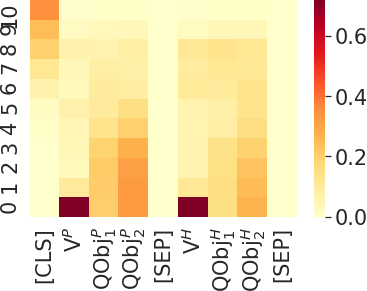}
        \caption{$\text{QP}_\text{Obj}$ Interchange success.}
        \label{fig:QP_Obj_interx}
    \end{subfigure}
    }

    \makebox[\textwidth][c]{
    \begin{subfigure}[t]{0.24\textwidth}
        \centering
        \includegraphics[width=0.9\textwidth]{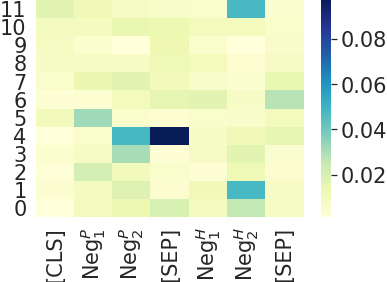}
        \caption{$\text{Neg}$ Probe selectivity.}
        \label{fig:Neg_probing_sel}
    \end{subfigure}\hfill
    \begin{subfigure}[t]{0.24\textwidth}
        \centering
        \includegraphics[width=0.9\textwidth]{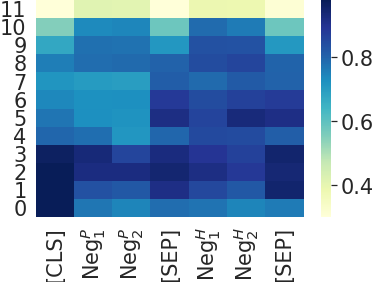}
        \caption{$\text{Neg}$ Probe accuracy.
        Random Baseline 25\%}
        \label{fig:Neg_probing_acc}
    \end{subfigure}\hfill
    \begin{subfigure}[t]{0.24\textwidth}
        \centering
        \includegraphics[width=0.9\textwidth]{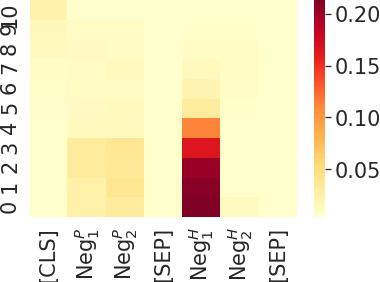}
        \caption{$\text{Neg}$ Clique size.}
        \label{fig:Neg_clq_size}
    \end{subfigure}\hfill
    \begin{subfigure}[t]{0.24\textwidth}
        \centering
        \includegraphics[width=0.9\textwidth]{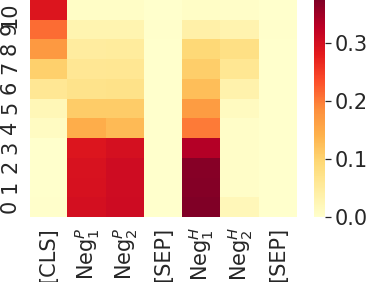}
        \caption{$\text{Neg}$ Interchange success.}
        \label{fig:Neg_interx}
    \end{subfigure}
    }

    \makebox[\textwidth][c]{
    \begin{subfigure}[t]{0.24\textwidth}
        \centering
        \includegraphics[width=0.9\textwidth]{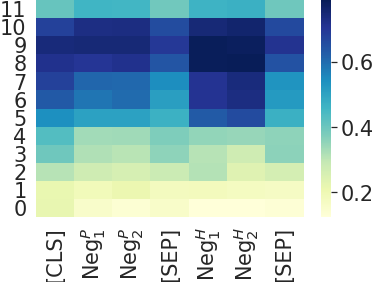}
        \caption{$\text{NegP}$ Probe selectivity.}
        \label{fig:NegP_probing_sel}
    \end{subfigure}\hfill
    \begin{subfigure}[t]{0.24\textwidth}
        \centering
        \includegraphics[width=0.9\textwidth]{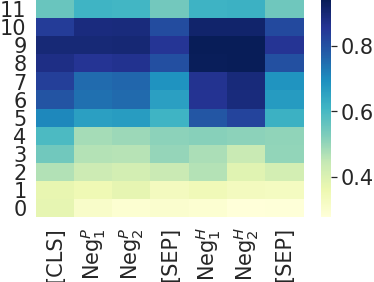}
        \caption{$\text{NegP}$ Probe accuracy.
        Random Baseline 25\%}
        \label{fig:NegP_probing_acc}
    \end{subfigure}\hfill
    \begin{subfigure}[t]{0.24\textwidth}
        \centering
        \includegraphics[width=0.9\textwidth]{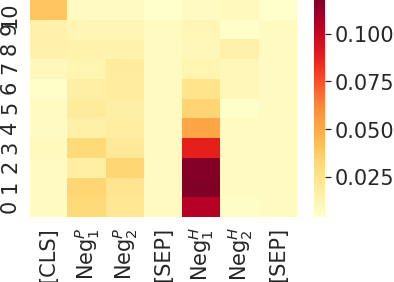}
        \caption{$\text{NegP}$ Clique size.}
        \label{fig:NegP_clq_size}
    \end{subfigure}\hfill
    \begin{subfigure}[t]{0.24\textwidth}
        \centering
        \includegraphics[width=0.9\textwidth]{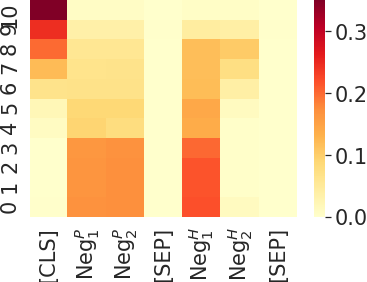}
        \caption{$\text{NegP}$ Interchange success.}
        \label{fig:NegP_interx}
    \end{subfigure}
    }
    \makebox[\textwidth][c]{
    \begin{subfigure}[t]{0.24\textwidth}
        \centering
        \includegraphics[width=0.9\textwidth]{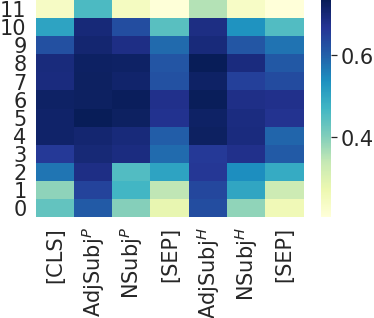}
        \caption{$\text{NP}_\text{Subj}$ Probe selectivity. 
        }
        \label{fig:NP_Subj_probing_sel}
    \end{subfigure}\hfill
    \begin{subfigure}[t]{0.24\textwidth}
        \centering
        \includegraphics[width=0.9\textwidth]{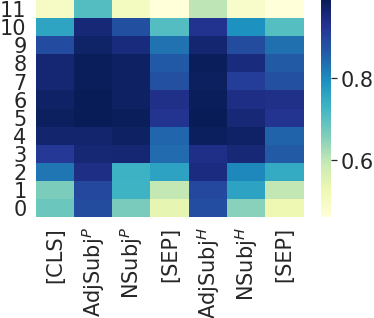}
        \caption{$\text{NP}_\text{Subj}$ Probe accuracy.
        Random baseline: 25\%}
        \label{fig:NP_Subj_probing_acc}
    \end{subfigure}\hfill
    \begin{subfigure}[t]{0.24\textwidth}
        \centering
        \includegraphics[width=0.9\textwidth]{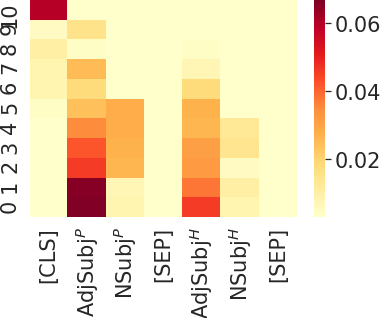}
        \caption{$\text{NP}_\text{Subj}$ Clique size.}
        \label{fig:NP_Subj_clq_size}
    \end{subfigure}\hfill
    \begin{subfigure}[t]{0.24\textwidth}
        \centering
        \includegraphics[width=0.9\textwidth]{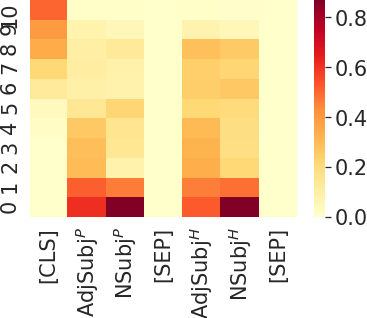}
        \caption{$\text{NP}_\text{Subj}$ Interchange success.}
        \label{fig:NP_Subj_interx}
    \end{subfigure}
    }
    
    \makebox[\textwidth][c]{
    \begin{subfigure}[t]{0.24\textwidth}
        \centering
        \includegraphics[width=0.9\textwidth]{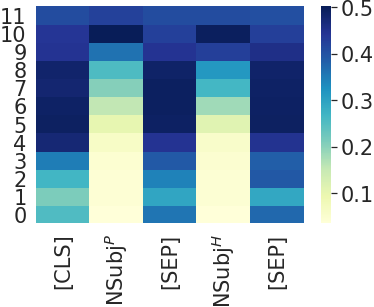}
        \caption{$\text{N}_\text{Subj}$ Probe selectivity.
        }
        \label{fig:N_Subj_probing_sel:app}
    \end{subfigure}\hfill
    \begin{subfigure}[t]{0.24\textwidth}
        \centering
        \includegraphics[width=0.9\textwidth]{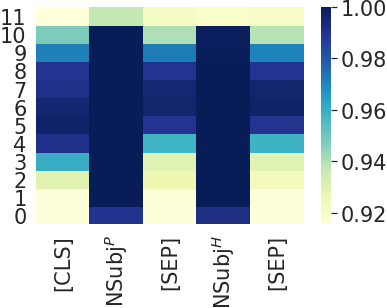}
        \caption{$\text{N}_\text{Subj}$ Probe accuracy.
        Random baseline: 14.3\%}
        \label{fig:N_Subj_probing_acc:app}
    \end{subfigure}\hfill
    \begin{subfigure}[t]{0.24\textwidth}
        \centering
        \includegraphics[width=0.9\textwidth]{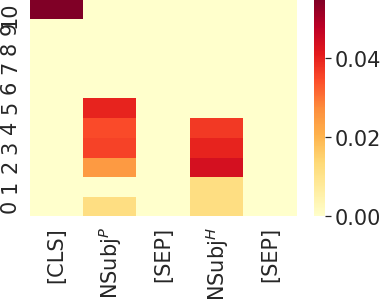}
        \caption{$\text{N}_\text{Subj}$ Clique size.}
        \label{fig:N_Subj_clq_size:app}
    \end{subfigure}\hfill
    \begin{subfigure}[t]{0.24\textwidth}
        \centering
        \includegraphics[width=0.9\textwidth]{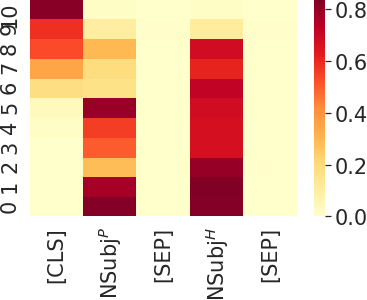}
        \caption{$\text{N}_\text{Subj}$ Interchange success.}
        \label{fig:N_Subj_interx:app}
    \end{subfigure}
    }

    \caption{Full probing and interchange intervention results on the high level nodes $\text{Q}_\text{Obj}$, $\text{QP}_\text{Obj}$, Neg, NegP, $\text{NP}_\text{Subj}$ and $\text{N}_\text{Subj}$. Vertical axes denote BERT layers and horizontal axes denote the token position of hidden representations. Intervention success rates are based on experiments with a change in the output label. Clique sizes are reported as a percentage of all examples.}
    \label{fig:probing-results-extended-2}
\end{figure*}

Probing experiments for relations between aligned multi-word subphrases (i.e., $\NPSubj$, $\VP$, $\NPObj$, $\QPObj$ and $\NegP$) show similar trends as shown in the row of figures \ref{fig:NegP_probing_sel} to \ref{fig:QP_Obj_interx}. As described in \secref{sec:probe_control_task}, all control probes for these achieve near-random performance, so selectivity and accuracy differ by the random baseline accuracy, which is evident by comparing figures \ref{fig:NegP_probing_sel} and \ref{fig:NegP_probing_acc}.

On the other hand, probing experiments for aligned closed-class words (quantifiers and negation) have near-zero selectivity, as shown in  \figref{fig:Q_Obj_probing_sel}. This is because the domain of the control function is the small set of closed-class word pairs, so memorizing the identity of these words becomes trivial for the probe.

\section{Probing and Intervention Heatmaps}\label{app:heatmaps}

\begin{figure*}[!ht]
    \makebox[\textwidth][c]{
    \begin{subfigure}[t]{0.24\textwidth}
        \centering
        \includegraphics[width=0.9\textwidth]{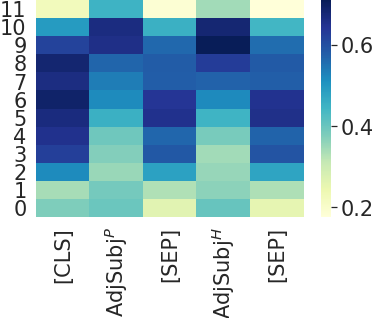}
        \caption{$\text{Adj}_\text{Subj}$ Probe selectivity.
        }
        \label{fig:Adj_Subj_probing_sel}
    \end{subfigure}\hfill
    \begin{subfigure}[t]{0.24\textwidth}
        \centering
        \includegraphics[width=0.9\textwidth]{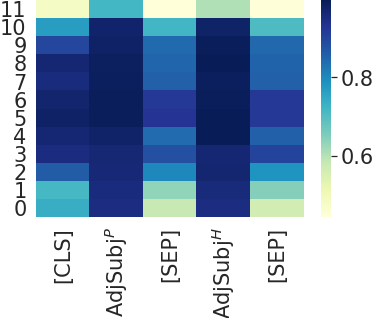}
        \caption{$\text{Adj}_\text{Subj}$ Probe accuracy.
        Random baseline: 14.3\%}
        \label{fig:Adj_Subj_probing_acc}
    \end{subfigure}\hfill
    \begin{subfigure}[t]{0.24\textwidth}
        \centering
        \includegraphics[width=0.9\textwidth]{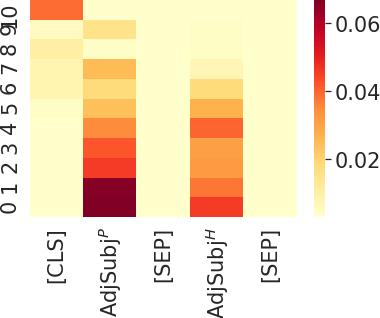}
        \caption{$\text{Adj}_\text{Subj}$ Clique size.}
        \label{fig:Adj_Subj_clq_size}
    \end{subfigure}\hfill
    \begin{subfigure}[t]{0.24\textwidth}
        \centering
        \includegraphics[width=0.9\textwidth]{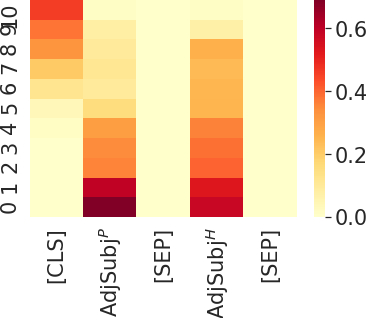}
        \caption{$\text{Adj}_\text{Subj}$ Interchange success.}
        \label{fig:Adj_Subj_interx}
    \end{subfigure}
    }

    \makebox[\textwidth][c]{
    \begin{subfigure}[t]{0.24\textwidth}
        \centering
        \includegraphics[width=0.9\textwidth]{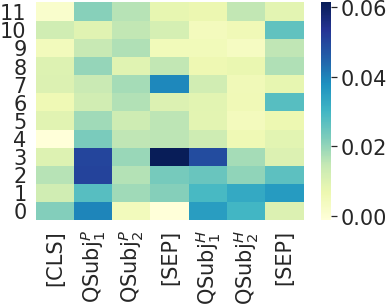}
        \caption{$\text{Q}_\text{Subj}$ Probe selectivity.}
        \label{fig:Q_Subj_probing_sel}
    \end{subfigure}\hfill
    \begin{subfigure}[t]{0.24\textwidth}
        \centering
        \includegraphics[width=0.9\textwidth]{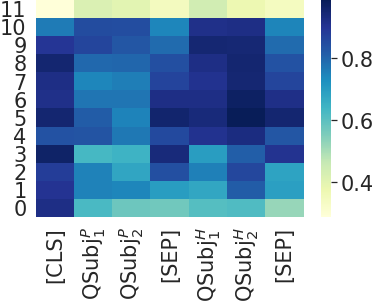}
        \caption{$\text{Q}_\text{Subj}$ Probe accuracy.
        Random Baseline: 6.25\%}
        \label{fig:Q_Subj_probing_acc}
    \end{subfigure}\hfill
    \begin{subfigure}[t]{0.24\textwidth}
        \centering
        \includegraphics[width=0.9\textwidth]{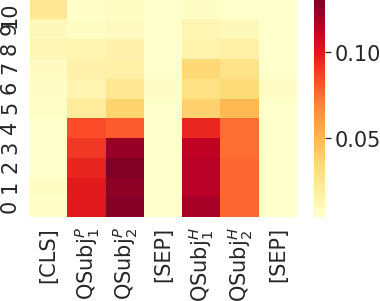}
        \caption{$\text{Q}_\text{Subj}$ Clique size.}
        \label{fig:Q_Subj_clq_size}
    \end{subfigure}\hfill
    \begin{subfigure}[t]{0.24\textwidth}
        \centering
        \includegraphics[width=0.9\textwidth]{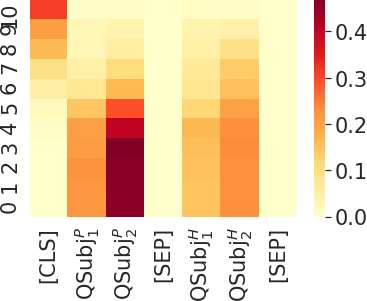}
        \caption{$\text{Q}_\text{Subj}$ Interchange success.}
        \label{fig:Q_Subj_interx}
    \end{subfigure}
    }

    \caption{Full probing and interchange intervention results for high-level nodes $\text{Adj}_\text{Subj}$ and $\text{Q}_\text{Subj}$. Vertical axes denote BERT layers and horizontal axes denote the token position of hidden representations. Intervention success rates are based on experiments with a change in the output label. Clique sizes are reported as a percentage of all examples.}
    \label{fig:probing-results-extended-3}
\end{figure*}

\section{Integrated Gradients}\label{app:IG}
\begin{figure}[t!]
\centering
\includegraphics[width=0.55\linewidth]{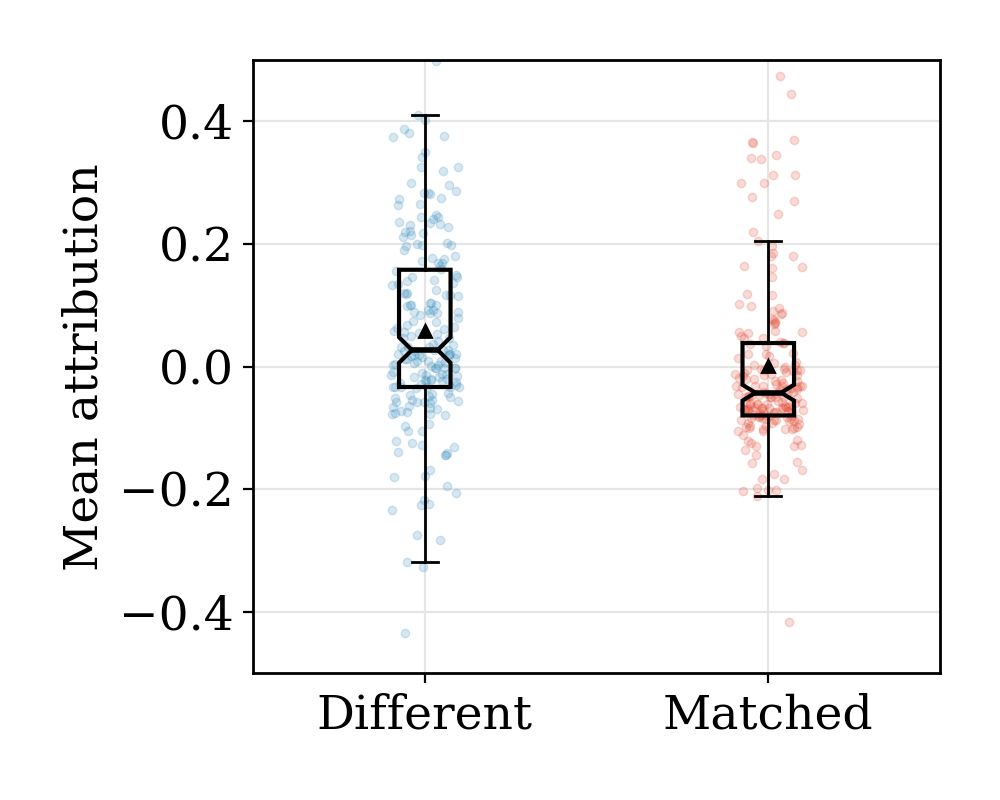}
\caption{Integrated Gradients values for examples in which the premise and hypothesis differ by in exactly one aligned position. `Different' refers to the IG value for this position, and `Matched' is a randomly selected different position from each example. The two populations are different according to a Wilcoxon signed-rank test ($p < 0.00001$). The `Different' positions have positive attribution on average, aligning with our expectation that they tend to be decisive for the output prediction.}
\label{fig:ig-cmp}
\end{figure}

We report attributions for the first BERT layer; later layers tend to concentrate importance onto the [CLS] token, since it is the direct basis for the classifier head in our model. To simplify the analysis, we restrict attention to examples in which exactly one position is different across the premise and hypothesis, and `Matched' is a randomly selected position from elsewhere in the example.  We see that the `Matched' are positive in general, which aligns with our expectation that they are the most important positions in these examples (\figref{fig:ig-cmp}).

\section{Background on Causal Models and Causal Abstraction}\label{app:causalbackground}
In this appendix we provide relevant background on causal models and causal abstraction, sufficient to define the notion of \emph{constructive abstraction}. 

\subsection{Causal Models}

\begin{definition}{(Signatures)}
A signature $S$ is a pair $(\mathcal{V}, \mathcal{R})$, where $\mathcal{V}$ is a set of variables and $\mathcal{R}$ is a function that associates with every variable $X\in \mathcal{V}$ a nonempty set $\mathcal{R}(X)$ of possible 
values. If $\mathbf{X} = (X_1, \dots , X_n)$, $\mathcal{R}(\mathbf{X})$ denotes the cross product $\mathcal{R}(X_1) \times \dots \times \mathcal{R}(X_n)$.
\end{definition} 

\begin{definition}{(Causal models)}
A causal model $M$ is a pair $(\mathcal{S}, \mathcal{F})$,
where $\mathcal{S}$ is a signature and $\mathcal{F}$ defines a function that associates with each variable $X$ a structural equation $\mathcal{F}^X$ giving the value of $X$ in terms of the values of other variables. Formally, the equation $\mathcal{F}^X$ maps $\mathcal{R}(\mathcal{V} - \{X\})$ to
$\mathcal{R}(X)$, so $\mathcal{F}^X$ determines the value of $X$, given the values of
all the other variables in $\mathcal{V}$. 
\end{definition} 

\begin{definition}{(Dependence)}
$X$ causes $Y$ according to $M$, denoted $X \rightsquigarrow Y$, if there is some
setting of the variables other than $X$ and $Y$ such that varying the
value of $X$ results in a variation in the value
of $Y$; that is, there is a setting $\mathbf{z}$ of the variables
$\mathbf{Z} = \mathcal{V} - \{X,Y\}$ and values $x$ and $x'$ of $X$ 
$\mathcal{F}^Y (x, \mathbf{z}) \neq \mathcal{F}^Y (x', \mathbf{z})$.
\end{definition}

\begin{definition}{(Intervention)}
An intervention $i$ has the form $\mathbf{X} \leftarrow \mathbf{x}$, where $\mathbf{X}$ is a vector of
variables. Intuitively, this means that the values
of the variables in $\mathbf{X}$ are set to $\mathbf{x}$.
Setting the value of some variables $\mathbf{X} \leftarrow \mathbf{x}$ in a causal
model $M = (\mathcal{S}, \mathcal{F})$ results in a new causal model, denoted
$i(M)$, which is identical to $M$, except that $\mathcal{F}$ is replaced
by $i(\mathcal{F})$: for each variable $Y \not \in \mathbf{X}$, $i(\mathcal{F}^Y) =  
\mathcal{F}^Y$ (i.e., the
equation for $Y$ is unchanged), while for each $X' \in \mathbf{X}$ ,
$i(F^{X'})$ is the constant function sending all arguments to $x'$ (where $x'$ is the value in $\mathbf{x}$ corresponding to $X_i$).
\end{definition}

When we write out the structured equations for a variable $X$, for simplicity's sake, we treat $\mathcal{F}^X$ as a map from $\mathcal{R}(\{Y \in \mathcal{V}: Y \rightsquigarrow X\})$ to $\mathcal{R}(X)$. 

Note that interventions $\mathbf{X} \leftarrow \mathbf{x}$ correspond 1--1 with variable settings $\mathbf{x}$. We make use of this in what follows. 

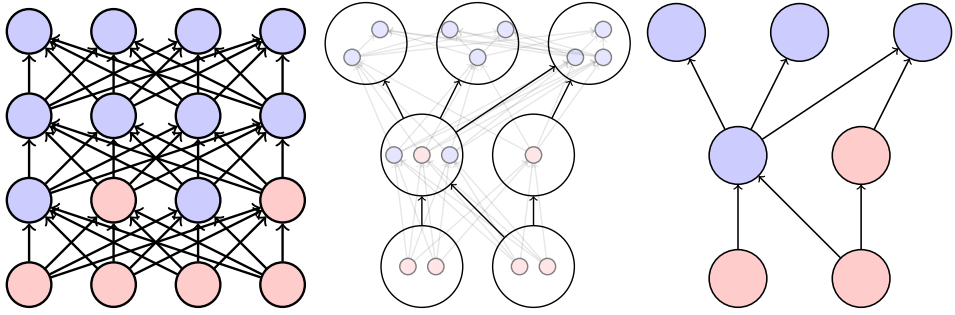
\begin{figure}[t!]
    \centering
    \begin{subfigure}[t]{0.3\textwidth}
\resizebox{\textwidth}{!}{
\begin{tikzpicture}[thick,scale=0.5, every node/.style={scale=0.6}]
\node[fill=red!20, draw, circle, minimum size=25pt] (N00) at (0,0) {};
\node[fill=red!20, draw, circle, minimum size=25pt] (N10) at (2,0) {};
\node[fill=red!20, draw, circle, minimum size=25pt] (N20) at (4,0) {};
\node[fill=red!20, draw, circle, minimum size=25pt] (N30) at (6,0) {};
\node[fill=blue!20, draw, circle, minimum size=25pt] (N01) at (0,2) {};
\node[fill=red!20, draw, circle, minimum size=25pt] (N11) at (2,2) {};
\node[fill=blue!20, draw, circle, minimum size=25pt] (N21) at (4,2) {};
\node[fill=red!20, draw, circle, minimum size=25pt] (N31) at (6,2) {};
\node[fill=blue!20, draw, circle, minimum size=25pt] (N02) at (0,4) {};
\node[fill=blue!20, draw, circle, minimum size=25pt] (N12) at (2,4) {};
\node[fill=blue!20, draw, circle, minimum size=25pt] (N22) at (4,4) {};
\node[fill=blue!20, draw, circle, minimum size=25pt] (N32) at (6,4) {};
\node[fill=blue!20, draw, circle, minimum size=25pt] (N03) at (0,6) {};
\node[fill=blue!20, draw, circle, minimum size=25pt] (N13) at (2,6) {};
\node[fill=blue!20, draw, circle, minimum size=25pt] (N23) at (4,6) {};
\node[fill=blue!20, draw, circle, minimum size=25pt] (N33) at (6,6) {};
\draw [->] (N00) -- (N01) ;
\draw [->] (N00) -- (N11) ;
\draw [->] (N00) -- (N21) ;
\draw [->] (N00) -- (N31) ;
\draw [->] (N10) -- (N01) ;
\draw [->] (N10) -- (N11) ;
\draw [->] (N10) -- (N21) ;
\draw [->] (N10) -- (N31) ;
\draw [->] (N20) -- (N01) ;
\draw [->] (N20) -- (N11) ;
\draw [->] (N20) -- (N21) ;
\draw [->] (N20) -- (N31) ;
\draw [->] (N30) -- (N01) ;
\draw [->] (N30) -- (N11) ;
\draw [->] (N30) -- (N21) ;
\draw [->] (N30) -- (N31) ;
\draw [->] (N01) -- (N02) ;
\draw [->] (N01) -- (N12) ;
\draw [->] (N01) -- (N22) ;
\draw [->] (N01) -- (N32) ;
\draw [->] (N11) -- (N02) ;
\draw [->] (N11) -- (N12) ;
\draw [->] (N11) -- (N22) ;
\draw [->] (N11) -- (N32) ;
\draw [->] (N21) -- (N02) ;
\draw [->] (N21) -- (N12) ;
\draw [->] (N21) -- (N22) ;
\draw [->] (N21) -- (N32) ;
\draw [->] (N31) -- (N02) ;
\draw [->] (N31) -- (N12) ;
\draw [->] (N31) -- (N22) ;
\draw [->] (N31) -- (N32) ;
\draw [->] (N02) -- (N03) ;
\draw [->] (N02) -- (N13) ;
\draw [->] (N02) -- (N23) ;
\draw [->] (N02) -- (N33) ;
\draw [->] (N12) -- (N03) ;
\draw [->] (N12) -- (N13) ;
\draw [->] (N12) -- (N23) ;
\draw [->] (N12) -- (N33) ;
\draw [->] (N22) -- (N03) ;
\draw [->] (N22) -- (N13) ;
\draw [->] (N22) -- (N23) ;
\draw [->] (N22) -- (N33) ;
\draw [->] (N32) -- (N03) ;
\draw [->] (N32) -- (N13) ;
\draw [->] (N32) -- (N23) ;
\draw [->] (N32) -- (N33) ;

\end{tikzpicture}
}
\end{subfigure}
\begin{subfigure}[t]{0.3\textwidth}
\resizebox{\textwidth}{!}{
\begin{tikzpicture}[thick,scale=0.37, every node/.style={scale=0.37}]
\node[draw, circle, minimum size=125pt] (B00) at (0+0,0+0) {};
\node[opacity = 0.5,fill=red!20, draw, circle, minimum size=25pt] (BN00) at (0+-0.75,0+0) {};
\node[opacity = 0.5,fill=red!20, draw, circle, minimum size=25pt] (BN10) at (0+0.75,0+0) {};
\node[draw, circle, minimum size=125pt] (B10) at (0+6,0+0) {};
\node[opacity = 0.5,fill=red!20, draw, circle, minimum size=25pt] (BN20) at (0+5.25,0+0) {};
\node[opacity = 0.5,fill=red!20, draw, circle, minimum size=25pt] (BN30) at (0+6.75,0+0) {};
\node[draw, circle, minimum size=125pt] (B01) at (0+0,0+6) {};
\node[opacity = 0.5,fill=blue!20, draw, circle, minimum size=25pt] (BN01) at (0+-1.5,0+6) {};
\node[opacity = 0.5,fill=red!20, draw, circle, minimum size=25pt] (BN11) at (0+0.0,0+6) {};
\node[opacity = 0.5,fill=blue!20, draw, circle, minimum size=25pt] (BN21) at (0+1.5,0+6) {};
\node[draw, circle, minimum size=125pt] (B11) at (0+6,0+6) {};
\node[opacity = 0.5,fill=red!20, draw, circle, minimum size=25pt] (BN31) at (0+6.0,0+6) {};
\node[draw, circle, minimum size=125pt] (B02) at (0+-3.0,0+12) {};
\node[opacity = 0.5,fill=blue!20, draw, circle, minimum size=25pt] (BN02) at (0+-3.75,0+11.25) {};
\node[opacity = 0.5,fill=blue!20, draw, circle, minimum size=25pt] (BN13) at (0+-2.25,0+12.75) {};
\node[draw, circle, minimum size=125pt] (B12) at (0+3.0,0+12) {};
\node[opacity = 0.5,fill=blue!20, draw, circle, minimum size=25pt] (BN03) at (0+1.5,0+12.75) {};
\node[opacity = 0.5,fill=blue!20, draw, circle, minimum size=25pt] (BN12) at (0+3.0,0+11.25) {};
\node[opacity = 0.5,fill=blue!20, draw, circle, minimum size=25pt] (BN23) at (0+4.5,0+12.75) {};
\node[draw, circle, minimum size=125pt] (B22) at (0+9.0,0+12) {};
\node[opacity = 0.5,fill=blue!20, draw, circle, minimum size=25pt] (BN22) at (0+8.25,0+11.25) {};
\node[opacity = 0.5,fill=blue!20, draw, circle, minimum size=25pt] (BN32) at (0+9.75,0+11.25) {};
\node[opacity = 0.5,fill=blue!20, draw, circle, minimum size=25pt] (BN33) at (0+9.75,0+12.75) {};
\draw [opacity =0.1, ->] (BN00) -- (BN01) ;
\draw [opacity =0.1, ->] (BN00) -- (BN11) ;
\draw [opacity =0.1, ->] (BN00) -- (BN21) ;
\draw [opacity =0.1, ->] (BN00) -- (BN31) ;
\draw [opacity =0.1, ->] (BN10) -- (BN01) ;
\draw [opacity =0.1, ->] (BN10) -- (BN11) ;
\draw [opacity =0.1, ->] (BN10) -- (BN21) ;
\draw [opacity =0.1, ->] (BN10) -- (BN31) ;
\draw [opacity =0.1, ->] (BN20) -- (BN01) ;
\draw [opacity =0.1, ->] (BN20) -- (BN11) ;
\draw [opacity =0.1, ->] (BN20) -- (BN21) ;
\draw [opacity =0.1, ->] (BN20) -- (BN31) ;
\draw [opacity =0.1, ->] (BN30) -- (BN01) ;
\draw [opacity =0.1, ->] (BN30) -- (BN11) ;
\draw [opacity =0.1, ->] (BN30) -- (BN21) ;
\draw [opacity =0.1, ->] (BN30) -- (BN31) ;
\draw [opacity =0.1, ->] (BN01) -- (BN02) ;
\draw [opacity =0.1, ->] (BN01) -- (BN12) ;
\draw [opacity =0.1, ->] (BN01) -- (BN22) ;
\draw [opacity =0.1, ->] (BN01) -- (BN32) ;
\draw [opacity =0.1, ->] (BN11) -- (BN02) ;
\draw [opacity =0.1, ->] (BN11) -- (BN12) ;
\draw [opacity =0.1, ->] (BN11) -- (BN22) ;
\draw [opacity =0.1, ->] (BN11) -- (BN32) ;
\draw [opacity =0.1, ->] (BN21) -- (BN02) ;
\draw [opacity =0.1, ->] (BN21) -- (BN12) ;
\draw [opacity =0.1, ->] (BN21) -- (BN22) ;
\draw [opacity =0.1, ->] (BN21) -- (BN32) ;
\draw [opacity =0.1, ->] (BN31) -- (BN02) ;
\draw [opacity =0.1, ->] (BN31) -- (BN12) ;
\draw [opacity =0.1, ->] (BN31) -- (BN22) ;
\draw [opacity =0.1, ->] (BN31) -- (BN32) ;
\draw [opacity =0.1, ->] (BN02) -- (BN03) ;
\draw [opacity =0.1, ->] (BN02) -- (BN13) ;
\draw [opacity =0.1, ->] (BN02) -- (BN23) ;
\draw [opacity =0.1, ->] (BN02) -- (BN33) ;
\draw [opacity =0.1, ->] (BN12) -- (BN03) ;
\draw [opacity =0.1, ->] (BN12) -- (BN13) ;
\draw [opacity =0.1, ->] (BN12) -- (BN23) ;
\draw [opacity =0.1, ->] (BN12) -- (BN33) ;
\draw [opacity =0.1, ->] (BN22) -- (BN03) ;
\draw [opacity =0.1, ->] (BN22) -- (BN13) ;
\draw [opacity =0.1, ->] (BN22) -- (BN23) ;
\draw [opacity =0.1, ->] (BN22) -- (BN33) ;
\draw [opacity =0.1, ->] (BN32) -- (BN03) ;
\draw [opacity =0.1, ->] (BN32) -- (BN13) ;
\draw [opacity =0.1, ->] (BN32) -- (BN23) ;
\draw [opacity =0.1, ->] (BN32) -- (BN33) ;
\draw [->] (B00) -- (B01) ;
\draw [->] (B10) -- (B01) ;
\draw [->] (B10) -- (B11) ;
\draw [->] (B01) -- (B02) ;
\draw [->] (B01) -- (B12) ;
\draw [->] (B01) -- (B22) ;
\draw [->] (B11) -- (B22) ;
\end{tikzpicture}
}
\end{subfigure}
\begin{subfigure}[t]{0.3\textwidth}
\resizebox{\textwidth}{!}{
\begin{tikzpicture}[thick,scale=0.37, every node/.style={scale=0.37}]
\node[fill=red!20,draw, circle, minimum size=80pt] (BB00) at (0+0,0+0) {};
\node[fill=red!20,draw, circle, minimum size=80pt] (BB10) at (0+6,0+0) {};
\node[fill=blue!20,draw, circle, minimum size=80pt] (BB01) at (0+0,0+6) {};
\node[fill=red!20,draw, circle, minimum size=80pt] (BB11) at (0+6,0+6) {};
\node[fill=blue!20,draw, circle, minimum size=80pt] (BB02) at (0+-3.0,0+12) {};
\node[fill=blue!20,draw, circle, minimum size=80pt] (BB12) at (0+3.0,0+12) {};
\node[fill=blue!20,draw, circle, minimum size=80pt] (BB22) at (0+9.0,0+12) {};
\draw [->] (BB00) -- (BB01) ;
\draw [->] (BB10) -- (BB01) ;
\draw [->] (BB10) -- (BB11) ;
\draw [->] (BB01) -- (BB02) ;
\draw [->] (BB01) -- (BB12) ;
\draw [->] (BB01) -- (BB22) ;
\draw [->] (BB11) -- (BB22) ;
\end{tikzpicture}
}
\end{subfigure}
    \caption{Schematic depicting constructive abstraction \citep{Beckers_Halpern_2019}. The variables of the low-level model (left) are divided into partitions (center) such that each low-level partition corresponds to a high level variable from the high-level model (right). The circles represent variables and the arrows represent causal dependencies. Blue circles are variables that are not being intervened on and red circles are variables that are being intervened on. Observe that a low-level causal dependence between partitions does not necessarily result in a high-level causal dependence between variables and that not every low-level intervention results in a high level intervention.}
    \label{fig:abstraction}
\end{figure}

\subsection{Constructive Abstraction}
\newcommand{\project}[2]{\mathsf{Proj}(#1, #2)}
\newcommand{\inverseproject}[1]{\mathsf{Proj}^{-1}(#1)}

The following definitions are in agreement with the definitions from \citet{Beckers_Halpern_2019}, but differ somewhat in presentation. We additionally omit exogenous variables, as they play no role in our deterministic setting. In this section we take causal models to be pairs $(M,\mathcal{I})$, with a set $\mathcal{I}$ of \emph{admissible interventions} made explicit. 

\begin{definition}{(Projection and Inverse Projection)}
Given some $\mathbf{v} \in \mathcal{R}(\mathcal{V})$ and $\mathbf{X} \subseteq \mathcal{V}$, define $\project{\mathbf{v}}{\mathbf{X}}$ to be the restriction of $\mathbf{v}$ to the variables in $\mathbf{X}$.
Given some $\mathbf{x} \subseteq \mathcal{V}(\mathbf{X})$, the inverse 
$\inverseproject{\mathbf{x}}$ is defined as usual: $$\{\mathbf{v} \in \mathcal{R}(\mathcal{V}) : \mathbf{x} \text{ is the restriction of } \mathbf{v}\text{ to } \mathbf{X} \}.$$
\end{definition}

We are interested in (possibly partial) functions $\tau: \mathcal{R}_L(\mathcal{V}_L) \rightarrow \mathcal{R}_H(\mathcal{V}_H)$ mapping settings of low-level variables to settings of high-level variables. Such a function $\tau$ naturally induces a function $\omega_{\tau}$ between sets of interventions, where $\omega_\tau(\mathbf{x}) = \mathbf{y}$ just in case \begin{equation*}
    \tau(\inverseproject{\mathbf{x}})  =  \inverseproject{\mathbf{y}}.
\end{equation*}

We are now in a position to define $\tau$-abstraction:

\begin{definition}{($\tau$-abstraction)} Fix a function $\tau : \mathcal{R}_L(\mathcal{V}_L) \rightarrow \mathcal{R}_H(\mathcal{V}_H)$, which in turn fixes  $\omega_\tau: \mathcal{I}_L \rightarrow \mathcal{I}_H$. We say 
$(M_H, \mathcal{I}_H)$ is a $\tau$-abstraction of $(M_L, \mathcal{I}_L)$ 
if the following hold:
\begin{enumerate}\setlength{\itemsep}{0pt}
    \item $\tau$ is surjective.
    \item $\omega_\tau$ is surjective.
    \item for all $i_L \in \mathcal{I}_L$ we have $\tau(i_L(M_L)) = \omega_{\tau}(i_L)(M_H)$.
\end{enumerate} \label{def-tau}
\end{definition}
One way to think of this is: $\tau$ is a map from $\mathcal{R}(\mathcal{V}_L)$ to $\mathcal{R}(\mathcal{V}_H)$, which in turn induces a map $\omega_\tau$ from the space of \emph{projections} on $\mathcal{R}(\mathcal{V}_L)$ to \emph{projections} on $\mathcal{R}(\mathcal{V}_H)$. The conditions on $\tau$-abstraction below then simply become that $\tau$ and $\omega_\tau$ are both total and surjective on their respective (co)domains, and a second condition that can be easily encoded in terms of potential outcomes. For any setting/projection $\mathbf{x}$ at the low-level, we require that $M_L \models \mathbf{v}_{\mathbf{x}}$ iff $M_H \models \tau(\mathbf{v})_{\omega_\tau(\mathbf{x})}$. 

Finally, to be a \emph{constructive} $\tau$-abstraction we simply require that $\tau$ decompose into a family of ``component'' functions, as below. 

\begin{definition}[Constructive $\tau$-abstraction] $(M_H,\mathcal{I}_H)$ is a constructive $\tau$-abstraction of $(M_L,\mathcal{I}_L)$ if, in addition to being a  $\tau$-abstraction,  we can associate with each $X_H$ a subset $P_{X_H}$ of $\mathcal{V}_L$, such that the mapping $\tau:\mathcal{R}(\mathcal{V}_L) \rightarrow \mathcal{R}(\mathcal{V}_H)$ decomposes into a family of functions $\tau_{X_H}:\mathcal{R}(P_{X_H}) \rightarrow \mathcal{R}(X_H)$. 
We say $M_H$ is a constructive abstraction of $M_L$ if it is a constructive $\tau$-abstraction for some $\tau$.
\end{definition}
In other words, for a constructive abstraction it suffices to define the component functions $\tau_{X_H}$, as these completely determine $\tau$. In fact, the maps $\tau_{X_H}$ more generally induce a (partial) function from projections of $\mathcal{R}(\mathcal{V}_L)$ to (in fact, onto) projections of $\mathcal{R}(\mathcal{V}_H)$ in the following sense. For any setting $\mathbf{h} = [h_1\dots h_k]$ of high-level variables $H_1,\dots,H_k$ we can find low-level setting $\mathbf{y}$ such that projections of $\mathbf{y}$ map via $\tau_{H_i}$ to $h_i$. Slightly abusing notation,  denote this (partial) low-level setting $\mathbf{y}$ as $\tau^{-1}(\mathbf{h})$. So, in particular when $\mathbf{h}$ corresponds to an intervention in $\mathcal{I}_H$, the setting $\tau^{-1}(\mathbf{h})$ should specify a corresponding intervention in $\mathcal{I}_L$. Indeed, point (2) of Def.~\ref{def-tau} tells us that (the intervention corresponding to) $\tau^{-1}(\mathbf{h})$ should be mapped via $\omega_\tau$ to (the intervention corresponding to) $\mathbf{h}$. 

\label{section:constructive-abstraction}


\section{Causal Abstraction Analysis of $\addmod$}\label{app:addition}

\subsection{Formal Definition of $\addmod$}
We define the causal model $\addmod = ( \mathcal{V}_+, \mathcal{R}_+,\mathcal{F}_+)$ as follows (where $\mathbb{N}_k = \{0,\dots,k\})$: 
\begin{align*}
\mathcal{V}_+ &= \{X,Y,Z,W, S_1, S_2\} &  &
\\[1ex]
\mathcal{R}_+(V) &= \mathbb{N}_9\text{, for }V \in \{X,Y,Z,W\} \\
\mathcal{R}_+(S_1) &= \mathbb{N}_{18} && \\ \mathcal{R}_+(S_2)& = \mathbb{N}_{27} & & 
\\[1ex]
\mathcal{F}^{X}_+& = \mathcal{F}^{Y}_+ = \mathcal{F}^{Z}_+ = 0 \\
\forall z &\in \mathcal{R}(Z):  \mathcal{F}^{W}_+(z) =  z\\
\forall (x, y) & \in  \mathcal{R}(X) \times \mathcal{R}(Y):  \mathcal{F}^{S_1}_+(x,y) =  x + y \\
\forall (s_1, w) &\in \mathcal{R}(S_1) \times \mathcal{R}(W): \mathcal{F}^{S_2}_+(s_1,w) =  s_1 + w
\end{align*}


\subsection{Formal Definition of $\addnet$}

In the main text, we did not provide a specific identity for $\addnet$. Here, we define $\addnet$ to be a feed forward network, which we represent directly as a causal model $C_{N_+} = (\mathcal{V}_{N_+}, \mathcal{R}_{N_+}, \mathcal{F}_{N_+})$. The location $L_1$ from \Figref{fig:addmod-example} is the hidden unit $H_3$, the location $L_2$ is the hidden unit $H_1$.

Let  $W \in \mathbb{R}^{30 \times 3}$; for $k\in \{1,3\}$ let $W_{jk} = j \bmod 10$ if $0 \leq j \leq 20$, otherwise $W_{jk} = 0$, and let $W_{j2} = 0$ if $0 \leq j \leq 20$, otherwise $W_{j2} = j \bmod 10$. Let $U \in \mathbb{R}^{3}$ and $U = [1,1,0]$.
\begin{align*}
\mathcal{V}_{\addnet} &= \{D_x, D_y,D_z, H_1, H_2, H_3, O\} & &
\\[1ex]
\mathcal{R}_{\addnet}(D_x) &= \mathcal{R}_{\addnet}(D_y) = \mathcal{R}_{\addnet}(D_z) = \{0,1\}^{10} & &\\
\mathcal{R}_{\addnet}(O) &= \mathcal{R}_{\addnet}(H_1) =\mathcal{R}_{\addnet}(H_2) = \mathcal{R}_{\addnet}(H_3) = \mathbb{R} & &
\\[1ex]
\mathcal{F}^{D_x}_{\addnet} &= \mathcal{F}^{D_y}_{\addnet} = \mathcal{F}^{D_z}_{\addnet} = 0\\
\forall \mathbf{x}  \in  \mathcal{R}&_{\addnet}(D_x) \times \mathcal{R}_{\addnet}(D_y) \times \mathcal{R}_{\addnet}(D_z): [ \mathcal{F}_{\addnet}^{H_1}  (\mathbf{x}), \mathcal{F}_{\addnet}^{H_2}(\mathbf{x}), \mathcal{F}_{\addnet}^{H_3}(\mathbf{x}) ] = \text{ReLU}(\mathbf{x}W) \\
\forall \mathbf{h} \in \mathcal{R}&_{\addnet}(H_1) \times \mathcal{R}_{\addnet}(H_2) \times \mathcal{R}_{\addnet}(H_3):
\mathcal{F}_{\addnet}^O(\mathbf{h}) = \text{ReLU}(\mathbf{h}U) &
\end{align*}
This network uses one-hot representations $d_x, d_y, d_z \in \{0,1\}^{10}$ to represent 
inputs from $\mathbb{N}_9$.

\subsection{Proving $\addmod$ is an abstraction of $\addnet$}
We now prove that $C_+$ is an abstraction $C_{N_+}$

We define the mapping $\tau:  \mathcal{R}_{\addnet}(\mathcal{V}_{\addnet}) \to \mathcal{R}_+(\mathcal{V}_+)$ as follows. We first partition the variables of $N_+$ into cells: 
$P_{X} = \{D_x\}$, $P_{Y} = \{D_y\}$,  $P_{Z} = \{D_z\}$, 
$P_{W} = \{H_1\}$, $P_{S_1} = \{H_3\}$,   $P_{S_2} = \{O\}$, $P_{\varnothing} = \{H_2\}$. To define $\tau$ it suffices to define the component functions $\tau_V$ for $V \in \mathcal{V}_+$. Let $B :\{0,1\}^{10} \to \mathbb{N}_9$ be the partial function s.t.\ $B([v_1, v_2, \dots, v_{10}]) = k$ if $v_k = 1$ and $v_j = 0$ for $j \not = k$. Set $\tau_{X},\tau_{Y},\tau_{Z}$ all equal to $B$, and let $\tau_{W},\tau_{S_1},\tau_{S_2}$ all be the identity function. 

Let $\mathcal{I}_+$ be the set of all interventions on $\addmod$ that determine values for (at least) $X$, $Y$, and $Z$. Let $\mathcal{I}_{\addnet} = \textit{dom}(\omega_{\tau})$. That is, $\mathcal{I}_{N_+}$ includes exactly the (interventions corresponding to) projections of $\mathcal{R}_{\addnet}(\mathcal{V}_{\addnet})$ that map via $\omega_{\tau}$ to some admissible intervention on $C_+$. 
Because elements of $\mathcal{I}_+$ always determine values for $X,Y,Z$, every intervention in $\mathcal{I}_{N_+}$ determines a value for each of $D_x,D_y,D_z$. In fact, these values are guaranteed to be in the domains of $\tau_{X},\tau_{Y},\tau_{Z}$, respectively. 

We now prove the three conditions guaranteeing $(C_{+}, \mathcal{I}_+)$ is a $\tau$-abstraction of $(C_{N_+},\mathcal{I}_{N_+})$.

(1) The first point is that the map $\tau$ is surjective. Take an arbitrary $(x,y,z,w, s_1, s_2) \in \mathcal{R}_+(\mathcal{V}_+)$. We determine an element of $\mathcal{R}_{\addnet}(\mathcal{V}_{\addnet})$ as follows: $[d_x d_y d_z]= B^{-1}([x,y,z])$, $\left[h_1 h_2 h_3\right] = \left[s_1d_2s_1\right]$, and $o = s_2$. It's then clear that $\tau(d_x, d_y, d_z, h_1,h_2,h_3,o) = (x,y,z,w, s_1, s_2)$. As $(x,y,z,w, s_1, s_2)$ was chosen arbitrarily,  $\tau$ is surjective.



(2) The second point is that $\omega_\tau$ must also surject onto the set $\mathcal{I}_+$ of all interventions on $C_+$. Any intervention $i_+ \in \mathcal{I}_+$ can be identified with a vector $\mathbf{i}^+$ of values of variables in $\mathcal{V}_+$. By definition of $\mathcal{I}_+$, $i_+$ fixes at least the values of $X,Y,Z$. Consider the intervention $i_{N_+}$ that sets $D_x$, $D_y$, and $D_z$ to the one-hot representations of $X$, $Y$, and $Z$ for the values they were set. Furthermore, if $i_+$ sets $W$ to $w$ then $i_{N_+}$ sets $H_1$ to $w$ and if $i_+$ sets $S_2$ to $s_2$, then $i_{N_+}$ sets $H_3$ to $s_2$.
It suffices to show that $\omega_\tau(i_{N_+})=i_+$. In other words, we need to show that $\tau(\inverseproject{\mathbf{i}^{\addnet}}) = \inverseproject{\mathbf{i}^+}$.


First, we show for all $\mathbf{v}_L \in \inverseproject{\mathbf{i}^{\addnet}}$ that $\tau(\mathbf{v}_L) \in \inverseproject{\mathbf{i}^{+}}$. By construction of $i_+$, any variables fixed by $i_{\addnet}$ will correspond (via $\tau$ component functions) to values of variables fixed by $i_{+}$, except for the variable $H_3$, which has no corresponding high level variable. We merely need to observe that for any values of variables \textit{not} set by $i_{\addnet}$, there exist corresponding values for the variables that are \textit{not} set by $i_+$, such that the appropriate $\tau$ component functions map the former to the latter (with the exception of $H_3$, which has no corresponding high level variable).  This is obvious from the definition of the components of $\tau$.

Second, we show for all $\mathbf{v}_H \in \inverseproject{\mathbf{i}^+}$ there is $\mathbf{v}_L \in \inverseproject{\mathbf{i}^{\addnet}}$ such that $\tau(\mathbf{v}_L) = \mathbf{v}_H$. Again, by construction of $i_+$, any variables fixed by $i_{+}$ will correspond (via $\tau$ component functions) to values of  variables fixed by $i_{\addnet}$. We merely need to observe that for any values of variables \textit{not} set by $i_{+}$, there exist corresponding values for the variables \textit{not} set by $i_{\addnet}$, such that the appropriate $\tau$ component functions map the former to the latter, with $H_3$ taking on any value.  This is obvious from the definition of the components of $\tau$. This concludes the argument that $\omega_\tau(i_{N_+})=i_+$.

(3) Finally, we need to show for each $i_{N_+} \in \textit{dom}(\omega_\tau)$ that $\tau(i_{N_+}(C_{N_+})) = \omega_\tau(i_{N_+})(C_+)$. The point here is that the two causal processes unfold in the same way, under any intervention.

Indeed, pick any $i_{N_+}$ and suppose that $i_+ = \omega_\tau(i_{N_+})$. We know that $i_+$ fixes values $x,y,z$ of $X,Y,Z$, and likewise that $i_{N_+}$ fixes values $d_x,d_y,d_z$ of $D_x,D_y,D_z$ such that $\tau_{D_j}(x_j) = d_j$ for $j \in \{1,2,3\}$. Any other variables fixed by $i_+$ from among $W,S_1,S_2$ will likewise correspond (via $\tau_{W},\tau_{S_1},\tau_{S_2}$) to values of  $H_2,H_1,O$ fixed by $i_{N_+}$. We merely need to observe that any variables that are \emph{not} set by $i_{+}$ and $i_{N_+}$ will still correspond via the appropriate $\tau$-component, given their settings in $i_+(C_+)$ and $i_{N_+}(C_{N_+})$. The mechanisms in $C_{N_+}$ were devised  precisely to guarantee this.

Thus we have fulfilled the three requirements and we have shown that $C_+$ is an abstraction $C_{N_+}$.

The proof that $\natmod$ is a constructive abstraction of $\nlinet$ follows this same pattern.


\section{Causal Abstraction Analysis of $\natmod$}\label{app:natmod}

\newcommand{\natlog}{\textit{NatLog}}
\newcommand{\subj}{\textit{Subj}}
\newcommand{\obj}{\textit{Obj}}
\newcommand{\Adj}{\text{Adj}}
\newcommand{\NP}{\text{NP}}
\newcommand{\QP}{\text{QP}}
\newcommand{\N}{\text{N}}
\newcommand{\Q}{\text{Q}}

\subsection{Formal Definition of $\natmod$}

We formally define the model $\natmod = (\mathcal{V}_{\natlog}, \mathcal{R}_{\natlog}, \mathcal{F}_{\natlog})$ as follows:
\[
\mathcal{V}_{\natlog} = 
\left\{
\begin{array}{c}
\Q^P_{\subj}, \Q^H_{\subj}, \Neg^P_{\subj}, \Neg^H_{\subj}, \N^P_{\subj},
\N^H_{\subj},\Neg^P, \Neg^H, \Adv^P, \Adv^H, 
\\[1ex]
\V^P, \V^H,\Q^P_{\obj}, \Q^H_{\obj},\Neg^P_{\obj}, \Neg^H_{\obj}, \N^P_{\obj}, \N^H_{\obj}, 
Q_{\subj}, \Neg_{\subj}, N_{\subj}, \Neg, \Adv
\\[1ex]
Q_{\obj}, 
\Neg_{\obj}, \N_{\obj}, \NP_{\subj}, \VP, \NP_{\obj}, \QP_{\obj},
\NegP, \QP_{\subj}
\end{array}
\right\}
\]

\begin{align*}
\mathcal{R}_{\natlog}(\Q^P_{\subj}) &= \mathcal{R}_{\natlog}(\Q^H_{\subj}) = \mathcal{R}_{\natlog}(\Q^H_{\subj}) = \mathcal{R}_{\natlog}(\Q^H_{\subj})  \\
&= \{\textit{no, some, every, not every}\} & &\\
\mathcal{R}_{\natlog}(\mathit{\Neg}^P) & = \mathcal{R}_{\natlog}(\Neg^H) = \{not, \epsilon \} \\
\mathcal{R}_{\natlog}(\Neg^P_{\subj}) &= \mathcal{R}_{\natlog}(\Neg^H_{\subj}) = \bf{\Neg}_{\subj}\\
\mathcal{R}_{\natlog}(\N^P_{\subj}) &= \mathcal{R}_{\natlog}(\N^H_{\subj}) = \bf{N}_{\subj} \\
\mathcal{R}_{\natlog}(\Adv^P) &= \mathcal{R}_{\natlog}(\Adv^H) = \bf{\Adv}_{\subj}\\
\mathcal{R}_{\natlog}(\V^P) &= \mathcal{R}_{\natlog}(\V^H) = \bf{V}_{\subj} \\
\mathcal{R}_{\natlog}(\Neg^P_{\obj}) &= \mathcal{R}_{\natlog}(\Neg^H_{\obj}) = \bf{\Neg}_{\obj}\\
\mathcal{R}_{\natlog}(\N^P_{\obj}) &= \mathcal{R}_{\natlog}(\N^H_{\obj}) = \bf{N}_{\obj} \\
\mathcal{R}_{\natlog}(\Q_{\obj}) &= \mathcal{R}_{\natlog}(\Q_{\subj}) = \mathcal{Q}\\
\mathcal{R}_{\natlog}(\Neg) &= \mathcal{N} \\
\mathcal{R}_{\natlog}(\Neg_{\obj}) &= \mathcal{R}_{\natlog}(\Neg_{\subj}) = \mathcal{R}_{\natlog}(\Adv) = \mathcal{A} \\
\mathcal{R}_{\natlog}(\N_{\obj}) &= \mathcal{R}_{\natlog}(\N_{\subj}) = \mathcal{R}_{\natlog}(\V) =  \{\#, \equiv\} & & \\
\mathcal{R}_{\natlog}(\NP_{\subj}) &= \mathcal{R}_{\natlog}(\NP_{\obj}) = \mathcal{R}_{\natlog}(\VP) =  \{\#, \equiv, \sqsubset, \sqsupset \} & & \\ 
\mathcal{R}_{\natlog}(\QP_{\obj}) &= \mathcal{R}_{\natlog}(\NegP) = \mathcal{R}_{\natlog}(\QP_{\subj}) =  \{\#, \equiv, \sqsubset, \sqsupset, |,\ \hat{} \ , \smile \}
\\[1ex]
\mathcal{F}_{\N} &= \text{COMP} \text{ for } \N \in \{\VP, \NP_{\subj},\NP_{\obj}, \NegP, \QP_{\obj},\QP_{\subj}\}\\
\mathcal{F}_{\N} &= \text{REL} \text{ for } \N \in \{\V, \N_{\subj},\N_{\obj}\}\\
\mathcal{F}_{\N} &= \text{PROJ} \text{ for } \N \in \{\Q_{\obj}, \Q_{\subj}, \Adv, \Neg_{\subj}, \Neg_{\obj}, \Neg\}
\end{align*}
The set $\{\#, \equiv, \sqsubset, \sqsupset, |,\ \hat{} \ , \smile \}$ contains the seven relations used in the natural logic of \citet{MacCartney:07}. The set $\bf{N}_{\subj}$ contains the subject nouns used to create \MQNLI , $\bf{N}_{\obj}$ the set of object nouns, $\bf{Adj}_{\subj}$ the subject adjectives, $\bf{Adj}_{\obj}$ the object adjectives, $\bf{V}$ the verbs, and $\bf{\Adv}$ the adverbs. Additionally, $\mathcal{Q}$ is the set of joint projectivity signatures between \textit{every}, \textit{some}, \textit{not every}, and \textit{no}, $\mathcal{N}$ is the set of joint projectivity signatures between \textit{not} and $\epsilon$, $\mathcal{A}$ is the set of joint projectivity signatures between intersective adjectives and adverbs and $\epsilon$. REL$(x,y)$ outputs the lexical relation between $x$ and $y$. Finally, COMP$(f, x_1,x_2, \dots, x_n) = f(x_1,x_2, \dots, x_n)$ and PROJ$(f,g) = P_{f/g}$ where $P_{f/g}$ is the joint projectivity signature between $f$ and $g$. See \citet{Geiger-etal:2019} for details about these sets and functions.

\subsection{Formal Definition of $\natmod^N$}

For some non-leaf node $N$ of the tree in \figref{fig:bigtree}, we define $\natmod^{\nodevar}$ to be the marginalization of $\natmod$ where all variables are removed other than the input variables 
\[ 
\mathcal{V}_{\natlog}^{\textit{Input}} = \Q^P_{\subj}, \Q^H_{\subj}, \Neg^P_{\subj}, \Neg^H_{\subj}, \N^P_{\subj},
\N^H_{\subj},\Neg^P, \Neg^H, \Adv^P, \Adv^H, \V^P, \V^H,\]
\[\Q^P_{\obj}, \Q^H_{\obj},\Neg^P_{\obj}, \Neg^H_{\obj}, \N^P_{\obj}, \N^H_{\obj}
\]
along with the output variable $\QP_{\subj}$
and the intermediate variable $N$. For a definition of marginalization, see \citet{Bongers2016}.

\subsection{Formal definition of $\nlinet$}

In the main text, $\nlinet$ could represent either our BERT model or our LSTM model. We will maintain this ambiguity, because while these two models are drastically different at the highest level of detail, for the sake of our analysis we can view them both as creating a grid of neural representations where each representation in the grid is caused by all representations in the previous row and causes all representations in the following row. We will now formally define the causal model $C_{\nlinet}$.

\[\mathcal{V}_{\nlinet} = \{R_{11}, R_{12}, \dots, R_{1m}, \dots R_{nm}, O\}\]

For the LSTM model $n =2$ and for the BERT model $n = 12$. $m$ is the number of tokens in a tokenized version of an \MQNLI\ example. 
\[\mathcal{R}_{\nlinet}(R_{jk}) = \mathbb{R}^d \ \ \ \  \mathcal{R}_{\nlinet}(O) = \{\text{entailment, contradiction, neutral}\}\]

For all $j$ and $k$ and where $d$ is the dimension of the vector representations.
\[\forall (r_{(j-1)1}, r_{(j-1)2}, \dots ,r_{(j-1)m}) \in \mathcal{R}_{\nlinet}(R_{(j-1)1} \times R_{(j-1)2} \times \dots \times R_{(j-3)m})\]
\[ \mathcal{F}_{\nlinet}^{R_{jk}}(r_{(j-1)1}, r_{(j-1)2}, \dots ,r_{(j-1)m}) = \mathbf{NN}_{jk}(r_{(j-1)1}, r_{(j-1)2}, \dots ,r_{(j-1)m})\]

where $\mathbf{NN}_{jk}$ is either the LSTM function or the BERT function that creates the neural representation at the $j$th row and $k$th column.

\[\forall r_{n1} \in \mathcal{R}_{\nlinet}(R_{n1}) 
\mathcal{F}_{\nlinet}^{O}(r_{n1}) = \mathbf{NN}_O(r_{n1})\]

where $\mathbf{NN}_O$ is the neural network that makes a three class prediction using the final representation of the [CLS] token. 

See \appref{app:trainingdetails} for details about these functions.

\subsection{Proving $\natmod^N$ is an abstraction of $\nlinet$}

We will now formally prove that that $\natmod^N$ is a constructive abstraction of $\nlinet$ if the following holds for all $e, e' \in \MQNLI$, where the representation location $L$ is equivalent to the variable $R_{jk}$ for some $j$ and $k$. This would mean that every single one of our intervention experiments at this location are successful.
\begin{equation}
    \natmod^{N \leftarrow e'}(e) = \nlinet^{L \leftarrow e'}(e) 
\end{equation} 
We define the mapping $\tau: \mathcal{R}_{\nlinet}(\mathcal{V}_{\nlinet}) \to \mathcal{R}^N_{\natlog}(\mathcal{V}_{\natlog})$ as follows. We first partition the ``low level'' variables of $\nlinet$ into partition cells:
$$\begin{array}{lll}
P_{N} = \{L\} &
P_{\QP_{\subj}} = \{O\} &
\forall X \in \mathcal{V}_{\natlog}^{\textit{Input}}\\
P_{X} = \{R_{1j},R_{1(j+1)}, \dots, R_{1(j+k)} \}
\end{array}$$
where $R_{1j},R_{1(j+1)}, \dots, R_{1(j+k)}$ are the token vectors associated with the input variable $X$. Some of our causal model's input variables are tokenized into several tokens (see \appref{app:trainingdetails} for details).
 
 To define $\tau$, it then suffices to define the component functions $\tau_{V}$ for each $V \in \mathcal{V}_{\natlog}$. Let $T: (\mathbb{R}^d)^+ \to \mathcal{V}_{\natlog}^{\textit{Input}}$ be the partial function mapping sequences of token vectors to the input variable they correspond to, where $+$ is the Kleene plus operator. Let $P: \mathcal{R}^3 \to \{\text{entailment, neutral, contradiction}\}$ be the partial function mapping a vector of logits to the output prediction they correspond to. Finally, let $Q_L: \mathbb{R}^d \to \mathcal{R}_{\natlog}(N)$  be the partial function such that for all $e \in \MQNLI$, if $\mathbf{v}$ is the vector created by $\nlinet$ at location $L$ when processing input $e$ and $x$ is the value realized by $\natmod$ for the variable $N$ when processing input $e$, then $Q_L(\mathbf{v}) = x$. 
 
 For all $\forall X \in \mathcal{V}_{\natlog}^{\textit{Input}}$, we set $\tau_{X}$ to be $T$. We additionally set $\tau_N$ to be $Q_L$ and $\tau_{\QP_{\subj}}$ to be $P$.
 
 Let $\mathcal{I}_{\natlog}$ be the set of all interventions on $\natmod$ that intervene on (i.e., determine the values for) at least the elements of $\mathcal{V}_{\natlog}^{\textit{Input}}$. Let $\mathcal{I}_{\nlinet}$ be the set of interventions that is the domain of the partial function $\omega_{\tau}$. In other words, $\mathcal{I}_{\nlinet}$ includes exactly the projections of $\mathcal{R}_{\nlinet}(\mathcal{V}_{\nlinet})$ that map via $\omega_{\tau}$ to some intervention on $\addmod$. The fact that $P$, $Q_L$, an $T$ are all proper partial functions prevent $\mathcal{I}_{\nlinet}$ from including all possible interventions on $C_{\nlinet}$.
 
 We now prove the three conditions that must hold for $(\natmod, \mathcal{I}_{\natlog})$ to be a $\tau$-abstraction of $(C_{\nlinet}, \mathcal{I}_{\nlinet})$.
 
 (1) The first point is to show the map $\tau$ is surjective. So take an arbitrary element $(\vec{v}^{\textit{input}}, n, q) \in \mathcal{R}_{\natlog}(\mathcal{V}_{\natlog})$. We specify an element of $\mathcal{R}_{\nlinet}(\mathcal{V}_{\nlinet})$ as follows:
 
 $$\begin{array}{lll}
l = Q_L^{-1}(n) & o = P^{-1}(q) \\
\\
\forall v^{\textit{input}} \in \vec{v}^{\textit{input}}  T^{-1}(v^{\textit{input}}) = (r_{1j},r_{1(j+1)}, \dots, r_{1(j+k)})
\end{array}$$
where $r_{1j},r_{1(j+1)}, \dots, r_{1(j+k)}$ are the token vectors corresponding to the input variable $v^{\textit{input}}$. 

It's then patent that $\tau(r_{11}, \dots, r_{n1}, r_{12}, \dots r_{nm}, o) = (\vec{v}^{\textit{input}}, n, q)$. As $(\vec{v}^{\textit{input}}, n, q)$ was chosen arbitrarily, we have shown $\tau$ is surjective.

(2) The second point is that $\omega_{\tau}$ must also be surjective onto the set $\mathcal{I}_{\natlog}$ of interventions on $\natmod$. Any intervention $i_{\natlog} \in \mathcal{I}_{\natlog}$ can be identified with with a vector $\mathbf{i}^{\natlog}$ of values of variables in $\mathcal{V}_{\natlog}$. By the definition of $\mathcal{I}_{\natlog}$, $i_{\natlog}$ fixes the values of the variables in $\V^{input}$ and may also determine $N$ and/or $\QP_{\textit{\subj}}$. Consider the intervention $i_{\nlinet}$ corresponding to $\mathbf{i^{\nlinet}} = \tau^{-1}(\mathbf{i}^{\natlog})$ as described in Section \ref{section:constructive-abstraction}. It suffices to show that $\omega_\tau(i_{\nlinet})=i_{\natlog}$. In other words, we need to show parts 1, 2, and 3 from the definition above.

Part 1 is clear, since by the definition of $\mathcal{I}_{\natlog}$ we are guaranteed that $\mathbf{i}^{\natlog}$ determines values for $\mathbf{V}^{\textit{input}}$, and hence $\mathbf{i}^{\nlinet}$ fixes values for $R_{11}, \dots, R_{1m}$ in the domains of $\tau_{\V^{\textit{input}}}$ for $V \in \mathbf{V}^{\textit{input}}$. Then any intervention that intervenes only on the values of 

Part 2 requires that for every $\mathbf{v}_{\nlinet} \in \inverseproject{\mathbf{i}^{\nlinet}}$, we have $\tau(\mathbf{v}_{\nlinet}) \in \inverseproject{\mathbf{i^{\natlog}}}$. Because of how we defined $i_{\natlog}$, any variables fixed by $i_{\nlinet}$ will correspond (via $\tau$ component functions) to values of variables fixed by $i_{\natlog}$, except for the variables $R_{jk} \not \in \mathbf{V}^{\textit{input}} \cup \{ L\}$, which have no corresponding high level variables. We merely need to observe that, for any values for the variables that are \textit{not} set by $i_{\nlinet}$, there exists corresponding values for the variables that are \textit{not} set by $i_{\natlog}$ such that the appropriate $\tau$ component functions map the former to the latter, except for the variables $R_{jk} \not \in \mathbf{V}^{input} \cup \{ L\}$, which, again, have no corresponding high level variables. This is plainly obvious from the definition of the components of $\tau$. 

Part 3 requires that for any $\mathbf{v}_{\natlog} \in \inverseproject{\mathbf{i}^{\natlog}}$, there exists a $\mathbf{v}_{\nlinet} \in \inverseproject{\mathbf{i}^{\nlinet}}$ such that $\tau(\mathbf{v}_{\nlinet}) = \mathbf{v}_{\natlog}$. Again, because of how we defined $i_{\natlog}$, any variables fixed by $i_{\natlog}$ will correspond (via $\tau$ component functions) to values of variables fixed by $i_{\nlinet}$. We merely need to observe that for any values for the variables that are \textit{not} set by $i_{\natlog}$, there exists corresponding values for the variables that are \textit{not} set by $i_{\nlinet}$, such that the appropriate $\tau$ component functions map the former to the latter, with  $R_{jk} \not \in \mathbf{V}^{\textit{input}} \cup \{ L\}$ taking on any value. This is plainly obvious from the definition of the components of $\tau$. 

Thus, we have shown that $\omega_{\tau}(i_{\nlinet}) = i_{\natlog}$. 

(3) Finally, we need to show for each $i_{\nlinet} \in \textit{dom}(\omega_{\tau})$ that $\tau(i_{\nlinet}(C_{\nlinet})) = \omega_{\tau}(i_{\nlinet})(\natmod)$. The point here is that the two causal processes unfold in the same way, under any intervention. 
Indeed, pick any $i_{\nlinet}$ and suppose that $i_{\natlog} = \omega_{\tau}(i_{\nlinet})$. We know that $i_{\natlog}$ fixes values for the variables in $\mathbf{V}^{input}$, and likewise that $i_{\nlinet}$ fixes values for the variables $R_{11}, \dots, R_{1m}$. Any other variables fixed by $i_{\natlog}$ from among $N, \QP_{\subj}$ will likewise correspond (via the component functions of $\tau$) to values of $L$ and $O$. We merely need to observe that any variables that are \textit{not} set by $i_{\natlog}$ and $i_{\nlinet}$ will still correspond via the appropriate $\tau$-component, given their settings in $i_{\natlog}(\natmod)$ and $i_{\nlinet}(C_{\nlinet})$. The intervention experiments on $\nlinet$ that we are assuming were successful were devised precisely to guarantee this.

We have thus fulfilled the three requirements and shown that $\natmod$ is an abstraction of $C_{\nlinet}$.

\end{document}


\section{Causal Abstraction Analysis of $\natmod$}\label{app:natmod}
This document provides a formal proof that interchange experiments in this codebase show that $\natmode{C}^N_{\NatLog}$ for some $N$ is an abstraction of $\nlinet$. 

\subsection{Formal Definition of $\natmod$}

We formally define the model $\natmod = (\mathcal{V}_{NatLog}, \mathcal{R}_{NatLog}, \mathcal{F}_{NatLog})$ as follows.

$\mathcal{V}_{NatLog} = 
\{Q^P_{Subj}, Q^H_{Subj}, Adj^P_{Subj}, Adj^H_{Subj}, N^P_{Subj},
N^H_{Subj},Neg^P, Neg^H, Adv^P, Adv^H, V^P, V^H, \\
Q^P_{Obj}, Q^H_{Obj},Adj^P_{Obj}, Adj^H_{Obj}, N^P_{Obj}, N^H_{Obj}, 
Q_{Subj}, Adj_{Subj}, N_{Subj}, Neg, Adv, V, Q_{Obj}, \\
Adj_{Obj}, N_{Obj}, NP_{Subj}, VP, NP_{Obj}, QP_{Obj},
NegP, QP_{Subj}\} $
\newline
\newline
$\mathcal{R}_{NatLog}(Q^P_{Subj}) = \mathcal{R}_{NatLog}(Q^H_{Subj}) = \mathcal{R}_{NatLog}(Q^H_{Subj}) = \mathcal{R}_{NatLog}(Q^H_{Subj}) = \{no, some, every, not every\}$
\newline
\newline
$\mathcal{R}_{NatLog}(Neg^P) = \mathcal{R}_{NatLog}(Neg^H) = \{not, \epsilon \}$
\newline
\newline
$\mathcal{R}_{NatLog}(Adj^P_{Subj}) = \mathcal{R}_{NatLog}(Adj^H_{Subj}) = \bf{Adj}_{Subj}$
\newline
\newline
$\mathcal{R}_{NatLog}(N^P_{Subj}) = \mathcal{R}_{NatLog}(N^H_{Subj}) = \bf{N}_{Subj}$
\newline
\newline
$\mathcal{R}_{NatLog}(Adv^P) = \mathcal{R}_{NatLog}(Adv^H) = \bf{Adv}_{Subj}$
\newline
\newline
$\mathcal{R}_{NatLog}(V^P) = \mathcal{R}_{NatLog}(V^H) = \bf{V}_{Subj}$
\newline
\newline
$\mathcal{R}_{NatLog}(Adj^P_{Obj}) = \mathcal{R}_{NatLog}(Adj^H_{Obj}) = \bf{Adj}_{Obj}$
\newline
\newline
$\mathcal{R}_{NatLog}(N^P_{Obj}) = \mathcal{R}_{NatLog}(N^H_{Obj}) = \bf{N}_{Obj}$
\newline
\newline
$\mathcal{R}_{NatLog}(Q_{Obj}) = \mathcal{R}_{NatLog}(Q_{Subj}) = \mathcal{Q}$
\newline
\newline
$\mathcal{R}_{NatLog}(Neg) = \mathcal{N}$
\newline
\newline
$\mathcal{R}_{NatLog}(Adj_{Obj}) = \mathcal{R}_{NatLog}(Adj_{Subj}) = \mathcal{R}_{NatLog}(Adv) = \mathcal{A}$
\newline
\newline
$\mathcal{R}_{NatLog}(N_{Obj}) = \mathcal{R}_{NatLog}(N_{Subj}) = \mathcal{R}_{NatLog}(V) =  \{\#, \equiv\}$
\newline
\newline
$\mathcal{R}_{NatLog}(NP_{Subj}) = \mathcal{R}_{NatLog}(NP_{Obj}) = \mathcal{R}_{NatLog}(VP) =  \{\#, \equiv, \sqsubset, \sqsupset \}$
\newline
\newline
$\mathcal{R}_{NatLog}(QP_{Obj}) = \mathcal{R}_{NatLog}(NegP) = \mathcal{R}_{NatLog}(QP_{Subj}) =  \{\#, \equiv, \sqsubset, \sqsupset, |,\ \hat{} \ , \smile \}$
\newline
\newline

The set $\{\#, \equiv, \sqsubset, \sqsupset, |,\ \hat{} \ , \smile \}$ contains the seven relations used in the natural logic of \citet{MacCartney:07}. The set $\bf{N}_{Subj}$ contains the subject nouns used to create \MQNLI , $\bf{N}_{Obj}$ the set of object nouns, $\bf{Adj}_{Subj}$ the subject adjectives, $\bf{Adj}_{Obj}$ the object adjectives, $\bf{V}$ the verbs, and $\bf{Adv}$ the adverbs. Additionally, $\mathcal{Q}$ is the set of joint projectivity signatures between \textit{every}, \textit{some}, \textit{not every}, and \textit{no}, $\mathcal{N}$ is the set of joint projectivity signatures between \textit{not} and $\epsilon$, $\mathcal{A}$ is the set of joint projectivity signatures between intersective adjectives and adverbs and $\epsilon$. Finally, COMP$(f, x_1,x_2, \dots, x_n) = f(x_1,x_2, \dots, x_n)$ and PROJ$(f,g) = P_{f/g}$ where $P_{f/g}$ is the joint projectivity signature between $f$ and $g$. See \citet{Geiger-etal:2019} for details about these sets and functions.

\subsection{Formal Definition of $\natmod^N$}

For a non-leaf node $N$ of the tree in Figure 2, define $C^N_{Natlog}$ to be the \emph{marginalization} of $C_{Natlog}$, with all variables removed other than the input variables $\mathcal{V}_{NatLog}^{Input} = \{Q^P_{Subj}, Q^H_{Subj}, Adj^P_{Subj}, Adj^H_{Subj}, N^P_{Subj}$,
$N^H_{Subj},Neg^P, Neg^H, Adv^P, Adv^H, V^P, V^H,
Q^P_{Obj}, Q^H_{Obj},Adj^P_{Obj}, Adj^H_{Obj}, N^P_{Obj}, N^H_{Obj}\}$, the ``output'' variable $QP_{Subj}$, and the intermediate variable $N$.  (For a definition of marginalization, see, e.g.,  \citet{Bongers2016}. In the present deterministic setting the probabilistic aspects of marginalize are not relevant.)

\subsection{Formal definition of $\nlinet$}

In the main text, $\nlinet$ could represent either our BERT model or our LSTM model. We will maintain this ambiguity, because while these two models are drastically different at the finest level of detail, for the sake of our analysis we can view them both as creating a grid of neural representations where each representation in the grid is caused by all representations in the previous row and causes all representations in the following row. We will now formally define the causal model $C_{\nlinet}$.
\newline

$\mathcal{V}_{\nlinet} = \{R_{11}, R_{12}, \dots, R_{1m}, \dots R_{nm}, O\}$
\newline

\noindent For the LSTM model $n =2$ and for the BERT model $n = 12$. $m$ is the number of tokens in a tokenized version of an \MQNLI\ example. 
\newline
\newline
$\mathcal{R}_{\nlinet}(R_{jk}) = \mathbb{R}^d \\ \mathcal{R}_{\nlinet}(O) = \{\text{entailment, contradiction, neutral}\}$ 
\newline

\noindent This is for all $j$ and $k$ and where $d$ is the dimension of the vector representations.
\newline

$\forall (r_{(j-1)1}, r_{(j-1)2}, \dots ,r_{(j-1)m}) \in \mathcal{R}_{\nlinet}(R_{(j-1)1} \times R_{(j-1)2} \times \dots \times R_{(j-3)m}): \\ \mathcal{F}_{\nlinet}^{R_{jk}}(r_{(j-1)1}, r_{(j-1)2}, \dots ,r_{(j-1)m}) = \mathbf{NN}_{jk}(r_{(j-1)1}, r_{(j-1)2}, \dots ,r_{(j-1)m})$
\newline

\noindent where $\mathbf{NN}_{jk}$ is either the LSTM function or the BERT function that creates the neural representation at the $j$th row and $k$th column.
\newline
\newline
$\forall r_{n1} \in \mathcal{R}_{\nlinet}(R_{n1}): 
\mathcal{F}_{\nlinet}^{O}(r_{n1}) = \mathbf{NN}_O(r_{n1})$ 
\newline

\noindent where $\mathbf{NN}_O$ is the neural network that makes a three class prediction using the final representation of the [CLS] token. 

\subsection{Proving $\natmod^N$ is an abstraction of $\nlinet$}

We will now formally prove that $\natmod^N$ is a constructive abstraction of $\nlinet$ if the following holds for all $e, e' \in MQNLI$ where the representation location $L$ is equivalent to the variable $R_{jk}$ for some $j$ and $k$. This would mean that every single one of our intervention experiments at this location is successful.

\begin{equation}
    \natmod^{N \leftarrow e'}(e) = \nlinet^{L \leftarrow e'}(e) 
\end{equation} 

We define the mapping $\tau: \mathcal{R}_{\nlinet}(\mathcal{V}_{\nlinet}) \to \mathcal{R}^N_{NatLog}(\mathcal{V}_{NatLog})$ as follows. We first partition the ``low level'' variables of $\nlinet$ into partition cells. 
$$\begin{array}{lll}
P_{N} = \{L\} & P_{QP_{Subj}} = \{O\} \\
\forall X \in \mathcal{V}_{NatLog}^{Input}: P_{X} = \{R_{1j},R_{1(j+1)}, \dots, R_{1(j+k)} \}
\end{array}$$

 \noindent where $R_{1j},R_{1(j+1)}, \dots, R_{1(j+k)}$ are the token vectors associated with the input variable $X$. Some of our causal model's input variables are tokenized into several tokens.
 
 To define $\tau$, it then suffices to define the component functions $\tau_{V}$ for each $V \in \mathcal{V}_{NatLog}$. Let $T: (\mathbb{R}^d)^+ \to \mathcal{V}_{NatLog}^{Input}$ be the partial function mapping sequences of token vectors to the input variable they correspond to, where $+$ is the Kleene plus operator. Let $P: \mathbb{R}^3 \to \{\text{entailment, neutral, contradiction}\}$ be the partial function mapping a vector of logits to the output prediction they correspond to. Finally, let $Q_L: \mathbb{R}^d \to \mathcal{R}_{NatLog}(N)$  be the partial function such that for all $e \in \MQNLI$, if $\mathbf{v}$ is the vector created by $\nlinet$ at location $L$ when processing input $e$ and $x$ is the value realized by $\natmod$ for the variable $N$ when processing input $e$, then $Q_L(\mathbf{v}) = x$. 
 
 For all $X \in \mathcal{V}_{NatLog}^{Input}$, we set $\tau_{X}$ to be $T$. We additionally set $\tau_N$ to be $Q_L$ and $\tau_{QP_{Subj}}$ to be $P$.
 
 Let $\mathcal{I}_{NatLog}$ be the set of all interventions on $\natmod$ that intervene on (i.e., determine the values for) at least the elements of $\mathcal{V}_{NatLog}^{Input}$. Let $\mathcal{I}_{\nlinet}$ be the set of interventions that is the domain of the partial function $\omega_{\tau}$. In other words, $\mathcal{I}_{\nlinet}$ includes exactly the projections of $\mathcal{R}_{\nlinet}(\mathcal{V}_{\nlinet})$ that map via $\omega_{\tau}$ to some intervention on $\addmod$. The fact that $P$, $Q_L$, an $T$ are all proper partial functions prevent $\mathcal{I}_{\nlinet}$ from including all possible interventions on $C_{\nlinet}$.
 
 We now prove the three conditions that must hold for $(\natmod, \mathcal{I}_{NatLog})$ to be a $\tau$-abstraction of $(C_{\nlinet}, \mathcal{I}_{\nlinet})$.
 
 (1) The first point is to show the map $\tau$ is surjective. So take an arbitrary element $(\mathbf{v}^{input}, n, q) \in \mathcal{R}_{NatLog}(\mathcal{V}_{NatLog})$. We specify an element of $\mathcal{R}_{\nlinet}(\mathcal{V}_{\nlinet})$ as follows:
 
 $$\begin{array}{lll}
l = Q_L^{-1}(n) & o = P^{-1}(q) \\
\forall v^{input} \in \mathbf{v}^{input}: 
T^{-1}(v^{input}) = (r_{1j},r_{1(j+1)}, \dots, r_{1(j+k)})
\end{array}$$
where $r_{1j},r_{1(j+1)}, \dots, r_{1(j+k)}$ are the token vectors corresponding to the input variable $v^{input}$. 

It's then patent that $\tau(r_{11}, \dots, r_{n1}, r_{12}, \dots r_{nm}, o) = (\vec{v}^{input}, n, q)$. As $(\vec{v}^{input}, n, q)$ was chosen arbitrarily, we have shown $\tau$ is surjective.

(2) The second point is that $\omega_{\tau}$ must also be surjective onto the set $\mathcal{I}_{NatLog}$ of interventions on $\natmod$. Any intervention $i_{NatLog} \in \mathcal{I}_{NatLog}$ can be identified with a vector $\mathbf{i}^{NatLog}$ of values of variables in $\mathcal{V}_{NatLog}$. By the definition of $\mathcal{I}_{NatLog}$, $i_{NatLog}$ fixes the values of the variables in $V^{input}$ and may also determine $N$ and/or $QP_{Subj}$. Consider the intervention $i_{\nlinet}$ corresponding to $\mathbf{i^{\nlinet}} = \tau^{-1}(\mathbf{i}^{NatLog})$. It suffices to show that $\omega_\tau(i_{\nlinet})=i_{NatLog}$. 


First, we show that for every $\mathbf{v}_{\nlinet} \in \inverseproject{\mathbf{i}^{\nlinet}}$, we have $\tau(\mathbf{v}_{\nlinet}) \in \inverseproject{\mathbf{i^{NatLog}}}$. Because of how we defined $i_{NatLog}$, any variables fixed by $i_{\nlinet}$ will correspond (via $\tau$ component functions) to values of variables fixed by $i_{NatLog}$, except for the variables $R_{jk} \not \in \mathbf{V}^{input} \cup \{ L\}$, which have no corresponding high level variables. We merely need to observe that, for any values for the variables that are \textit{not} set by $i_{\nlinet}$, there exist corresponding values for the variables that are \textit{not} set by $i_{NatLog}$ such that the appropriate $\tau$ component functions map the former to the latter, except for the variables $R_{jk} \not \in \mathbf{V}^{input} \cup \{ L\}$, which, again, have no corresponding high level variables. This is plainly obvious from the definition of the components of $\tau$. 

Second, we show that for any $\mathbf{v}_{NatLog} \in \inverseproject{\mathbf{i}^{NatLog}}$, there exists a $\mathbf{v}_{\nlinet} \in \inverseproject{\mathbf{i}^{\nlinet}}$ such that $\tau(\mathbf{v}_{\nlinet}) = \mathbf{v}_{NatLog}$. Again, because of how we defined $i_{NatLog}$, any variables fixed by $i_{NatLog}$ will correspond (via $\tau$ component functions) to values of variables fixed by $i_{\nlinet}$. We merely need to observe that for any values for the variables that are \textit{not} set by $i_{NatLog}$, there exists corresponding values for the variables that are \textit{not} set by $i_{\nlinet}$, such that the appropriate $\tau$ component functions map the former to the latter, with  $R_{jk} \not \in \mathbf{V}^{input} \cup \{ L\}$ taking on any value. This is plainly obvious from the definition of the components of $\tau$. 

Thus, we have shown that $\omega_{\tau}(i_{\nlinet}) = i_{NatLog}$. 

(3) Finally, we need to show for each $i_{\nlinet} \in dom(\omega_{tau})$ that $\tau(i_{\nlinet}(C_{\nlinet})) = \omega_{\tau}(i_{\nlinet})(\natmod)$. The point here is that the two causal processes unfold in the same way, under any intervention. 
Indeed, pick any $i_{\nlinet}$ and suppose that $i_{NatLog} = \omega_{\tau}(i_{\nlinet})$. We know that $i_{NatLog}$ fixes values for the variables in $\mathbf{V}^{input}$, and likewise that $i_{\nlinet}$ fixes values for the variables $R_{11}, \dots, R_{1m}$. Any other variables fixed by $i_{NatLog}$ from among $N, QP_{Subj}$ will likewise correspond (via the component functions of $\tau$) to values of $L$ and $O$. We merely need to observe that any variables that are \textit{not} set by $i_{NatLog}$ and $i_{\nlinet}$ will still correspond via the appropriate $\tau$-component, given their settings in $i_{NatLog}(\natmod)$ and $i_{\nlinet}(C_{\nlinet})$. The intervention experiments on $\nlinet$ that we are assuming were successful were devised precisely to gaurantee this.

Thus we have fulfilled the three requirements and we have shown that $\natmod$ is an abstraction of $C_{\nlinet}$.

\bibliography{anthology,custom}
\bibliographystyle{acl_natbib}